\documentclass[final, envcountsect]{svjour3}
\usepackage[utf8]{inputenc} 
\usepackage[T1]{fontenc}    
\usepackage{amsmath,amsfonts,amssymb}
\usepackage{lscape}
\usepackage{lmodern}
\usepackage{hyperref}       
\usepackage{url}            
\usepackage{booktabs}       
\usepackage{amsfonts}       
\usepackage{nicefrac}       
\usepackage{microtype}      
\usepackage{lipsum}         
\usepackage{graphicx}
\usepackage{siunitx}
\usepackage[numbers]{natbib}
\usepackage{pdfcomment}
\usepackage{threeparttable}
\usepackage{siunitx}
\usepackage{setspace}
\usepackage{geometry}
\usepackage{orcidlink}
\usepackage{doi}
\usepackage{subcaption}
\usepackage{enumitem}
\usepackage{calc}
\usepackage{multirow}
\usepackage{placeins}
\usepackage{multicol}
\usepackage{cleveref}
\usepackage{mdframed}
\usepackage{setspace}
\usepackage{tikz}
\usepackage{comment}
\usepackage{array}
\usepackage[linesnumbered,ruled,vlined]{algorithm2e}
\SetKwRepeat{Do}{do}{while}

\newcolumntype{C}{>{\centering\arraybackslash}m{6em}}
\SetKw{KwIn}{in}

\DeclareMathOperator{\vech}{vech}
\DeclareMathOperator{\vect}{vec}

\newlist{Properties}{enumerate}{2}
\setlist[Properties]{label=Property \arabic*., font=\textbf, itemindent=*}

\begin{document}
\raggedbottom

\title{Learning Active Subspaces and Discovering Important Features with Gaussian Radial Basis Functions Neural Networks}

\author{{Danny D'Agostino 
\orcidlink{https://orcid.org/0000-0003-0028-4720}}, Ilija Ilievski, Christine Annette Shoemaker}
\institute{National University of Singapore \\ Department of Industrial Systems Engineering and Management \\Singapore \\ \email{\texttt{dannydag@nus.edu.sg}}
}
\titlerunning{ }

\maketitle
\begin{abstract}
Providing a model that achieves a strong predictive performance and is simultaneously interpretable by humans is one of the most difficult challenges in machine learning research due to the conflicting nature of these two objectives. 
To address this challenge, we propose a modification of the radial basis function neural network model by equipping its Gaussian kernel with a learnable precision matrix.
We show that precious information is contained in the spectrum of the precision matrix that can be extracted once the training of the model is completed. 
In particular, the eigenvectors explain the directions of maximum sensitivity of the model revealing the active subspace and suggesting potential applications for supervised dimensionality reduction.
At the same time, the eigenvectors highlight the relationship in terms of absolute variation between the input and the latent variables, thereby allowing us to extract a ranking of the input variables based on their importance to the prediction task enhancing the model interpretability.
We conducted numerical experiments for regression, classification, and feature selection tasks, comparing our model against popular machine learning models, the state-of-the-art deep learning-based embedding feature selection techniques, and a transformer model for tabular data. 
Our results demonstrate that the proposed model does not only yield an attractive prediction performance compared to the competitors but also provides meaningful and interpretable results that potentially could assist the decision-making process in real-world applications. 
A PyTorch implementation of the model is available on GitHub at the following link.\footnote{\url{https://github.com/dannyzx/Gaussian-RBFNN}} 
\end{abstract}

%
\section{Introduction}
The radial basis function (RBF) is a family of models used for function interpolation and approximation that are defined as a linear combination of radially symmetric basis functions \cite{broomhead1988radial}. 
The RBF approach has many properties that make it attractive as a mathematical tool for interpolation \cite{micchelli1984interpolation, powell2001radial}.
Once the basis function and its hyperparameters are determined, the weights that multiply the basis functions can be found by solving a convex optimization problem or directly through matrix inversion.
The RBF model has been generalized in the context of approximation by using basis functions centered on a subset of the data, that can be interpreted as one hidden layer neural network (RBFNN) with RBF's activation function as shown in \cite{broomhead1988radial}. 
In \cite{park1991universal, park1993approximation} the authors showed that under some conditions on the basis function, RBFNNs are universal approximators as neural networks (NNs) \cite{hornik1989multilayer}. 

RBFs have been used for function interpolation or approximation for many decades in different applications. 
In the work presented in \cite{poggio1990networks,girosi1995regularization} the authors showed that from the regularization principles and through a solution of a variational problem, the RBF model is a subclass of regularization networks.
Within this family, the Gaussian radial basis function neural network (GRBFNN) is a particular case of the RBF approach, defined by employing Gaussian kernels as activation functions.
The Gaussian kernel exhibits flexibility by assuming various forms based on the covariance matrix, offering diverse strategies for capturing relationships within data.
The standard Gaussian kernel, frequently employed, assumes an isotropic covariance, meaning the covariance matrix is a diagonal matrix, and all elements on the diagonal are equal. Governed by its shape (i.e. width) parameter, this kernel treats each feature equally. Given that only one parameter requires determination, the shape parameter is commonly treated as a hyperparameter, allowing users to define it in advance \cite{bishop1995neural, mongillo2011choosing}.
However, the shape parameter need not be a fixed constant but can be learnable which means, that during the training process, the shape parameter is adjusted and optimized alongside the other model parameters, solving in this case a nonconvex optimization problem \cite{cavoretto2021search, zheng2020optimal}.
A diagonal covariance matrix, accommodates scenarios where distinct variances along each feature axis are necessary. 
Real-world situations often introduce challenges where features display different variances or correlations. To address this, employing a Gaussian kernel with a full covariance matrix becomes essential. Hence, the Mahalanobis distance Gaussian kernel, with its utilization of a full covariance matrix, adeptly captures both feature scaling and inter-feature correlations. 

In general, various methods have been proposed to estimate the model parameters in the context of approximation. 
Some of them are inspired by the work presented in \cite{poggio1990networks} where the location of the centers and a weighted norm (instead of the classical Euclidean norm) are considered part of the learning problem together with the weights. 
The possibility of using a superposition of kernels with a distinct set of hyperparameters has been also considered \cite{poggio1990networks}.
In \cite{moody1989fast} they propose to compute the width factors by the nearest neighbors heuristic and a clustering procedure for the centers. 
A different approach has been used in \cite{wettschereck1991improving}, where the centers' locations are considered as additional parameters of the optimization problem as well as the weights. In the same work, they also considered learning the width of the Gaussian kernel around each center. A similar approach has been presented in \cite{schwenker2001three} where a diagonal precision matrix is also considered for each RBF center. 
In \cite{buzzi2001convergent} they developed a Gauss-Seidel decomposition type algorithm \cite{bertsekas1997nonlinear} where the optimization of the weights and the centers are alternated along the iterations.
In \cite{zhang2019efficient}, a two-stage fuzzy clustering approach is introduced to partition the input space into multiple overlapping regions, subsequently utilized for constructing a local GRBFNN. Another noteworthy technique, referred to as variable projection \cite{zheng2023variable}, is employed to reduce the parameter count in the optimization problem associated with GRBFNN. Furthermore, in the study conducted in \cite{xing2023separable}, an enhancement to the computational speed of the model is achieved by leveraging the separability property inherent in the Gaussian basis function. In the work presented in \cite{lin2023optimized}, they suggest a modification to the activation function, incorporating a raised cosine radial basis function modulated by an exponential function. Recently, an accelerated gradient-based method has been presented in \cite{han2021accelerated} to improve the learning performance of the RBFNN model.
Important research to improve the generalization power of RBFNNs is in \cite{bishop1991improving}. This has been achieved by adding a regularization term that penalizes the second derivative of the output of each neuron. The technique is also known in the case of NNs in \cite{bishop1992curvature}. 

As in the case of NNs, RBFNNs are considered black-box models, and consequently, the underlying process of how the input features are used to make predictions is unclear to humans, including those who developed the models. 
Explainable AI (XAI) is a rapidly growing field of research that aims to make AI models more transparent and interpretable to humans. 
XAI techniques provide insight into the decision-making processes of AI models, allowing users to understand how models arrive at their outputs and to identify potential biases or errors.
For this reason, sometimes simpler models given just by a linear combination of the input variables are preferred since the coefficients can assess the importance of each feature in the prediction task. 
On the other hand, simpler models tend to be less accurate than complex ones. 
As a result, it is crucial to propose models with powerful predictive capabilities that can also provide simple explanations to support decision-making in complex real-world applications.
Thus, recognizing the importance of each input feature in a prediction task from a machine learning model has significant implications in various fields, including genomics \cite{bi2020interpretable}, environmental science \cite{dikshit2021interpretable}, emergency medicine \cite{wong2022development}, cancer research \cite{hauser2022explainable}, and finance \cite{ohana2021explainable}. In these domains, the model interpretability is crucial as the predictive performance as described in \cite{guidotti2018survey, arrieta2020explainable, adadi2018peeking}. 

Feature importance ranking and feature selection are two key techniques used in XAI.
Feature importance ranking is a fundamental aspect of machine learning that addresses the question of how individual features contribute to the predictive performance of a model. For example in linear regression, each of the coefficients represents the rate of change of the conditional mean of the target/response variable to the corresponding feature when all the others are kept fixed. In tree-based models, variable importance is assessed by considering the reduction in impurity or information gain attributed to each feature. Features with higher impurity reduction or information gain are deemed more crucial in the decision-making process of the model. 
Considering the complexity of modern machine learning models, understanding the significance of each input variable is crucial for model interpretability, transparency, and effective decision-making. 

Obtaining the feature importance ranking can accommodate performing the feature selection process. Feature selection involves choosing a subset of the data features to improve model performance due to the curse of dimensionality \cite{bellman1966dynamic}.
According to the taxonomy in \cite{guyon2003introduction} there are three kinds of feature selection methods: filter, wrapper, and embedded.
Filter methods for feature selection are techniques that use statistical measures to rank the importance of each feature independently of the model, such as Chi-squared feature selection and variance thresholding.  
These methods are called "filters" because they filter out irrelevant or redundant features from the dataset before the learning algorithm is applied. 
Wrapper methods work by selecting a subset of features, training the learning algorithm on that subset, evaluating its performance using cross-validation, and then repeating the process with different feature subsets. 
This iterative approach can be time-consuming and inefficient.
A popular wrapper method is the forward/backward feature elimination \cite{hastie2009elements}.
Embedded methods refer to learning algorithms that have feature selection incorporated. 
Embedded methods are optimal and time-efficient because they use the target learning algorithm to select the features. 
Some of the embedded methods are linear models such as the lasso (least absolute shrinkage and selection operator) regression \cite{tibshirani2011regression}, and tree-based methods such as the random forest (RF) \cite{breiman2001random} and gradient boosting (GB) \cite{friedman2001greedy}. 
The aforementioned models are inherently easier to interpret and have become prevalent tools across practitioners.
Recently, deep feature selection (DFS) \cite{li2016deep} and the approach proposed in \cite{wojtas2020feature} highlight the important features of NN architectures.
An emerging new kind of feature selection methods are the post-hoc explanators such as SHAP (Shapley additive explanations) \cite{lundberg2017unified} and LIME (local interpretable model-agnostic explanations) \cite{ribeiro2016should}.
They are applied after the model has made its predictions and provide insights into why the model has made a certain decision. 

In conjunction with XAI, models must proficiently extract and recognize essential features from the data, serving as effective feature extractors to ensure accurate predictions.
One approach to achieving this is through the discovery of underlying factors of variation that generated the data, which may live in a subspace of lower dimensionality than the input space. 
As an example, the success of deep learning models has also been imputed to their capability to learn and exploit latent representations through a cascade of multiple non-linear transformations of the dataset  \cite{bengio2013representation}. 
From an unsupervised learning perspective, the popular principal component analysis (PCA) \cite{pearson1901liii,hotelling1933analysis} can be used to prune out irrelevant directions in the data, constructing a latent space given as a linear combination of uncorrelated factors. 
In supervised learning, the Fisher linear discriminant \cite{fisher1936use} learns a new reduced representation searching for those vectors in the latent space that best discriminates among classes.
In statistical regression, extensive literature exists on approaches aimed at identifying a low-dimensional subspace within the input feature space that adequately captures the statistical relationship between input features and responses such as sufficient dimension reduction (SDR) \cite{adragni2009sufficient, cook2009regression, cook1994interpretation} and effective dimension reduction (EDR) \cite{li1991sliced}. 
These methodologies are closely related to the concept of active subspace. 

The active subspace method (ASM) \cite{constantine2014active}, can be used to discover directions of maximum variability of a particular function by applying a PCA to a dataset composed of its gradients. 
By eliminating directions defined by the eigenvectors associated with zero eigenvalues one can provide a reduced representation (i.e. the active subspace) where most of the original function variability is preserved.
The ASM has shown to be relevant in many areas of science and engineering such as in hydrology \cite{jefferson2015active}, shape optimization \cite{lukaczyk2014active}, and disease modeling \cite{loudon2016mathematical}. It enables efficient exploration of the model input space and can help reduce computational costs associated with sensitivity analysis, uncertainty quantification, and optimization. 

\section{Main contribution}
Our main contribution is to enhance the interpretability of the GRBFNN model while maintaining its attractive predictive performance. More in particular:
\begin{description}
    \item[$\bullet$ \textbf{Supervised Dimensionality Reduction in Active Subspaces}:] 
    We enhance the GRBFNN by incorporating a learnable precision matrix. After completing the model training, valuable latent information about the prediction task is extracted by analyzing the precision matrix spectrum. The eigenvalues of the precision matrix offer insights regarding the curvature of the Gaussian basis function at the GRBFNN centers along the corresponding eigenvectors. Dominant eigenvalues correspond to eigenvectors explaining a substantial portion of the variability within the GRBFNN model. Consequently, analyzing eigenvalues and eigenvectors respectively helps us understand the degrees of freedom and directions where the model exhibits the most variability. 
    As a result, our proposed model can be employed for supervised dimensionality reduction, such as projecting the learned model into a 2-dimensional active subspace for visualization and analysis but also for optimization purposes \cite{lukaczyk2014active, li2019surrogate}. 
    \item[$\bullet$ \textbf{Feature Importance Ranking Estimation:}] Simultaneously, for enhanced transparency and interpretability, we estimate the feature importance ranking of the learning task, enabling the use of our model for feature selection purposes. This is facilitated by recognizing that the eigenvectors also serve as the Jacobian of the linear transformation in a new coordinate system defined within the active/latent space thus enabling the assessment of the importance of the input feature to the GRBFNN model output.
    \item[$\bullet$ \textbf{Impact of Regularization}:] To improve the smoothness and the generalization capability of our model, we introduce two regularization parameters: one for the weights and the other one for the elements of the precision matrix of the Gaussian basis function. To better analyze the behavior of our model, we investigate the synergy between them. Interestingly, numerical results suggest that a stronger role is played by the regularizer of the precision matrix rather than the one that controls the magnitude of the weights.
\end{description}
In conclusion, we perform numerical experiments to compare the GRBFNN model against other well-established machine learning models, including support vector machines (SVMs) \cite{cortes1995support}, Random Forest (RF) \cite{breiman2001random}, multilayer perception (MLP) \cite{rosenblatt1958perceptron}, extreme gradient boosting (XGB) \cite{Chen_2016}, as well as state-of-the-art deep learning feature selection embedding methods presented in \cite{li2016deep, wojtas2020feature}. Additionally, we assess our model against a recent transformer architecture specifically designed for tabular data, known as the FT-Transformer (FT-T) \cite{gorishniy2021revisiting}. The outcomes demonstrate that our model not only attains competitive prediction performances but also furnishes meaningful feature importance ranking and interpretable insights that potentially can assist decision-making in real-world applications.

\section{Model Description}
Radial basis functions have been introduced for solving interpolation problems, which consist of building the following interpolant
\begin{equation}\label{eq:rbf_f}
    f(\mathbf{x}) = \sum_{m=1}^{M} w_m \varphi(||\mathbf{x} - \mathbf{x}_m||)
\end{equation}
where we have $M$ weights $w_m \in \mathbb{R}$, a continuous function $\varphi : \mathbb{R}^{+} \rightarrow \mathbb{R}$ which represents the basis function, and the centers $\mathbf{x}_m$. One can solve the interpolation problem by solving the following linear system by imposing the interpolation condition

\begin{equation}
    \boldsymbol{\Phi} \mathbf{w} = \mathbf{y} 
\end{equation}
where the $N\times M$ (in this case with $M=N$) symmetric matrix $\boldsymbol{\Phi}$ has elements $\boldsymbol{\Phi}_{nm} = \varphi(||\mathbf{x}_n - \mathbf{x}_m||)$, with $\mathbf{w} = (w_1, \dots, w_M)$, the response or target variable vector $\mathbf{y} = (y_1, \dots, y_N)$ and the $n$th data point $\mathbf{x}_n$ belonging to the $N \times D$ data matrix $\mathbf{X}$. It has been proven \cite{micchelli1984interpolation} that for some RBF (eg. the Gaussian) the matrix $\boldsymbol{\Phi}$ is not singular if all the data points are distinct with $N>2$. 

Our first modification to the model in Eq. \ref{eq:rbf_f} concerns the kernel. 
We are primarily interested in learning and exploiting hidden correlation structures in the dataset so that we can equip our RBF model with a Gaussian basis function with a symmetric positive definite matrix as follows
\begin{equation}\label{eq:md_kernel}
	\begin{split}
			\varphi(||\mathbf{x} - \mathbf{x}_j||) =
			\exp\left\{-\frac{1}{2}(\mathbf{x}-\mathbf{x}_j)^T \mathbf{P} (\mathbf{x}-\mathbf{x}_j)\right\} \\ =\exp\left\{-\frac{1}{2}(\mathbf{x}-\mathbf{x}_j)^T \mathbf{U}^T\mathbf{U} (\mathbf{x}-\mathbf{x}_j)\right\}
		\end{split}	
\end{equation}
the matrix $\mathbf{P}$ is a $D \times D$ symmetric and positive definite precision matrix that can be expressed as upper triangular matrix multiplication using $\mathbf{U}$. 

The function approximation problem, in this case, can be solved by minimizing the following nonconvex optimization problem and defining the vector $\mathbf{u} = \vech(\mathbf{U})$, where the operator $\vech$ is the half vectorization of matrices which means that the upper triangular entries of the matrix $\mathbf{U}$ are collected inside the vector $\mathbf{u}$
\begin{equation}\label{eq:min_rbf}
	\begin{aligned}
		\min_{\mathbf{w}, \mathbf{u}} & \quad E(\mathbf{w}, \mathbf{u})\\
	\end{aligned}
\end{equation}
and the error function in the regression case takes the following form
\begin{equation}\label{eq:gbf}
		E(\mathbf{w}, \mathbf{u}) = \frac{1}{2}\sum_{n=1}^{N} (y_n - f(\mathbf{x}_n))^{2} =  \frac{1}{2} \sum_{n=1}^{N} \left(y_n - \sum_{m=1}^{M} w_m \exp\left\{-\frac{1}{2}(\mathbf{x}_n-\mathbf{x}_m)^T \mathbf{U}^T\mathbf{U} (\mathbf{x}_n-\mathbf{x}_m)\right\}\right)^2 
\end{equation}

The number of parameters to optimize in this case is $P = M + D + \frac{D \times (D - 1)}{2}$. 
From numerical experiments, the model $f$ defined in Eq. \ref{eq:rbf_f} can produce a very sharply peaked function at the end of the minimization of the error function defined in Eq. \ref{eq:gbf}. 
In such cases,  we encountered large values in the entries of the precision matrix $\mathbf{P}$. 
Then, it is natural to force the smoothness of $f$ through regularization.  
The measure of the bumpiness of the function $f$ is controlled by the second derivative of the function $f$ that depends on both the weights $\mathbf{w}$ and the precision matrix $\mathbf{P}$. 
Consequently, the regularizers have the responsibility to force the Gaussian kernel to be as flat as possible, penalizing large values of the entries of the matrix $\mathbf{P}$ along with the weights $\mathbf{w}$ and promoting the smoothness of $f$.
After the considerations above, the regularized error function becomes 
\begin{equation}\label{eq:reg_rbf}
R(\mathbf{w}, \mathbf{u}) = E(\mathbf{w}, \mathbf{u}) + G(\mathbf{w}, \mathbf{u})
\end{equation}
where the penalty function is given by 
\begin{equation}
   G(\mathbf{w}, \mathbf{u}) = \frac{1}{2}\lambda_{\mathbf{u}}||\mathbf{u}||^{2} +  \frac{1}{2}\lambda_{\mathbf{w}}||\mathbf{w}||^{2} 
\end{equation}
The regularization parameter $\lambda_{\mathbf{u}}$ influences the precision matrix $\mathbf{P}$, promoting the flatness of the Gaussian kernel in Eq. \ref{eq:md_kernel}. Meanwhile, $\lambda_{\mathbf{w}}$ penalizes large weights.
Then we solve the following nonconvex optimization problem 
\begin{equation}\label{eq:min_reg_rbf}
	\begin{aligned}
		\min_{\mathbf{w}, \mathbf{u}} & \quad R(\mathbf{w}, \mathbf{u})
	\end{aligned}
\end{equation}
with the partial gradients respect to $\mathbf{w}$ and $\mathbf{u}$ given as follows
\begin{gather}
	\nabla R({\mathbf{w}}) = \boldsymbol{\Phi}^T (\mathbf{y} - \boldsymbol{\Phi} \mathbf{w}) + \lambda_{\mathbf{w}} \mathbf{w} = \boldsymbol{\Phi}^{T} \mathbf{r} + \lambda_{\mathbf{w}} \mathbf{w} \label{eq:partial_w} \\
	\nabla R({\mathbf{u}}) = \vech\left(\sum_{n=1}^{N} r_n \sum_{m=1}^{M} w_m \mathbf{G}_{nm} \boldsymbol{\Phi}_{nm}\right) + \lambda_{\mathbf{u}}\mathbf{u} \label{eq:partial_u}
\end{gather}
where the $D \times D$ matrix $\mathbf{G}_{nm}$ defined as $\mathbf{G}_{nm} = (\mathbf{x}_n - \mathbf{x}_m)(\mathbf{x}_n - \mathbf{x}_m)^{T}\mathbf{U}$ and $r_n$ the $n$th component of the vector $\mathbf{r} = \mathbf{y} - \boldsymbol{\Phi} \mathbf{w}$.

Until now we assumed the centers are exactly given by our training dataset. 
This might be unfeasible and computationally very expensive for very large $N$. 
This issue can be easily solved by selecting the number of the $M$ centers collected in the $M \times D$ matrix $\mathbf{C}$ to be less than the number of data points $N$ as shown in \cite{broomhead1988radial}. Depending on how the centers are selected we can distinguish two different strategies:
\begin{enumerate}
    \item Unsupervised selection of the centers: in this case, one can choose an $M$ centers $\mathbf{c}_m$ at random among the data points or by running a clustering algorithm (e.g. $k$-means \cite{lloyd1982-IEEE}). Given the centers, the objective function is the same as in Eq.~\ref{eq:reg_rbf} except that now $M<N$
    \begin{equation}\label{eq:min_rbf_uns}
    	\begin{aligned}
    		\min_{\mathbf{w}, \mathbf{u}} & \quad R(\mathbf{w}, \mathbf{u})\\
    	\end{aligned}
    \end{equation}
    with
\begin{equation}
    R(\mathbf{w}, \mathbf{u})= \frac{1}{2} \sum_{n=1}^{N} \left(y_n - \sum_{m=1}^{M} w_m \exp\left\{-\frac{1}{2}(\mathbf{x}_n-\mathbf{c}_m)^T \mathbf{U}^T\mathbf{U} (\mathbf{x}_n-\mathbf{c}_m)\right\}\right)^2 + G(\mathbf{w}, \mathbf{u})
\end{equation}

The partial gradients are the same as in Eq. \ref{eq:partial_w} and in Eq. \ref{eq:partial_u}
together with the partial gradient respect to $\mathbf{u}$ in Eq. \ref{eq:partial_u} unchanged, together with the total number of parameters $P$. 
    \item Supervised selection of the centers: in this case the centers are considered learnable, adding $D\times M$ parameters in the optimization problem. With this variation, the model has the following form with recasting the matrix containing the centers as a vector $\mathbf{c} = \vect(\mathbf{C})$
    \begin{equation}\label{eq:min_rbf_sup}
    	\begin{aligned}
    		\min_{\mathbf{w}, \mathbf{u}, \mathbf{c}} & \quad R(\mathbf{w}, \mathbf{u}, \mathbf{c})\\
    	\end{aligned}
    \end{equation}
    with
\begin{equation}
    R(\mathbf{w}, \mathbf{u}, \mathbf{c})= \frac{1}{2} \sum_{n=1}^{N} \left(y_n - \sum_{m=1}^{M} w_m \exp\left\{-\frac{1}{2}(\mathbf{x}_n-\mathbf{c}_m)^T \mathbf{U}^T\mathbf{U} (\mathbf{x}_n-\mathbf{c}_m)\right\}\right)^2 + G(\mathbf{w}, \mathbf{u}, \mathbf{c})
\end{equation}
        The partial gradient with respect to the $m${th} center is the following
        \begin{equation}
            \nabla R({\mathbf{c}_m}) = \sum_{n=1}^{N} r_n  \mathbf{U}^T\mathbf{U}(\mathbf{x}_n - \mathbf{c}_m)\boldsymbol{\Phi}_{nm} + \lambda_{\mathbf{c}} \mathbf{c}_m \label{eq:partial_c}
        \end{equation}
    together with the partial gradients in Eq. \ref{eq:partial_w} and Eq. \ref{eq:partial_u} and $r_n$ the $n$th component of the vector $\mathbf{r} = \mathbf{y} - \boldsymbol{\Phi} \mathbf{w}$. 
    The number of parameters is in this case $P = M \times D + M + D + \frac{D \times (D - 1)}{2}$. 
    Where in the penalty function we introduced the possibility to regularize the position of the centers, controlled by $\lambda_{\mathbf{c}}$ as follows  $G(\mathbf{w}, \mathbf{u}, \mathbf{c}) = \frac{1}{2}\lambda_{\mathbf{u}}||\mathbf{u}||^{2} +  \frac{1}{2}\lambda_{\mathbf{w}}||\mathbf{w}||^{2} + \frac{1}{2}\lambda_{\mathbf{c}}||\mathbf{c}||^{2}$.
\end{enumerate}

%
\subsection{Extracting Insights from the GRBFNN: Feature Importance and Active Subspace}\label{sec:as_fs}
After obtaining the parameters of the GRBFNN, we can extract valuable information from the spectrum of the matrix $\mathbf{P}$.
Specifically, we aim to determine whether the variability of the fitted model $f$ is restricted to a lower-dimensional space compared to the original space, as well as to identify the directions in which the function $f$ is most sensitive. This allows us to establish the active subspace.
It is easy to observe that the exponent of Eq. \ref{eq:md_kernel} is the following quadratic form also known as the squared Mahalanobis distance
\begin{equation}\label{eq:mah_dist_x}
d_M^2(\mathbf{x}) = (\mathbf{x}-\mathbf{x}_j)^T \mathbf{P} (\mathbf{x}-\mathbf{x}_j)
\end{equation} 
which expresses the functional dependence of the Gaussian kernel on the input variable $\mathbf{x}$.
More insights can be revealed by expanding Eq. \ref{eq:mah_dist_x} in terms of eigenvectors and eigenvalues
\begin{equation}\label{eq:mah_dist_x_z}
d_M^2(\mathbf{x}) = (\mathbf{x}-\mathbf{x}_j)^T \mathbf{P} (\mathbf{x}-\mathbf{x}_j) =  (\mathbf{x}-\mathbf{x}_j)^T \mathbf{V}\boldsymbol{\Gamma}\mathbf{V}^T (\mathbf{x}-\mathbf{x}_j) =  (\mathbf{z}-\mathbf{z}_j)^T \boldsymbol{\Gamma} (\mathbf{z}-\mathbf{z}_j) 
\end{equation} 
shows that the second derivatives of Eq. \ref{eq:mah_dist_x_z} are represented by the eigenvalues in the diagonal matrix $\boldsymbol{\Gamma}$, after a rotation in the latent space $\mathbf{z} \in \mathbb{Z} \subset \mathbb{R}^K$ (with $K=D$) under the new basis defined by the eigenvectors in the ($D \times K$) matrix $\mathbf{V}$. 

The spectrum of the precision matrix $\mathbf{P}$ highlights the principal curvatures \cite{guggenheimer2012differential} of Eq. \ref{eq:mah_dist_x_z}.
The presence of zero eigenvalues indicates that the factors of variation in $f$ are manifested in a lower-dimensional subspace than the input dimensionality $D$. Additionally, the eigenvector $\mathbf{v}_k$ corresponding to the largest eigenvalue $\gamma_k$ identifies the direction of maximum curvature of the quadratic function in Eq. \ref{eq:mah_dist_x}, thereby pinpointing the direction in which $f$ is most globally sensitive.
Furthermore, the Gaussian kernel in Eq. \ref{eq:md_kernel} in the latent space $\mathbb{Z}$ is given by a product of $D$ independent contributions
\begin{equation}\label{eq:md_kernel_ind}
	\begin{split}
			\varphi(||\mathbf{x} - \mathbf{x}_j||) =
			\exp\left\{-\frac{1}{2}(\mathbf{x}-\mathbf{x}_j)^T \mathbf{P} (\mathbf{x}-\mathbf{x}_j)\right\} \\ =\exp\left\{-\frac{1}{2} \sum_{d=1}^{D} \gamma_d (z_d - z_{jd})\right\} \\ = \prod_{D=1}^{D}\exp\left\{-\frac{1}{2} \gamma_d (z_d - z_{jd})\right\}
		\end{split}	
\end{equation}
enhancing the fact that the variability of the model $f$ is axis aligned within the latent space.

To identify which input variables $x_d$ are more critical in the prediction task of our model, we can observe that the matrix $\mathbf{V}$, which contains the eigenvectors of the matrix $\mathbf{P}$, represents the Jacobian of the linear transformation that maps the input space to the latent space, as demonstrated in Eq. \ref{eq:mah_dist_x_z}.
Considering the original input vector $\mathbf{x}$ as generated from a linear combination of latent variables $\mathbf{z}$ and the eigenvectors $\mathbf{V}$
\begin{equation}
    \mathbf{x} = \mathbf{V}\mathbf{z}
\end{equation}
they represent simply the following derivative $\frac{\partial \mathbf{x}}{\partial \mathbf{z}} = \mathbf{V}$
showing that the $k$th eigenvector $\mathbf{v}_k$ can be interpret as the contribution of the $k$th latent variable $z_k$ in the variation of $\mathbf{x}$. 

Each element of the matrix $\mathbf{V}$ has to be transformed in its absolute value to obtain meaningful results. So we can define the matrix $\bar{\mathbf{V}}$ where each component is given by $\bar{v}_{dk} = |v_{dk}|$.
To obtain the feature importance ranking vector, we need to scale the eigenvectors $\mathbf{v}_k$ by their corresponding eigenvalues $\gamma_k$ as the eigenvectors are returned typically normalized to the unitary norm from numerical procedures. 
This scaling ensures that more importance is given to the directions with the most significant variation. 
The resulting $D$-dimensional feature importance ranking vector can be defined as follows
\begin{equation}\label{eq:feature_importance}
    \text{Feature Importance} = \sum_{k=1}^{K}\gamma_k\bar{\mathbf{v}}_k
\end{equation}
A final normalization step is performed so the feature importance vector ranges between zero and one.

\subsection{Numerical Examples}
In this section, we want to provide some simple examples to highlight graphically the behavior of the proposed model. 
We first start with two simple classification problems with $N=100$ and $D=2$. 
In the first problem there two classes $c_1$ and $c_2$ that are normally distributed with mean $\boldsymbol{\mu}_{c_1}^T= [1, 1 ]^T$ and $\boldsymbol{\mu}_{c_2}^T = [2.8, 2.8]^T$, respectively, and same covariance matrix $\boldsymbol{\Sigma} = \bigl[ \begin{smallmatrix}0.81 & 0.72\\ 0.72 & 0.66 \end{smallmatrix}\bigr]$. 

\begin{figure}[!htb]
\centering
    \includegraphics[scale=0.4]{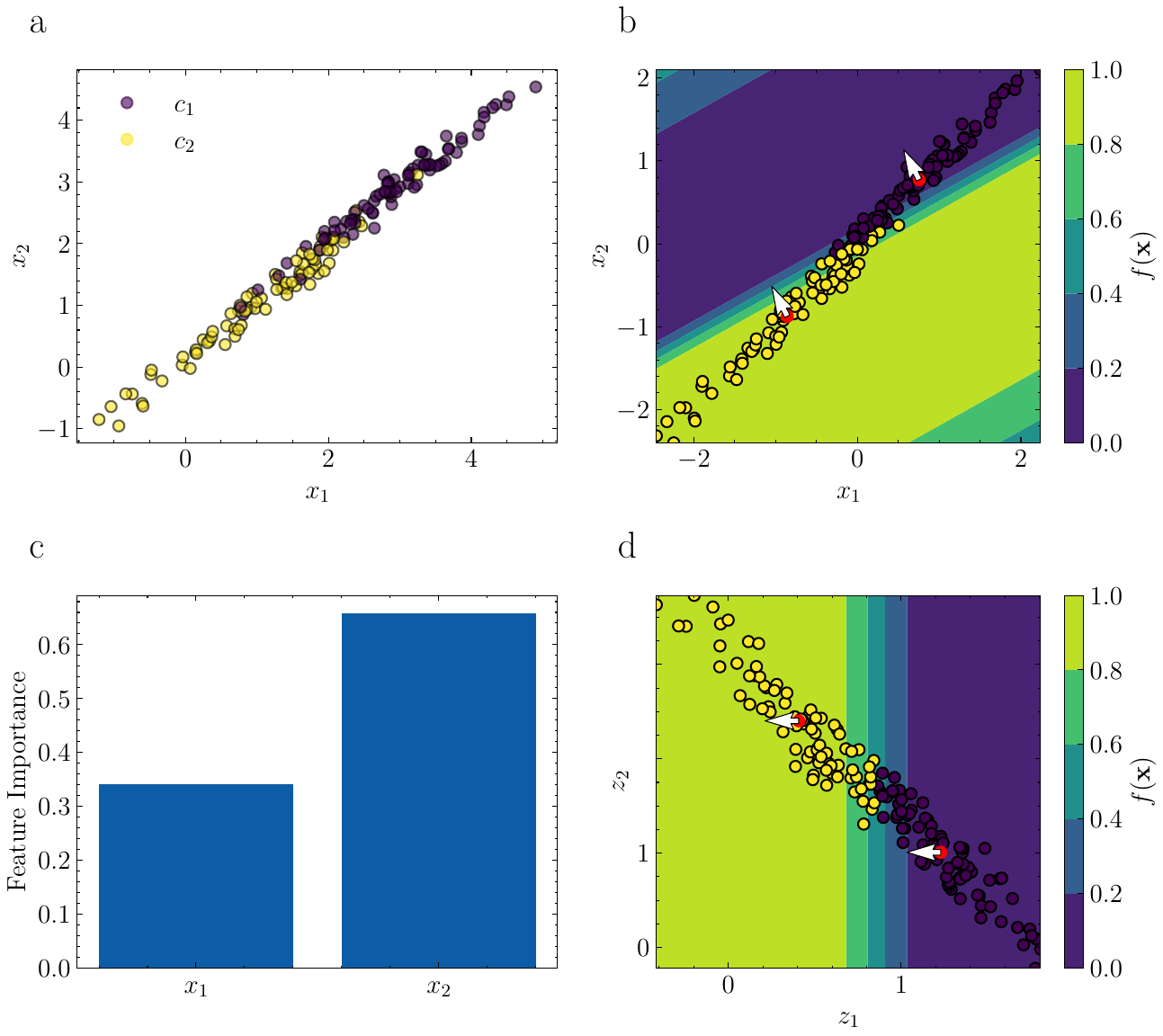}
    \caption{The GRBFNN behavior is graphically represented in four subfigures: (a) shows the classification problem with purple and yellow dots representing the two classes. The subfigure (b) shows the fitted GRBFNN in the input space, while the (c) figure shows the fitted GRBFNN model in the active subspace. Contour levels show estimated class probabilities.
    The red dotted points represent the GRBFNN centers. The white arrow highlights the direction of the dominant eigenvector $\mathbf{v}_1$.
    Finally, in (c) the subfigure shows the feature importance estimated from the GRBFNN.}
    \label{fig:c1_problem_2}
\end{figure}

The scatter plot of the two classes is shown in Fig. \ref{fig:c1_problem_2}a where the yellow and purple dotted points represent class $c_1$ and $c_2$, respectively. In Fig. \ref{fig:c1_problem_2}b, we display the fitted GRBFNN model with $M=2$ centers (highlighted by the red points) obtained with the unsupervised selection strategy by $k$-means clustering. 
Furthermore, we plot the eigenvector $\mathbf{v}_1$ corresponding to the dominant eigenvalue $\gamma_1$ of the matrix $\mathbf{P}$ as white arrows with the origin at the two centers. 
This shows that the fitted model $f$ obtains most of its variability along the direction of $\mathbf{v}_1$, which is orthogonal in this case to the contour levels of $f$. 

In Fig. \ref{fig:c1_problem_2}d, we show the fitted model in the latent space obtained by projecting the dataset $\mathbf{X}$ to the new basis defined by the eigenvectors of $\mathbf{P}$, defining the projected dataset $\mathbf{Z} = \mathbf{X}\mathbf{V}$. We observe that all the variation of $f$ is aligned to the first latent variable $z_1$, which indicates that the fraction $\frac{\gamma_1}{\sum_k^{K}\gamma_k}$ is approximately equal to 1. Fig. \ref{fig:c1_problem_2}c shows the feature importance estimated by our model using Eq. \ref{eq:feature_importance}, which validates that the input feature $x_2$ plays a more significant role in the discrimination power between the classes than $x_1$.

\begin{figure}[!htb]
\centering
    \includegraphics[scale=0.45]{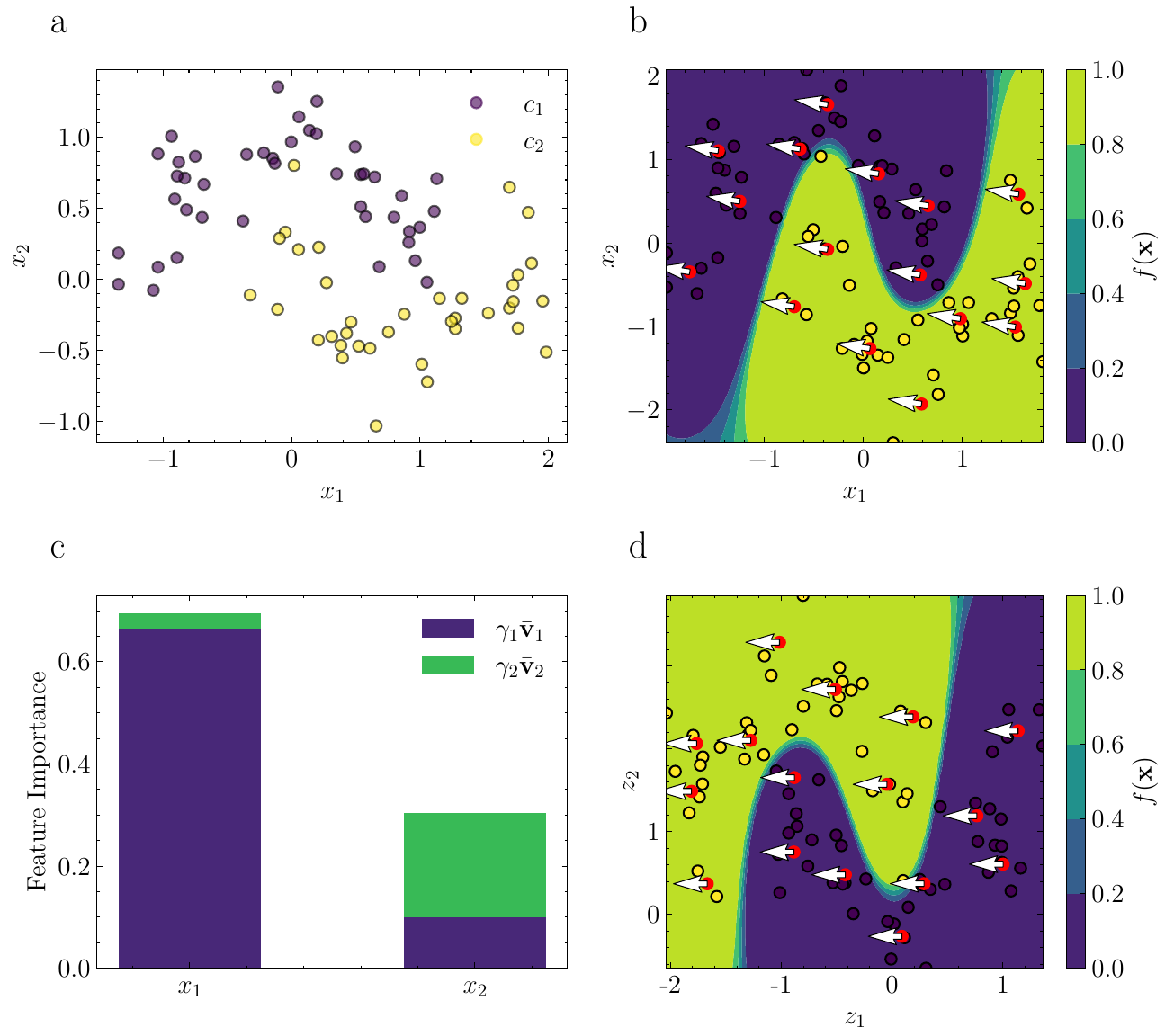}
    \caption{The GRBFNN behavior is graphically represented in four subfigures: (a) shows the classification problem with purple and yellow dots representing the two classes. The subfigure (b) shows the fitted GRBFNN in the input space, while the (c) figure shows the fitted GRBFNN model in the active subspace. Contour levels show estimated class probabilities.
    The red dotted points represent the GRBFNN centers. The white arrow highlights the direction of the dominant eigenvector $\mathbf{v}_1$.
    Finally, in (c) the subfigure shows the feature importance estimated from the GRBFNN and highlights its composition.}
    \label{fig:c1_problem_4}
\end{figure}

Another example of a classification problem is shown in Fig. \ref{fig:c1_problem_4}. In this case, we have two noisy interleaving half circles with $N=100$ and $D=2$, as seen in Fig. \ref{fig:c1_problem_4}a. 
To achieve a stronger discriminative power from the model, we choose $M=16$ centers.

In contrast to the previous example, not all of the model $f$ variability is concentrated along the direction identified by the eigenvector $\mathbf{v}_1$ related to the dominant eigenvalue $\gamma_1$. 
In this case, the fraction $\frac{\gamma_1}{\sum_k^{K}\gamma_k}$ is approximately 0.8, meaning that the resulting feature importance in Eq. \ref{eq:feature_importance} includes the contribution of the eigenvector $\mathbf{v}_2$. This is highlighted in the barplot in Fig \ref{fig:c1_problem_4}d, where we decomposed the feature importance showing the contribution of each term in Eq. \ref{eq:feature_importance}.
Since $\mathbf{v}_2$ is orthogonal to $\mathbf{v}_1$, it gives more importance to the feature $x_2$ because $\mathbf{v}_1$ is quasi-parallel to the input $x_1$.

\begin{figure}[!htb]
\centering
    \includegraphics[scale=0.45]{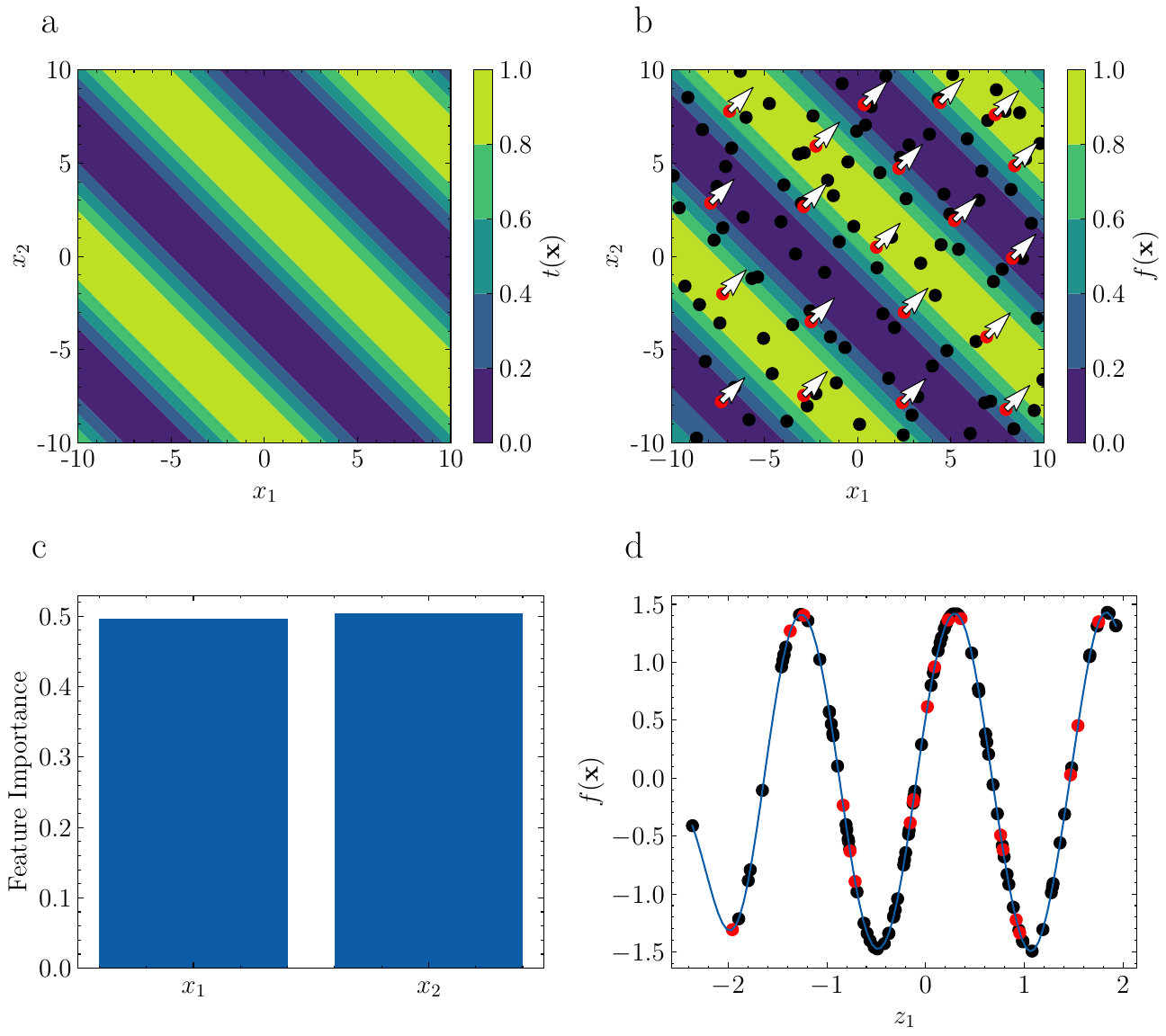}
    \caption{The GRBFNN behavior is depicted in four subfigures: (a) shows the regression problem with target function $t(\mathbf{x}) = \sin(0.5x_1 + 0.5x_2)$, while (b) displays the fitted GRBFNN in the input space. 
    The dominant eigenvector $\mathbf{v}_1$ is indicated by a white arrow, and the GRBFNN centers are shown as red dotted points. The subfigure (d) shows the fitted GRBFNN model projected in the one-dimensional active subspace. The function values at the input data and at the centers are represented by black and red dotted points, respectively. Finally, in (c) the subfigure displays the feature importance estimated from the GRBFNN. Function values are normalized.}
    \label{fig:reg_problem_1}
\end{figure}

We present an example of regression, where the function to be approximated is $t(\mathbf{x}) = \sin(ax_1 + bx_2)$, with $a$ and $b$ being real scalars. 
Fig. \ref{fig:reg_problem_1} shows the case where $a$ and $b$ are equal to 0.5, and the true function is depicted in Fig. \ref{fig:reg_problem_1}a. We then use our proposed model to obtain an approximation, as shown in Fig. \ref{fig:reg_problem_1}b along the direction given by the eigenvector $\mathbf{v}_1$, with the centers represented by dotted red points.

\begin{figure}[!htb]
\centering
    \includegraphics[scale=0.45]{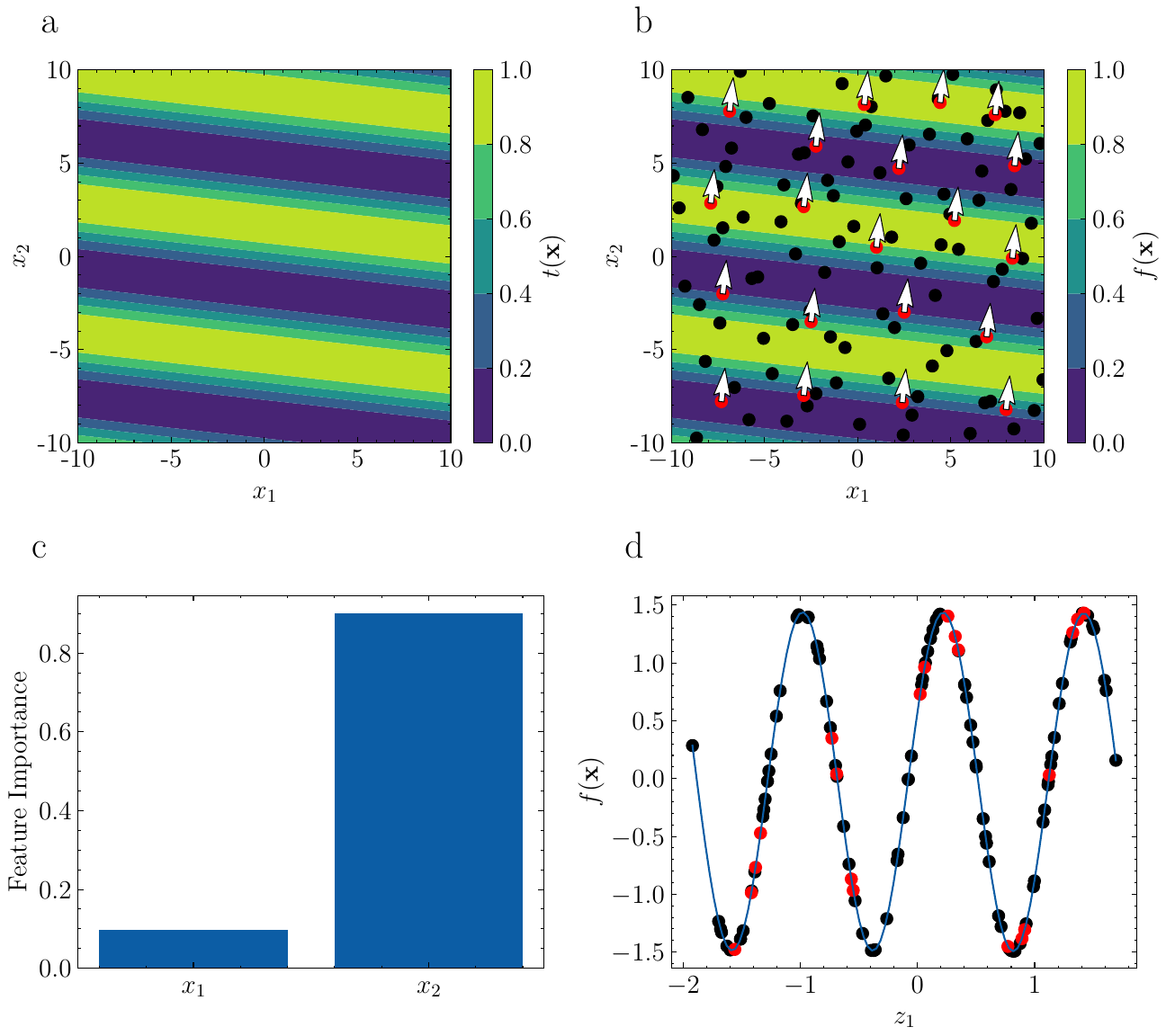}
    \caption{The GRBFNN behavior is depicted in four subfigures: (a) shows the regression problem with target function $t(\mathbf{x}) = \sin(0.1x_1 + 0.9x_2)$, while (b) displays the fitted GRBFNN in the input space. 
    The dominant eigenvector $\mathbf{v}_1$ is indicated by a white arrow, and the GRBFNN centers are shown as red dotted points. The subfigure (d) shows the fitted GRBFNN model projected in the one-dimensional active subspace. The function values at the input data and at the centers are represented by black and red dotted points, respectively. Finally, in (c) the subfigure displays the feature importance estimated from the GRBFNN. Function values are normalized.}
    \label{fig:reg_problem_2}
\end{figure}

Furthermore, we perform a supervised dimensionality reduction from the original two-dimensional space to the one-dimensional subspace defined by the first eigenvector $\mathbf{v}_1$. This subspace captures the 'active' part of the function where most of the variation is realized, as illustrated in Fig. \ref{fig:reg_problem_1}d. Finally, we estimate the feature importance using Eq. \ref{eq:feature_importance} and find that $x_1$ and $x_2$ contribute equally, as expected. This result is depicted in Fig. \ref{fig:reg_problem_1}c.
In this final example, we altered the values of the scalars $a$ and $b$ to 0.1 and 0.9, respectively. This modification resulted in a change in the feature importance estimated by our model, as depicted in Fig. \ref{fig:reg_problem_2}.
 
In summary, the GRBFNN model beyond solving a classical regression/classification model provides the user with valuable information about the model behavior such as allowing visualization of the fitted model $f$ of the active subspace thereby recognizing the underlying factors of variation of the data, and in parallel allowing to discover which are the most important input features related to the learned model about the prediction task.

\section{Numerical Experiments}
This section aims to provide a comprehensive evaluation of the proposed model by assessing its predictive performance along with the feature selection and feature importance ranking quality. 
We consider two variants of the same GRBFNN model, one with unsupervised center selection (GRBFNN$_k$) as given in Eq. \ref{eq:min_rbf_uns} and the other one with supervised center selection (GRBFNN$_c$) as given in Eq. \ref{eq:min_rbf_sup} for comparison purposes. 

We compare the performance of these models with other popular models such as multi-layer perceptron (MLP) \cite{rosenblatt1958perceptron} and support vector machines (SVMs) \cite{cortes1995support}.
As the GRBFNN model incorporates feature selection, it can be classified as an embedding method. To provide a comprehensive benchmark, we also include other widely used embedding methods such as random forest (RF) \cite{friedman1991multivariate} and extreme gradient boosting (XGB) \cite{Chen_2016}, which have shown strong performance for tabular data.

We include state-of-the-art embedding deep learning methods such as deep feature selection (DFS) \cite{li2016deep} and the method proposed in \cite{wojtas2020feature}, referred to as FIDL (feature importance for deep learning) in this comparison for simplicity. We also consider in this benchmark a recent transformer model designed for tabular data, known as the FT-Transformer (FT-T) \cite{gorishniy2021revisiting}.

We perform a 5-fold cross-validation to identify the best set of hyperparameters for the models. Once the best set of hyperparameters is determined, we conduct another 5-fold cross-validation using 20 different seeds. We performed a Wilcoxon signed-rank test \cite{wilcoxon1992individual} at $\alpha = 0.05$ significance level to assess the statistical significance of the numerical results.
For the FIDL model, we were able to run the cross-validation procedure to find its best set of hyperparameters varying only one random seed due to the severe time and memory complexity of the model.
For the regression problems we use a root mean squared error (RMSE) while for the classification problems, we use accuracy as a metric to evaluate the models. 

\subsection{Datasets}\label{sec:datasets}
To test the predictive performance of our model we consider 20 different real-world problems as summarized in Tab. \ref{tab:datasets}. 
We have a total of 6 binary classifications, 4 multiclass, 1 time series, and 9 regression problems. 
A detailed description of the datasets can be found in Section \ref{sec:supplementary_data} of supplementary material.
\begin{table*}[htb!]
	\centering
    \begin{threeparttable}
	\caption{Datasets considered in the benchmark.}
	\begin{tabular}{lccccc}\toprule    
    Name &  $N$ & $D$ & Task & Reference  \\ 
    \hline
   Digits & 357 & 64 & Binary classification & \cite{alimoglu1996methods} \\
   Iris & 150 & 4 & Multiclass classification & \cite{fisher1936use} \\
   Breast Cancer & 569 & 30 & Binary classification & \cite{street1993nuclear} \\
   Wine & 173 & 13 & Multiclass classification & \cite{aeberhard1994comparative}\\
   Australian & 600 & 15 & Binary classification & \cite{Dua:2019}\\
   Credit-g & 1000 & 20 & Binary classification & \cite{Dua:2019}\\
   Glass & 214 & 9 & Multiclass classification & \cite{evett:1987:rifs}\\
   Blood & 748 & 4 & Binary classification &\cite{YEH20095866}\\
   Heart Disease & 270 & 13 & Binary classification & \cite{Dua:2019}\\
   Vowel & 990 & 12 & Multiclass classification &  \cite{deterding1989speaker}\\
   Delhi Weather & 1461 & 7 & Time series & \cite{DWdataset}\\
   Boston Housing & 506 & 14 & Regression & \cite{pace1997sparse}\\
   Diabetes & 214 & 9 & Regression &\cite{smith1988using}\\
   Prostatic Cancer & 97 & 4 & Regression &\cite{stamey1989prostate}\\
   Liver & 345 & 5 & Regression & \cite{MCDERMOTT201641}\\
   Plasma & 315 & 16 & Regression & \cite{nierenberg1989determinants} \\
   Cloud & 108 & 5 & Regression  & \cite{Dua:2019} \\
   DTMB-$5415^{1}$  & 42 & 21 & Regression& \cite{d2024generative}\\
   DTMB-$5415^{2}$ & 42 & 21 & Regression& \cite{d2024generative}\\
   Body Fat & 252 & 14 & Regression & \cite{penrose1985generalized}\\
	\bottomrule
\end{tabular}
\label{tab:datasets}
\end{threeparttable}
\end{table*} 

Furthermore, we use synthetic datasets to provide a deeper comparison of the feature importance and feature selection results obtained by the methods since the ground truth of the feature importance related to the learning task is known. 
This allows for the evaluation of the quality of the feature selection and ranking provided by the methods, as the true feature importance can be compared to the estimates obtained by the models. 
The synthetic datasets considered are the following:
\begin{itemize}
\item \textbf{Binary classification} \cite{hastie2009elements} (P1): given $y=-1$, the ten input features are generated with $(x_1, \dots, x_{10}) \sim \mathcal{N}(0, \mathbf{I})$. 
Given $y=1$, $x_1$ through $x_4$ are standard normal conditioned on $9 \leq \sum_{j=1}^4 x_j^2 \leq 16$, and $(x_5, \dots, x_{10})\sim \mathcal{N}(0, \mathbf{I})$. \\
The first four features are relevant for P1.   
\item \textbf{$3$-dimensional XOR as 4-way classification} \cite{chen2017kernel} (P2): Consider the 8 corners of the 3-dimensional hypercube $(v_1, v_2, v_3) \in {-1, 1}^3$, and group them by the tuples $(v_1v_3, v_2v_3)$, leaving 4 sets
    of vectors paired with their negations ${v^{(i)}, -v^{(i)}}$. Given a class $i$, a point is generated from the mixture distribution $(1/2)\mathcal{N}(v^{(i)}, 0.5 \mathbf{I}) + (1/2)\mathcal{N}(-v^{(i)}, 0.5\mathbf{I})$. Each example additionally has 7 standard normal noise features for a total of $D=10$ dimensions. \\
    The first three features are relevant for P2.    
\item \textbf{Nonlinear regression} \cite{friedman1991multivariate} (P3): The $10$-dimensional inputs $x$ are independent features uniformly distributed on the interval $[0, 1]$. The output $y$ is created according to the formula  $y = 10  \sin(\pi x_1 x_2) + 20 (x_3 - 0.5)^2 + 10 x_4 + 5 x_5 + \epsilon $ with $\epsilon \sim \mathcal{N}(0, 1)$. \\
The first five features are relevant for P3. 
\end{itemize}  
In all those three cases we varied the number of data points with $N \in \{100, 500, 1000\}$.
\subsection{Evaluation of the Predictive Performance}

Tab. \ref{tab:num_res} presents a summary of our numerical results, showing the mean from the cross-validation procedure for each model. 
Significantly, the GRBFNN exhibits robust competitiveness when compared to other models. Specifically, the GRBFNN\textsubscript{$c$} emerged as the top-performing method, demonstrating statistically significant superiority in 4 out of 20 datasets. Notably, these datasets exclusively involve regression tasks, highlighting the particular aptitude of GRBFNN\textsubscript{$c$} in addressing such problems. 

Conversely, the GRBFNN\textsubscript{$k$} proved to be more effective in classification tasks, emerging as the top-performing method in two datasets. This observation suggests that the unsupervised selection of centers might be particularly suitable for such tasks. 
It's important to notice that the selection of the center strategy is merely a hyperparameter in the GRBFNN model. Numerical results suggest that there is no clear superiority between the strategies for selecting centers, and both approaches are equally competitive. Users might consider experimenting with both strategies, as aggregating the results between the  GRBFNN\textsubscript{$c$} and the GRBFNN\textsubscript{$k$} reveals that the GRBFNN model outperforms other models with statistically significant results in 7 out of 20 datasets.

Numerical results, confirm that standard methods such as the SVM, RF, and MLP are still very good baselines.
In some datasets, the optimizer used for the training process of FIDL did not seem to converge and the resulting performance in those datasets is not reported. 
In conclusion, the no free lunch theorem \cite{wolpert1996lack} reminds us that there is no singularly superior model applicable to all machine learning problems. The choice of method depends on the specific dataset and task requirements.



%
\begin{table*}[!htb]
	\centering
	\caption{Numerical results summary: The first ten datasets are for binary and multiclass classification, showing average accuracy achieved via cross-validation. The last ten are for regression and time series, showing average RMSE via cross-validation. The bold numbers signify the method with the best performance on the test data, demonstrating statistical significance based on the Wilcoxon test ($\alpha=0.05$). An asterisk indicates that, although the method is the overall best, statistical significance has not been established.}
\scalebox{0.495}{
	\begin{tabular}{*{21}{l}}\toprule
	\multirow{2}{*}{} &
      \multicolumn{2}{c}{GRBFNN\textsubscript{$k$}} & \multicolumn{2}{c}{GRBFNN\textsubscript{$c$}} & \multicolumn{2}{c}{SVM} & \multicolumn{2}{c}{RF} & \multicolumn{2}{c}{XGB} & \multicolumn{2}{c}{MLP}
      & \multicolumn{2}{c}{DFS} & \multicolumn{2}{c}{FIDL} & \multicolumn{2}{c}{FT-T}&\\ 
    &Training& Test& Training & Test & Training & Test & Training & Test & Training & Test& Training & Test& Training & Test &Training & Test&Training & Test\\
	\midrule
Digits  & 1.000 & \textbf{0.994}
 & 0.995 & 0.989
 & 1.000 & 0.992
 & 1.000 & 0.983
 & 1.000 & 0.983
 & 1.000 & 0.990
 & 1.000 & 0.990
 & 1.000 & 0.975
 & 1.000 & 0.975\\
Iris & 0.982 & \textbf{0.971}
  & 0.943 & 0.941
  & 0.974 & 0.958
  & 0.989 & 0.951
  & 0.980 & 0.955
  & 0.989 & 0.949
  & 0.985 & 0.959
  & 0.962 & 0.967
  & 0.998 & 0.958\\
Breast Cancer & 0.987 & $0.978^{*}$
  & 0.988 & 0.976
  & 0.988 & 0.977
  & 0.990 & 0.956
  & 1.000 & 0.965
  & 1.000 & 0.971
  & 0.985 & 0.976
  & 0.992 & 0.974
  & 0.999 & 0.970\\
Wine & 0.996 & $0.985^{*}$
  & 0.991 & 0.979
  & 0.992 & 0.983
  & 1.000 & 0.980
  & 1.000 & 0.965
  & 1.000 & 0.971
  & 1.000 & 0.984
  & 1.000 & 0.961
  & 1.000 & 0.981\\
Australian  & 0.911 & 0.861
  & 0.853 & 0.847
  & 0.884 & 0.858
  & 0.998 & 0.865
  & 1.000 & 0.848
  & 0.924 & \textbf{0.868}
  & 0.818 & 0.812
  & 0.883 & 0.851
  & 0.988 & 0.842\\
Credit-g & 0.787 & 0.760
  & 0.812 & 0.758
  & 0.841 & $0.765^{*}$
  & 1.000 & 0.760
  & 0.951 & 0.764
  & 0.874 & 0.744
  & 0.778 & 0.757
  & 0.793 & 0.747
  & 0.985 & 0.715\\
Glass & 0.912 & 0.657
  & 0.969 & 0.686
  & 0.857 & 0.685
  & 1.000 & \textbf{0.780}
  & 1.000 & 0.761
  & 0.978 & 0.714
  & 0.947 & 0.710
  & 0.699 & 0.630
  & 0.992 & 0.756\\
Blood  & 0.791 & 0.781
  & 0.794 & 0.782
  & 0.829 & 0.784
  & 0.818 & 0.790
  & 0.829 & 0.777
  & 0.801 & $0.792^{*}$
  & 0.789 & 0.777
  & 0.762 & 0.762
  & 0.836 & 0.752\\
Heart Disease  & 0.785 & 0.764
  & 0.846 & 0.814
  & 0.943 & 0.802
  & 0.916 & 0.834
  & 0.951 & 0.816
  & 0.872 & 0.842
  & 0.891 & 0.826
  & 0.878 & $0.844^{*}$
  & 0.996 & 0.788\\

 Vowel & 0.995 & 0.959
  & 0.998 & 0.969
  & 0.999 & \textbf{0.992}
  & 1.000 & 0.967
  & 1.000 & 0.920
  & 1.000 & 0.968
  & 0.995 & 0.953
  & - & -
  & 0.998 & 0.984\\
 \midrule
Delhi Weather  & 0.115 & 0.111
 & 0.098 & \textbf{0.107}
 & 0.109 & 0.108
 & 0.071 & 0.190
 & 0.063 & 0.144
 & 0.106 & 0.110
 & 0.541 & 0.371
 & - & - 
 & 0.081  & 0.120\\
Boston Housing  & 0.224 & 0.344
  & 0.217 & 0.361
  & 0.181 & 0.361
  & 0.135 & 0.364
  & 0.018 & 0.338
  & 0.211 & 0.359
  & 0.280 & 0.388
  & - & -
  & 0.163 & $0.336^{*}$\\

Diabetes  & 0.703 & 0.718
  & 0.669 & $0.706^{*}$
  & 0.680 & 0.711
  & 0.662 & 0.745
  & 0.617 & 0.739
  & 0.655 & 0.706
  & 0.661 & 0.711
  & - & -
  & 0.371 & 0.861\\
Prostatic Cancer & 0.581 & 0.646
  & 0.581 & 0.646
  & 0.570 & \textbf{0.625}
  & 0.610 & 0.715
  & 0.377 & 0.713
  & 0.595 & 0.662
  & 0.460 & 0.697
  & 0.408 & 0.644
  & 0.278 & 0.864\\
Liver & 0.834 & \textbf{0.895}
  & 0.838 & 0.902
  & 0.825 & 0.913
  & 0.770 & 0.923
  & 0.743 & 0.928
  & 0.857 & 0.911
  & 0.846 & 0.934
  & - & -
  & 0.491 & 1.113\\

 Plasma & 0.958 & 0.991
  & 0.941 & \textbf{0.982}
  & 0.937 & 0.987
  & 0.890 & 0.989
  & 0.865 & 0.999
  & 1.000 & 1.002
  & 0.917 & 1.070
  & - & -
  & 0.094 & 1.143\\

 Cloud & 0.340 & 0.451
  & 0.323 & 0.385
  & 0.344 & 0.373
  & 0.418 & 0.521
  & 0.216 & 0.438
  & 0.378 & 0.417
  & 0.615 & 0.613
  & 0.138 & $0.206^{*}$
  & 0.190 & 0.490\\

DTMB-$5415^{(1)}$ & 0.513 & 0.752
  & 0.103 & 0.783
  & 0.096 & 0.886
  & 0.413 & 1.014
  & 0.057 & 0.929
  & 0.009 & 0.816
  & 0.777 & 0.898
  & 0.899 & 0.932
  & 0.222 & $0.745^{*}$\\

 DTMB-$5415^{(2)}$ & 0.031 & 0.112
  & 0.020 & \textbf{0.071}
  & 0.088 & 0.226
  & 0.342 & 0.902
  & 0.001 & 0.794
  & 0.052 & 0.238
  & 0.048 & 0.300
  & 0.073 & 0.077
  & 0.293 & 0.575\\

Body Fat & 0.090 & 0.132
  & 0.086 & \textbf{0.127}
  & 0.153 & 0.148
  & 0.070 & 0.170
  & 0.112 & 0.185
  & 0.074 & 0.131
  & 0.113 & 0.137
  & 1.523 & 2.650
  & 0.076 & 0.142\\
	\bottomrule
\end{tabular}}
\label{tab:num_res}
\end{table*}
%
\subsection{The Impact of Regularization on the GRBFNN} 
\begin{figure}[!htb]
\centering
    \begin{subfigure}{0.4\linewidth}
    \centering
        \includegraphics[width=\linewidth]{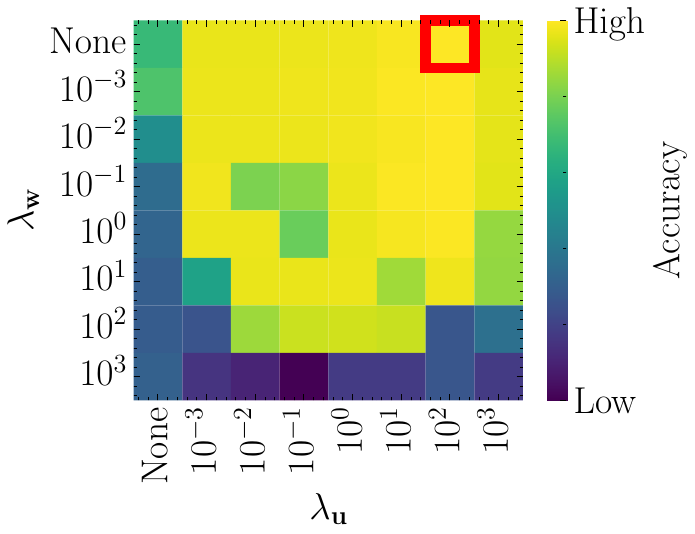}
    \caption{Breast Cancer dataset test accuracies.}
    \label{fig:reg_analysis_test_breast_cancer_GR-RBF}
    \end{subfigure}
    \begin{subfigure}{0.4\linewidth}
    \centering
        \includegraphics[width=\linewidth]{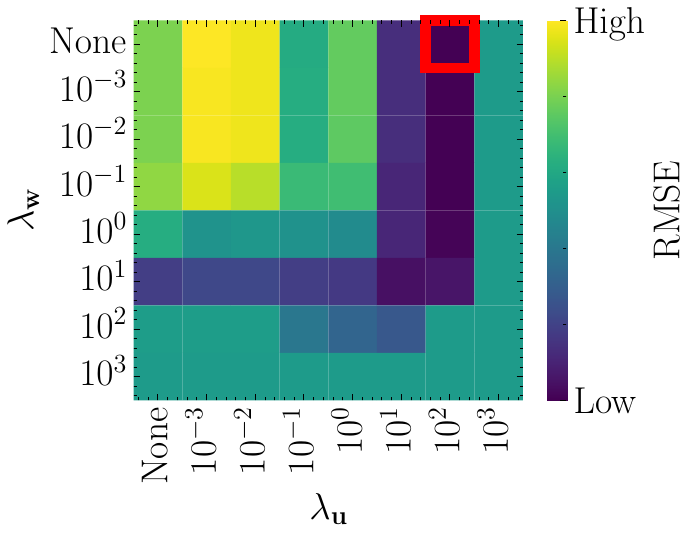}
    \caption{Prostatic cancer dataset test RMSE.}
    \label{fig:reg_analysis_test_prostate_cancer_GR-RBF}
    \end{subfigure}
\caption{Graphical interpretation of the sensitivity analysis concerning the two regularizers $\lambda_{\mathbf{w}}$ and $\lambda_{\mathbf{u}}$ on the Breast Cancer dataset (binary classification) in (a) and on the Prostatic Cancer dataset (regression) in (b). The red frame highlights the best combination of hyperparameters. A lighter color indicates higher accuracy and higher RMSE.}
\label{fig:reg_analysis_test_breast_prostatic_GR-RBF}   
\end{figure}

We aim to provide further insights into the behavior of the GRBFNN model by examining the relationship between its two regularizers, $\lambda_{\mathbf{w}}$ and $\lambda_{\mathbf{u}}$. 
We show the results of the hyperparameter search procedure on the test data for Breast Cancer and Prostatic Cancer in Fig. \ref{fig:reg_analysis_test_breast_prostatic_GR-RBF}.  
For the regression problems, dark colors indicate lower error while for classification problems lighter color indicates higher accuracy. The red frame indicates the best set of regularizers.

Interestingly, we observe that in Fig. \ref{fig:reg_analysis_test_breast_cancer_GR-RBF} and in Fig. \ref{fig:reg_analysis_test_prostate_cancer_GR-RBF} the regularizer of the precision matrix $\lambda_{\mathbf{u}}$ impacts the performance of the GRBFNN more than the regularizer of the weights $\lambda_{\mathbf{w}}$. 
This phenomenon occurs in many datasets such as Digits, Breast Cancer, Credit-g, Glass, Diabetes, Prostatic Cancer, Cloud, DTMB-$5415^{(2)}$ and the Body Fat dataset (see Fig. \ref{fig:reg_analysis} in supplementary Section \ref{sec:supplementary}). In these cases, we obtain the best combination of regularizers on the test data when $\lambda_{\mathbf{w}}$ is set to 0. 
This suggests that the regularization term $\lambda_{\mathbf{w}}$ has minimal influence on the learning task on those datasets.  

This can indicate that promoting the 'flatness' of the Gaussian basis function through penalizing the entries of the precision matrix $\mathbf{P}$ through $\lambda_{\mathbf{u}}$ may have a more pronounced regularization and generalization impact on the model than merely penalizing the amplitudes via $\lambda_{\mathbf{w}}$.
Therefore, in situations where conducting a large hyperparameter search is not feasible due to computational constraints, it might be beneficial to prioritize the hyperparameter search solely on $\lambda_{\mathbf{u}}$. 
\subsection{Supervised Dimensionality Reduction in the Active Subspace}
In Section \ref{sec:as_fs}, we mentioned that valuable insights into the behavior of the GRBFNN can be obtained by examining the eigenvalues and projecting the model onto the active subspace spanned by the eigenvectors of $\mathbf{P}$.
This enables us to visualize the function the GRBFNN is attempting to model (for example, in $2D$), offering users an additional approach to comprehending the representation learned by our model. In particular, we project the dataset $\mathbf{X}$, along the first two eigenvectors relative to the largest eigenvalues of the matrix $\mathbf{P}$, for visualization in $2D$. The function values the then computed by projecting back the latent variables in the input space.

\begin{figure}[!htb]
\centering
    \begin{subfigure}{0.35\linewidth}
    \centering
        \includegraphics[width=\linewidth]{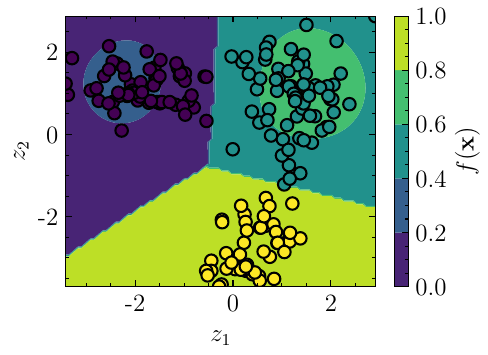}
    \caption{Wine dataset active subspace.}
    \label{fig:LR_wine_GR-RBF_2}
    \end{subfigure}
    \begin{subfigure}{0.333\linewidth}
    \centering
        \includegraphics[width=\linewidth]{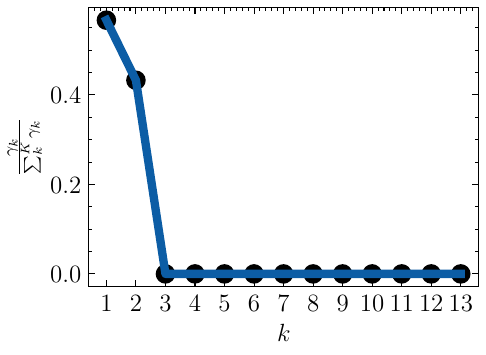}
    \caption{Wine dataset eigenvalues decay.}
    \label{fig:gammas_wine_GR-RBF_2}
    \end{subfigure}\\
\caption{Graphical interpretation of the active subspace in two dimensions (a) and corresponding eigenvalues decay (b) for the Wine dataset (multiclass classification). Function values are normalized between zero and one.}
\label{fig:wine_active}   
\end{figure}

In Fig. \ref{fig:wine_active} we illustrate the active subspace for the Wine dataset.
The eigenvalues' decay (Fig. \ref{fig:gammas_wine_GR-RBF_2}) given by $\frac{\gamma_1}{\sum_k^{K}\gamma_k}$, indicates that the GRBFNN model's variability is primarily captured in two dimensions, as the first two eigenvalues are the only ones significantly different from zero. It is worth recalling that the eigenvalue $\gamma_k$ of the matrix $\mathbf{P}$ represents the second derivative of the argument of the Gaussian basis function after rotating it in the latent space, along the corresponding principal axis $\mathbf{v}_k$.
Therefore, very low magnitude eigenvalues mean a lack of variability of our model in those principal directions, indicating that the true underlying factors of variation of the original problem may develop in a lower dimensional space.
This assertion is validated by the corresponding active subspace displayed in Fig. \ref{fig:LR_wine_GR-RBF_2}, revealing variations along both latent variables $z_1$ and $z_2$. Additionally, it is noteworthy that the learned decision boundaries appear almost linear, given the near-linear separability of classes in this problem. 

At the end of Section \ref{sec:supplementary} of the supplementary material we show in Fig. \ref{fig:eigen_embedding} the active subspace and the related eigenvalue decay for all the datasets. For the Digits, Iris, and Breast Cancer datasets, the embedding shows that the function $f$ provides a remarkable discriminative power where most of the variability is obtained in just one dimension, along with the latent variable $z_1$. 
This is confirmed by the relative eigenvalues decays, which show that the first eigenvalue $\gamma_1$ is the only one significantly different from zero.
In general, for almost linearly separable classification problems, the GRBFNN likely detects a one-dimensional active subspace as is the case for the Digits dataset.
In the regression cases, we can, for example, observe that the GRBFNN model identifies an active subspace of dimension $K=1$ in many datasets such as the Prostate Cancer, Plasma, Cloud, DTMB-$5415^{(1)}$ and the DTMB-$5415^{(2)}$ dataset. In other regression problems, the active subspace might reveal more complicated patterns as is the case of the Liver and Boston Housing datasets.

\subsection{Evaluation of the Feature Importance Ranking}
In addition to analyzing the predictive performance and performing a supervised dimensionality reduction in the active subspace for visualization, we can also obtain information about the importance of input features $\mathbf{x}$. This provides additional insights into the model behavior and enables the user to perform feature selection.
To evaluate the significance of the feature importance ranking, we will only consider the embedding methods. This means that SVM, MLP, and FT-T will not be taken into account in this part of the benchmark since they do not directly provide information on the importance of each input feature.

\subsubsection{Feature Importance Ranking for the Digits Dataset}
Some of the real-world datasets used to evaluate the predictive performance of the models can also be used to evaluate the quality of the feature importance ranking obtained once the model training has terminated such as the Digits and DTMB-$5415^{(2)}$ datasets.
We train all the models using their best set of hyperparameters from the cross-validation procedure performed precedently to the whole dataset.
\begin{figure}[!htb]
\centering
    \begin{subfigure}{0.14\linewidth}
    \centering
        \includegraphics[width=\linewidth]{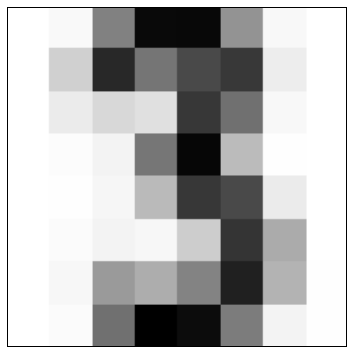}
    \caption{Mean of the class '3'.}
    \label{fig:mean_c0_digits}
    \end{subfigure}
    \begin{subfigure}{0.14\linewidth}
    \centering
        \includegraphics[width=\linewidth]{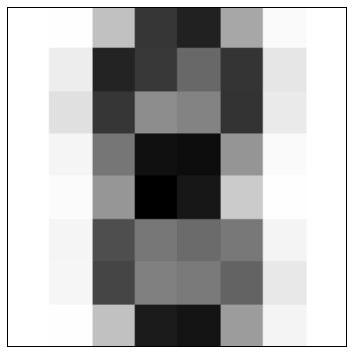}
    \caption{Mean of the class '8'.}
    \label{fig:mean_c1_digits}
    \end{subfigure}\\
    \begin{subfigure}{0.18\linewidth}
    \centering
        \includegraphics[width=\linewidth]{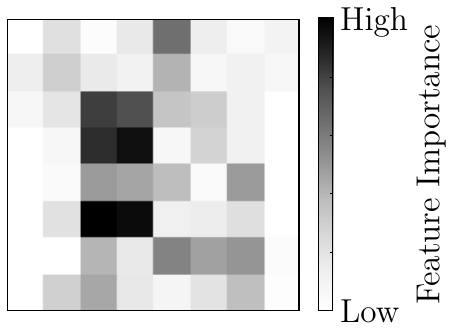}
    \caption{GRBFNN.}
    \label{fig:FI_digits_GR-RBF}
    \end{subfigure}
        \begin{subfigure}{0.18\linewidth}
    \centering
        \includegraphics[width=\linewidth]{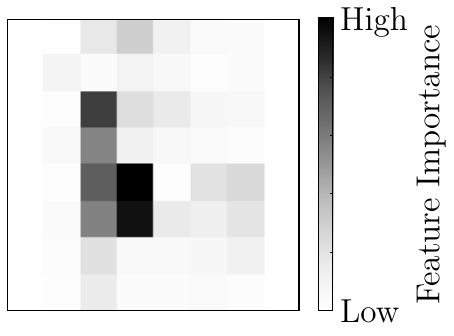}
    \caption{RF.}
    \label{fig:FI_digits_RF}
    \end{subfigure}
    \begin{subfigure}{0.18\linewidth}
    \centering
        \includegraphics[width=\linewidth]{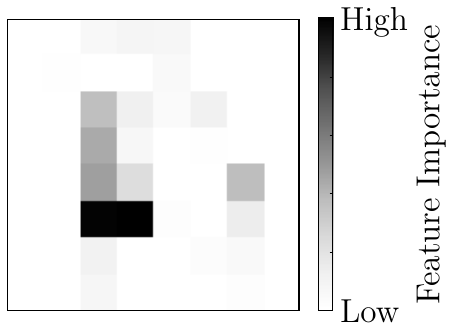}
    \caption{XGB.}
    \label{fig:FI_digits_XGB}
    \end{subfigure}
        \begin{subfigure}{0.18\linewidth}
    \centering
        \includegraphics[width=\linewidth]{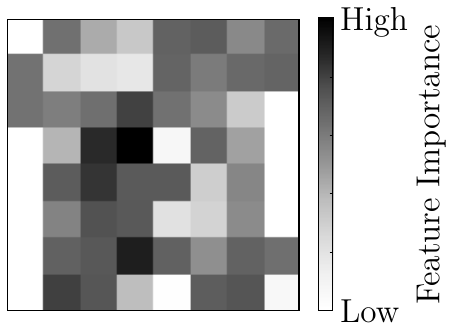}
    \caption{DFS.}
    \label{fig:FI_digits_DFS}
    \end{subfigure}
    \begin{subfigure}{0.18\linewidth}
    \centering
        \includegraphics[width=\linewidth]{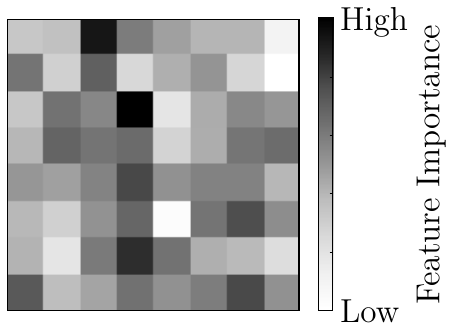}
    \caption{FIDL.}
    \label{fig:FI_digits_FIDL}
    \end{subfigure}
\caption{Graphical interpretation of the feature importance for the Digits dataset for all the models considered in this experiment. The feature importance should highlight the pixels where (a) and (b) differ. The GRBFNN employs an unsupervised center selection method (i.e. GRBFNN\textsubscript{$k$}) for this dataset, as evidenced by the superior performance in Tab. \ref{tab:num_res} with respect to  GRBFNN\textsubscript{$c$}.}
\label{fig:digits_FI}   
\end{figure}

We can easily show the feature importance obtained from the Digits dataset, which is composed only of the digit '8' and '3'. The mean of those two classes is visible in Fig. \ref{fig:mean_c0_digits} and Fig. \ref{fig:mean_c1_digits},
where each feature corresponds to a particular pixel intensity of greyscale value. 
This suggests that is easy to interpret the feature importance identified from the models because the important features should highlight where the digits '8' and '3' differ the most. 
In Fig. \ref{fig:digits_FI} we can verify the feature importance detected from the models. 
The GRBFNN employs an unsupervised center selection method for this dataset, as evidenced by the superior performance in Tab. \ref{tab:num_res} compared to the model with supervised center selection.

Meaningful feature importance is observable for the GRBFNN, RF, and XGB models. The GRBFNN (Fig. \ref{fig:FI_digits_GR-RBF}), similar to the RF (Fig. \ref{fig:FI_digits_RF}) the feature importance enhances pixels where those two classes differ. In contrast, the XGB (Fig. \ref{fig:FI_digits_XGB}) provides a sparser representation since almost all the importance is concentrated in only a few pixels. 
From Fig. \ref{fig:FI_digits_DFS} and Fig. \ref{fig:FI_digits_FIDL} seems that the two methods for feature importance learning for deep learning fail to provide explainable feature importances. 
\begin{figure}[!htb]
\centering
    \begin{subfigure}{0.49\linewidth}
    \centering
        \includegraphics[width=\linewidth]{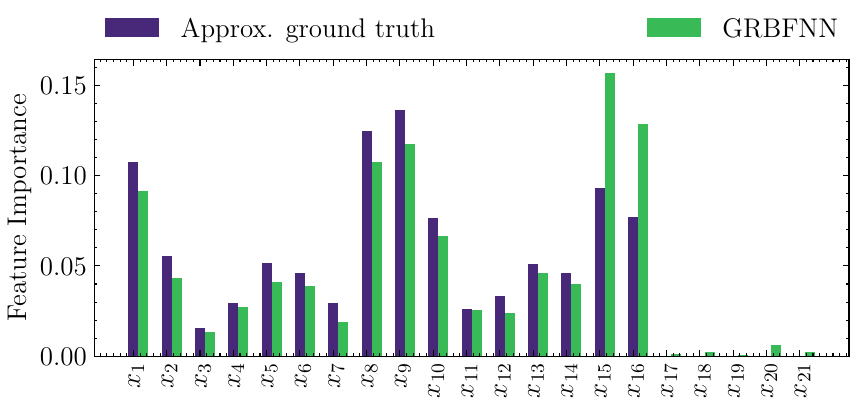}
    \caption{GRBFNN.}
    \label{fig:FIR_DTMB_5415_GR-RBF}
    \end{subfigure}\\
        \begin{subfigure}{0.49\linewidth}
    \centering
        \includegraphics[width=\linewidth]{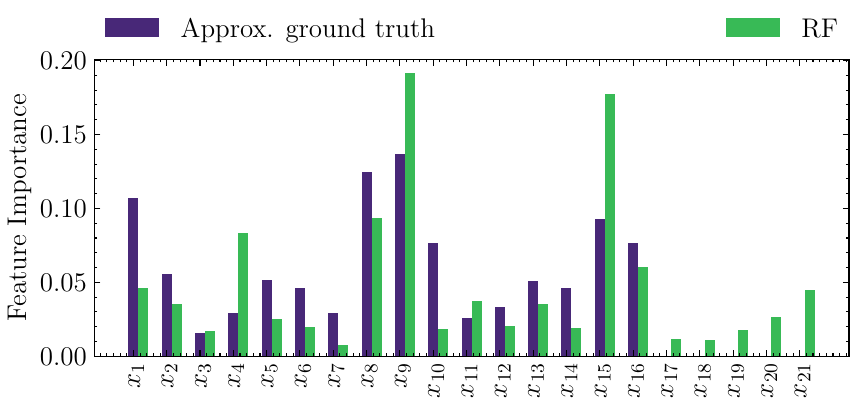}
    \caption{RF.}
    \label{fig:FIR_DTMB_5415_RF}
    \end{subfigure}
    \begin{subfigure}{0.49\linewidth}
    \centering
        \includegraphics[width=\linewidth]{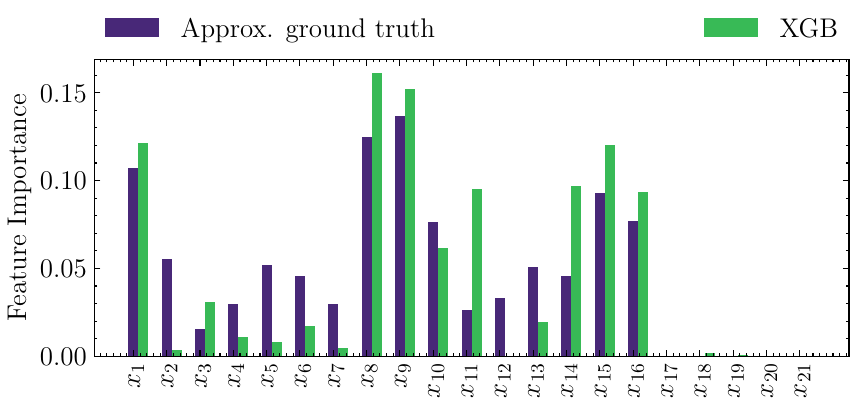}
    \caption{XGB.}
    \label{fig:FIR_DTMB_5415_XGB}
    \end{subfigure}
        \begin{subfigure}{0.49\linewidth}
    \centering
        \includegraphics[width=\linewidth]{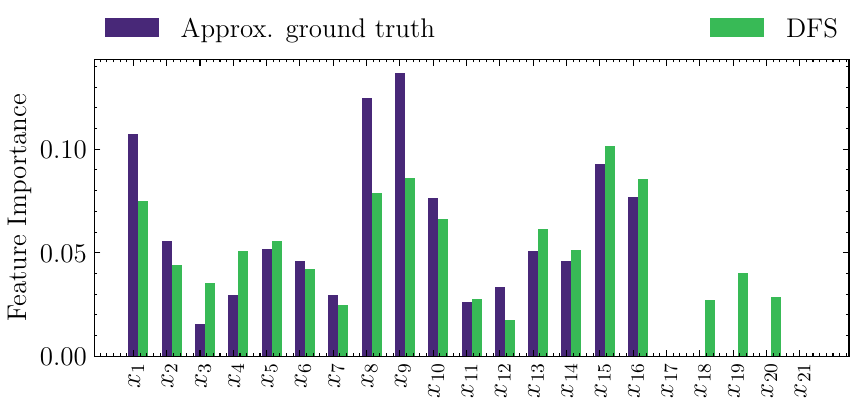}
    \caption{DFS.}
    \label{fig:FIR_DTMB_5415_DFS}
    \end{subfigure}
    \begin{subfigure}{0.49\linewidth}
    \centering
        \includegraphics[width=\linewidth]{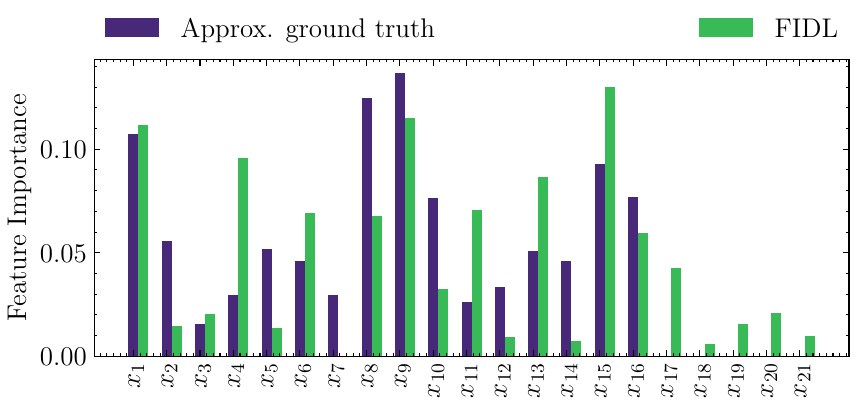}
    \caption{FIDL.}
    \label{fig:FIR_DTMB_5415_FIDL}
    \end{subfigure}
\caption{Graphical interpretation of the feature importance for the DTMB-$5415^{(2)}$ dataset for all the models considered in this experiment. Blue bars represent the approximated ground truth feature importance, in green is the one estimated by the models. The GRBFNN employs a supervised center selection method (i.e. GRBFNN\textsubscript{$c$}) for this dataset, as evidenced by the superior performance in Tab. \ref{tab:num_res} with respect to GRBFNN\textsubscript{$k$}.}
\label{fig:DTMB_FI}   
\end{figure}

\subsubsection{Feature Importance Ranking for the DTMB-5415 Dataset}
We provide a similar validation analysis for the DTMB-$5415^{(2)}$ dataset.
Based on the results presented in Tab. \ref{tab:num_res}, the GRBFNN model achieved better performance on this dataset using the supervised selection of the centers compared to the unsupervised selection of the centers.

For this particular problem, we can estimate an approximate ground truth for the feature importance of the true function, by computing the gradients at specific input variables $\mathbf{x}$. To do this, we employ a finite difference process, evaluating gradients at 84 different points. The estimated ground truth is obtained by averaging the absolute values of the gradient vectors at these 84 points.
In Fig. \ref{fig:DTMB_FI}, we show the approximated ground truth in blue in each bar plot. It should be noted that the true feature importance of the last five features is zero as these are not related to the true function. Consequently, the feature importance obtained from the models shown in green should be able to detect this.
\begin{figure}[!htb]
\centering
    \includegraphics[width=5cm]{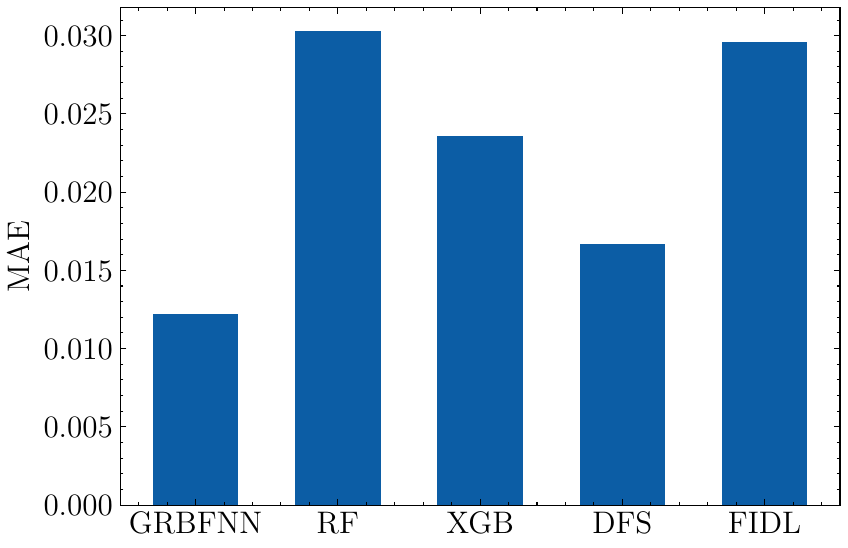}
    \caption{Bars represent the mean absolute error (MAE) ($y$-axis) between the approximated ground truth feature importance and the one estimated from the models ($x$-axis).}
    \label{fig:FIR_DTMB_5415_ERR}
\end{figure}

Only the GRBFNN (Fig. \ref{fig:FIR_DTMB_5415_GR-RBF}) and the XGB (Fig. \ref{fig:FIR_DTMB_5415_XGB}) can recognize that the last five features are not related to the problem, while the RF (Fig. \ref{fig:FIR_DTMB_5415_RF}),  DFS (Fig. \ref{fig:FIR_DTMB_5415_DFS}) and FIDL (Fig. \ref{fig:FIR_DTMB_5415_FIDL}) fail to identify that. 
To summarize the results, in Fig. \ref{fig:FIR_DTMB_5415_ERR} we show the bar plot regarding the mean absolute error between the approximated ground truth feature importance and the one estimated from all the methods, showing that in this case, the GRBFNN obtains the lowest error in detecting the underlying important variables of the regression task.

\subsubsection{Feature Importance Ranking for the Synthetic Datasets}
We test the same models on other synthetic datasets presented at the end of Section \ref{sec:datasets} designed specifically to evaluate feature selection and feature importance ranking models. 
In Tab. \ref{tab:num_res2} we resume the numerical results obtained with the same cross-validation procedure as in the previous numerical experiments.
\begin{table}[!htb]
\centering
\caption{Numerical results summary for the three synthetic problems. Problem 1 (P1) and Problem 2 (P2) are binary and multiclass classification tasks respectively and numbers represent the accuracy values. Problem 3 (P3) is a regression task and the numbers represent RMSE values. 
Bold numbers indicate the best-performing method on the test data that is statistically significant according to the Wilcoxon test ($\alpha=0.05$). The asterisk means that the method is the best on average but not demonstrated to be statistically significant.}
\scalebox{0.75}{
\begin{tabular}{llllllllllllll}
\toprule
 &
 &
  \multicolumn{2}{c}{GRBFNN\textsubscript{$k$}} &
  \multicolumn{2}{c}{GRBFNN\textsubscript{$c$}} &
  \multicolumn{2}{c}{RF} &
  \multicolumn{2}{c}{XGB} &
  \multicolumn{2}{c}{DFS} &
  \multicolumn{2}{c}{FIDL} \\
 &
  $N$ &
  Training &
  Test &
  Training &
  Test &
  Training &
  Test &
  Training &
  Test &
  Training &
  Test &
  Training &
  Test \\ \hline
\multirow{3}{*}{P1} 
& 100 & 0.980 & 0.609
  & 1.000 & 0.652
  & 0.992 & 0.672
  & 1.000 & 0.736
  & 1.000 & 0.762
  & 0.992 & $0.820^{*}$\\

  & 500 & 0.958 & 0.898
  & 0.981 & 0.892
  & 1.000 & 0.882
  & 1.000 & 0.803
  & 1.000 & 0.905
  & 0.942 & $0.914^{*}$\\
  
  & 1000 & 0.962 & \textbf{0.931}
  & 0.973 & 0.925
  & 1.000 & 0.901
  & 1.000 & 0.910
  & 0.991 & 0.897 
  & 0.934 & 0.918\\ \hline
\multirow{3}{*}{P2}          
  &100& 0.235 & 0.310
  & 0.250 & 0.249
  & 1.000 & 0.290
  & 0.993 & 0.289
  & 0.973 & \textbf{0.338}
  & 0.762 & 0.340\\
   
& 500  & 0.710 & 0.428
  & 0.706 & $0.508^{*}$
  & 1.000 & 0.486
  & 1.000 & 0.483
  & 0.898 & 0.497 
  & 0.568 & 0.496\\
  
  & 1000 &  0.643 & 0.538
  & 0.645 & $0.551^{*}$
  & 0.871 & 0.546
  & 1.000 & 0.507
  & 0.633 & 0.537  
  & 0.497 & 0.455\\ \hline
\multirow{3}{*}{P3} 
& 100  &  0.499 & 0.570
  & 0.497 & 0.570
  & 0.236 & 0.615
  & 0.001 & 0.506
  & 0.157 & \textbf{0.447} 
  & - & -   \\
  
  & 500 & 0.232 & 0.283
  & 0.184 & \textbf{0.255}
  & 0.163 & 0.440
  & 0.062 & 0.316
  & 0.190 & 0.289
  & - & -    \\
  
  & 1000 & 0.258 & 0.284
  & 0.198 & \textbf{0.221}
  & 0.142 & 0.387
  & 0.101 & 0.264
  & 0.196 & 0.240 
  & - & - \\ 
  \bottomrule
\end{tabular}}
\label{tab:num_res2}
\end{table}

Problem P1 is a binary classification problem and the FIDL shows the best performance for $N=100$ and $N=500$ (not statistically significant), while for $N = 1000$ the GRBFNN\textsubscript{$k$} (statistically significant) obtained the higher accuracy. 
In Tab. \ref{tab:P1_FI}, we have the feature importance related to problem P1.
We discuss GRBFNN\textsubscript{$c$} for $N=100$ while for $N=500$ and $N=1000$ we show GRBFNN\textsubscript{$k$} and refer as GRBFNN. 

For $N=100$, the GRBFNN provides meaningless feature importance across the methods due also to the lack of predictive performance obtained in this case, while the XGB and FIDL are the only models to recognize that the first four features are relevant. 
For $N=500$ and $N=1000$, the feature importance from the GRBFNN improves substantially together with its predictive performance. 
The DFS even if it provides competitive accuracy compared with other methods has some difficulty in highlighting the importance of the first four features from the remaining ones especially for $N=500$.

\begin{figure}[!htb]
\centering
    \begin{subfigure}{0.3\linewidth}
        \includegraphics[width=\linewidth]{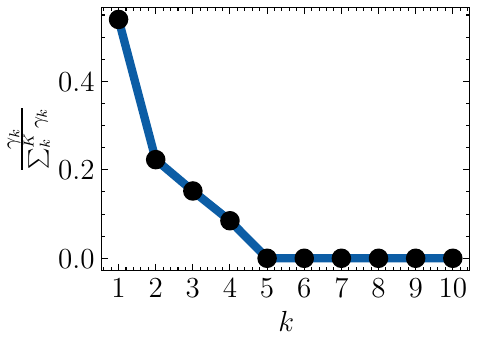}
    \caption{$N=100$.}
    \label{fig:gammas_bc_friedman_100_GR-RBF}
    \end{subfigure}
    \begin{subfigure}{0.3\linewidth}
        \includegraphics[width=\linewidth]{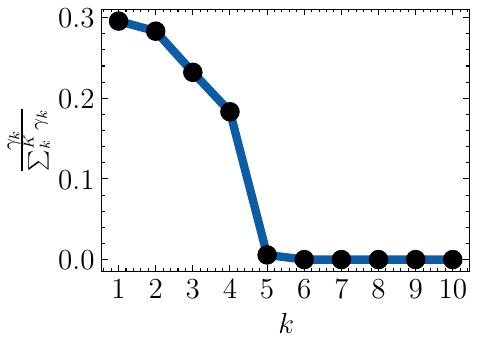}
    \caption{$N=500$.}
    \label{fig:gammas_bc_friedman_500_GR-RBF}
    \end{subfigure}
    \begin{subfigure}{0.3\linewidth}
        \includegraphics[width=\linewidth]{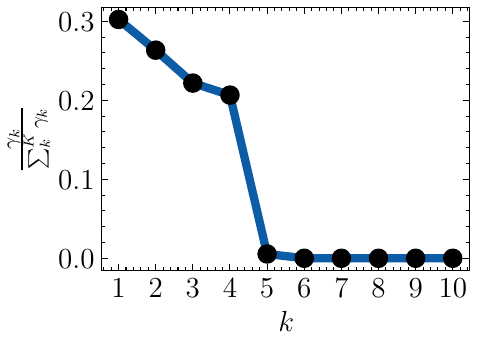}
    \caption{$N=1000$.}
    \label{fig:gammas_bc_friedman_1000_GR-RBF}
    \end{subfigure}
\caption{Eigenvalues decay for problem P1.}
\label{fig:gammas_P1}   
\end{figure}

As further support and analysis, we evaluate the eigenvalues decay that can help us to highlight the numbers of factors of variation of the learned model $f$.
In Fig. \ref{fig:gammas_P1}, we have the eigenvalues decay for problem P1 for the values of $N$ considered. It is possible to notice that for $N=1000$ and $N=500$ the GRBFNN varies mainly along four components which are also the number of the important features for P1. For $N=100$, there is not a clear identification of those factors within the latent space/active subspace due to the low predictive performance of the model.
\begin{table}[!htb]
\centering
\caption{Summary of the models feature importance obtained on problem P1. Note that the first four features are relevant in P1. The GRBFNN\textsubscript{$c$} or GRBFNN\textsubscript{$k$} are renamed as GRBFNN depending on which one achieved superior performance according to Table \ref{tab:num_res2}.}
\begin{tabular}{l*5{C}@{}}
\toprule
Model & GRBFNN & RF & XGB & DFS & FIDL \\ 
\midrule
P1 ($N=100$) & \includegraphics[width=\linewidth]{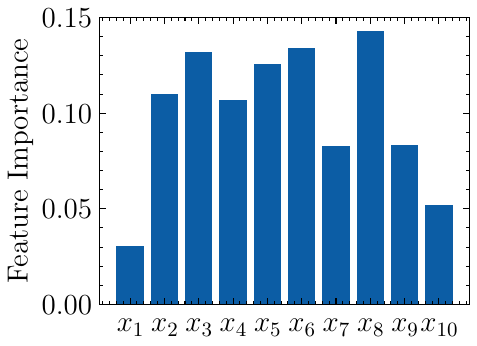} & \includegraphics[width=\linewidth]{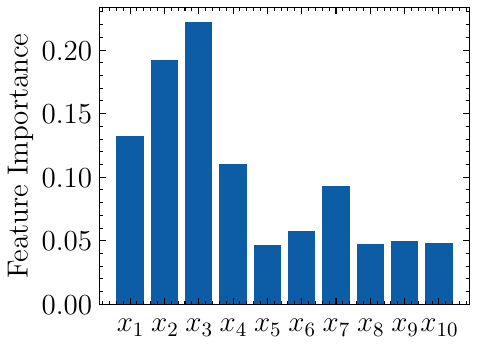} & \includegraphics[width=\linewidth]{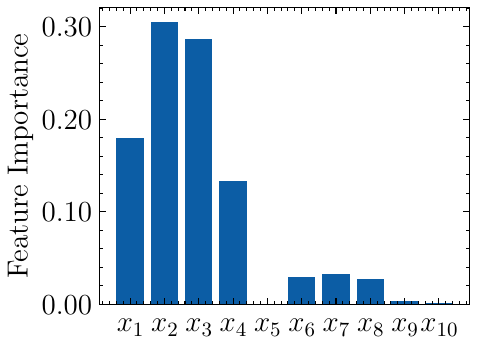} & \includegraphics[width=\linewidth]{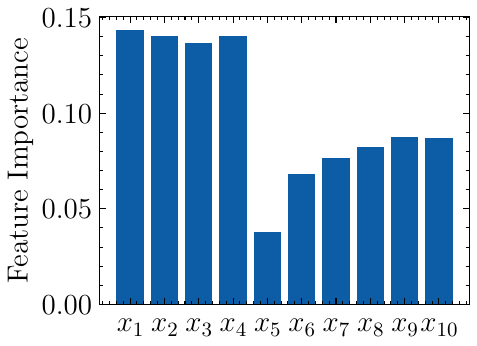} &
\includegraphics[width=\linewidth]{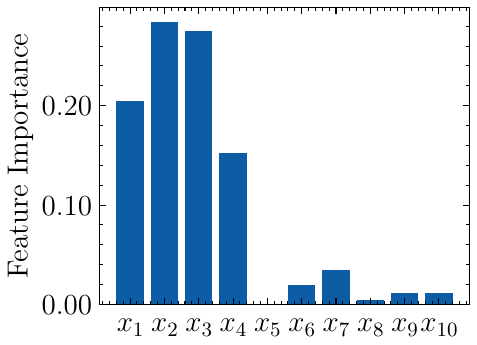} \\ 
P1 ($N=500$)  &  \includegraphics[width=\linewidth]{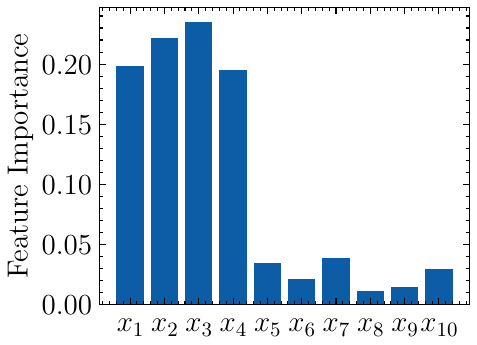} & \includegraphics[width=\linewidth]{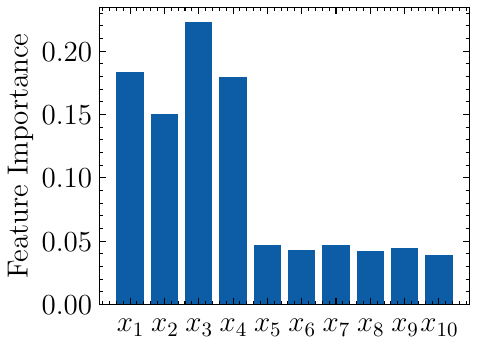} & \includegraphics[width=\linewidth]{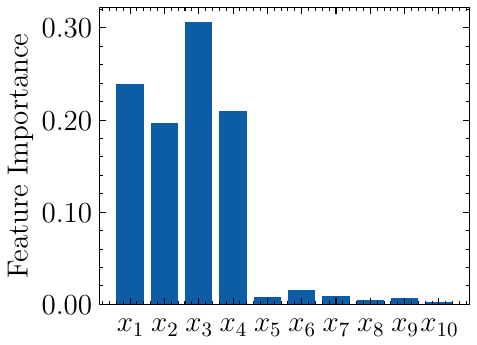} & \includegraphics[width=\linewidth]{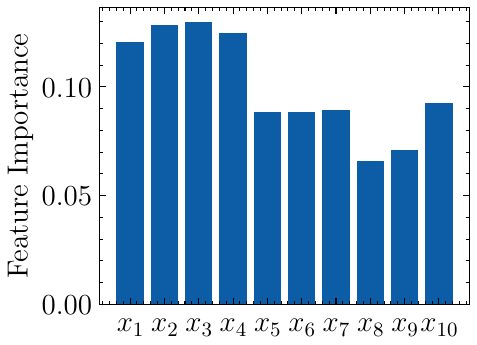} &
\includegraphics[width=\linewidth]{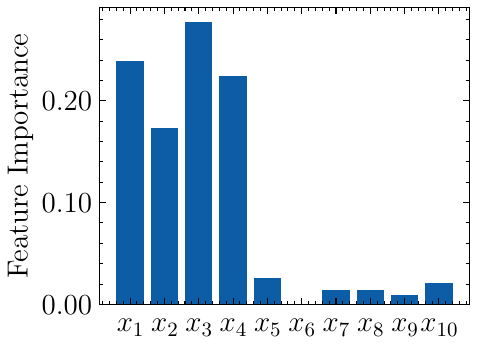} \\ 
P1 ($N=1000$)  & \includegraphics[width=\linewidth]{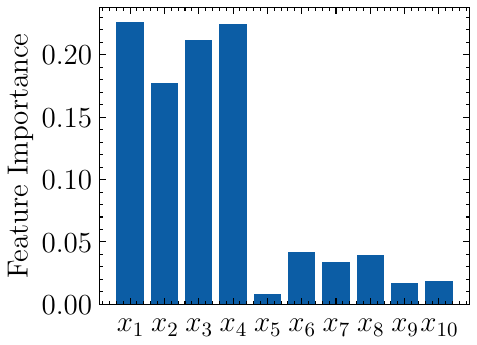} & \includegraphics[width=\linewidth]{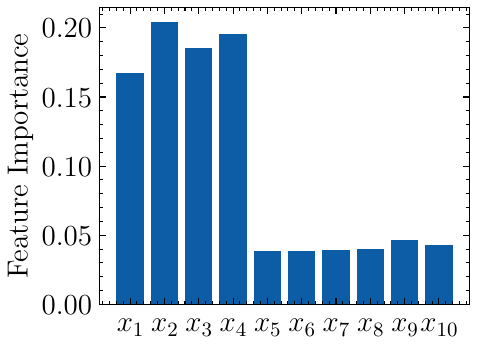} & \includegraphics[width=\linewidth]{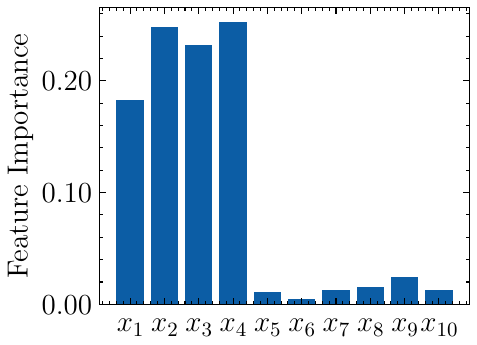} & \includegraphics[width=\linewidth]{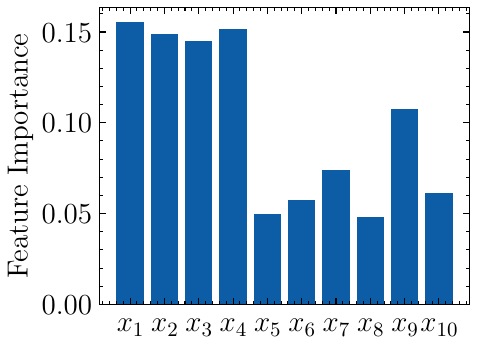} &
\includegraphics[width=\linewidth]{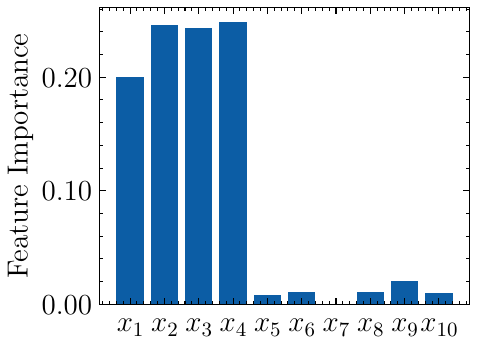} \\  
\bottomrule 
\end{tabular}
\label{tab:P1_FI}
\end{table}

Problem P2 is a difficult multiclass classification problem and DFS shows the best performance for $N=100$ (statistically significant) while for $N=500$ and $N=1000$ the GRBFNN\textsubscript{$c$} (not statistically significant) obtains the highest accuracy.  
In Tab. \ref{tab:P2_FI}, we have the feature importance related to problem P2.
We discuss GRBFNN\textsubscript{$k$} for $N=100$ while for $N=500$ and $N=1000$ we show GRBFNN\textsubscript{$c$} and both referred as GRBFNN.

Similarly as in the previous case for $N=100$, the GRBFNN and RF have some difficulty in detecting that the first three features are the most important while the FIDL provides the best feature ranking.  
Similar to problem P1, the feature importance from the GRBFNN improves significantly for $N=500$ and $N=1000$ along with its predictive performance. The GRBFNN and FIDL provide the most meaningful feature importance ranking in these cases, while the other methods fail to provide a clear separation between important and non-important variables. 

\begin{figure}[!htb]
\centering
    \begin{subfigure}{0.32\linewidth}
        \includegraphics[width=\linewidth]{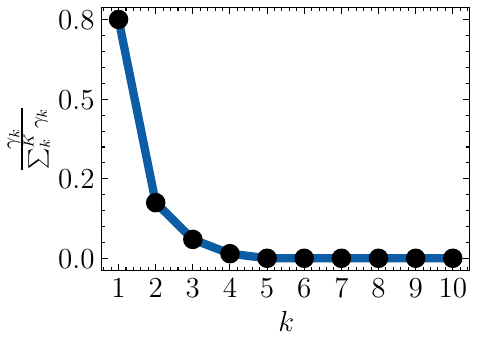}
    \caption{$N=100$.}
    \label{fig:gammas_XOR_100_GR-RBF}
    \end{subfigure}
    \begin{subfigure}{0.32\linewidth}
        \includegraphics[width=\linewidth]{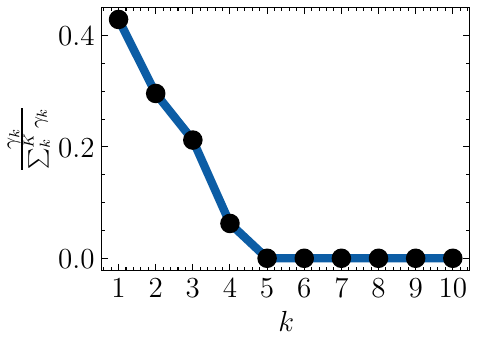}
    \caption{$N=500$.}
    \label{fig:gammas_XOR_500_GR-RBF}
    \end{subfigure}
    \begin{subfigure}{0.32\linewidth}
        \includegraphics[width=\linewidth]{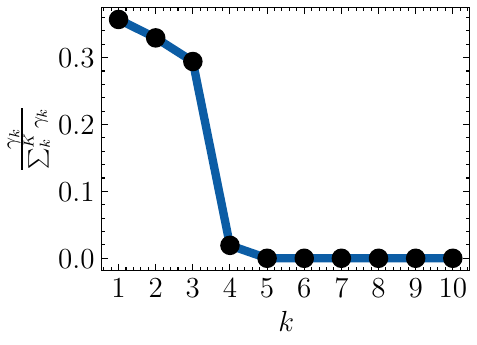}
    \caption{$N=1000$.}
    \label{fig:gammas_XOR_1000_GR-RBF}
    \end{subfigure}
\caption{Eigenvalues decay for problem P2.}
\label{fig:gammas_P2}   
\end{figure}

In Fig. \ref{fig:gammas_P2}, we have the eigenvalues decay for problem P2 for the values of $N$ considered. For $N=1000$, the GRBFNN varies mainly along three components which are also the number of the important features for P2, this behavior is less visible but still present for $N=100$ and $N=500$.
\begin{table}[!htb]
\centering
\caption{Summary of the models feature importance obtained on problem P2. Note that the first three features are relevant in P2. The GRBFNN\textsubscript{$c$} or GRBFNN\textsubscript{$k$} are renamed as GRBFNN depending on which one achieved superior performance according to Table \ref{tab:num_res2}.}
\begin{tabular}{l*5{C}@{}}
\toprule
Model & GRBFNN & RF & XGB & DFS & FIDL \\ 
\midrule
P2 ($N=100$) & \includegraphics[width=\linewidth]{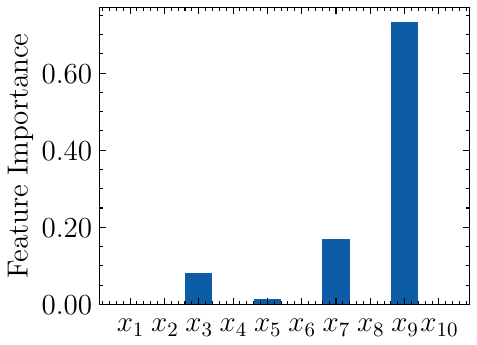} & \includegraphics[width=\linewidth]{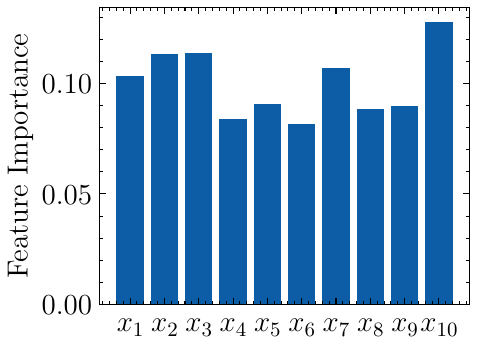} & \includegraphics[width=\linewidth]{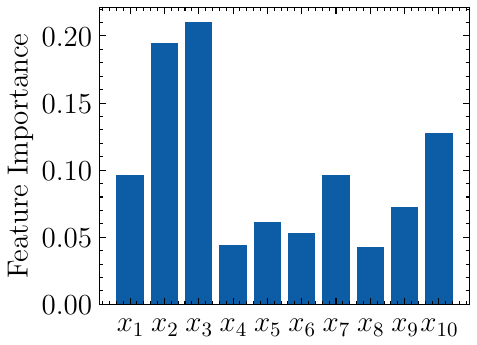} & \includegraphics[width=\linewidth]{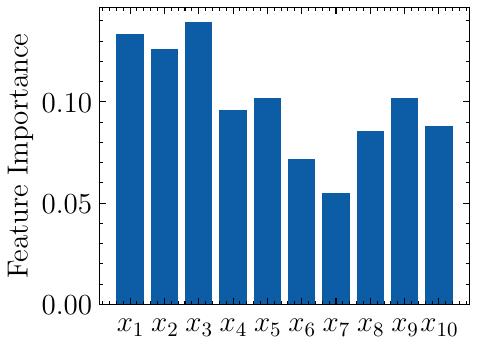} &
\includegraphics[width=\linewidth]{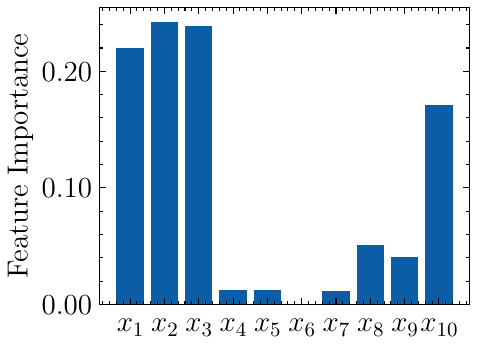} \\ 
P2 ($N=500$)  & \includegraphics[width=\linewidth]{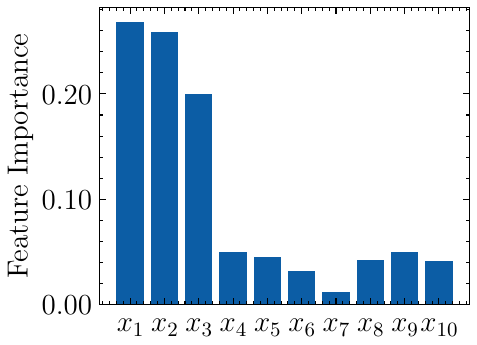} & \includegraphics[width=\linewidth]{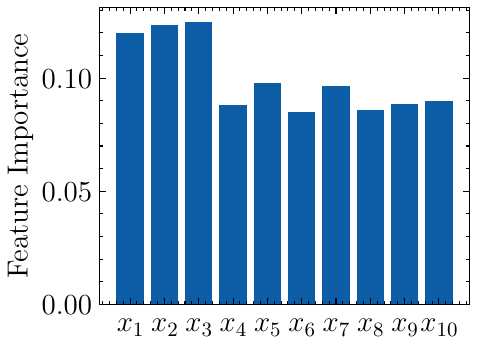} & \includegraphics[width=\linewidth]{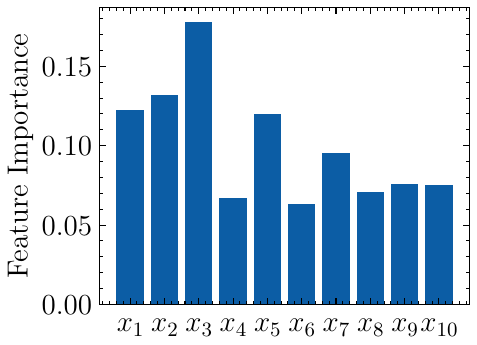} & \includegraphics[width=\linewidth]{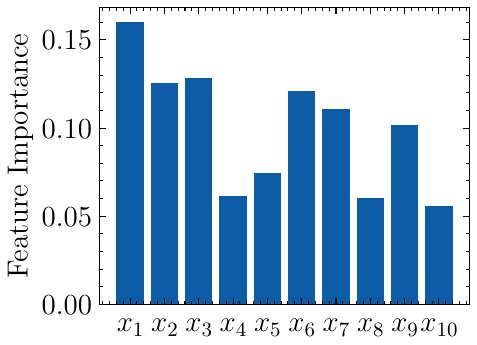} &
\includegraphics[width=\linewidth]{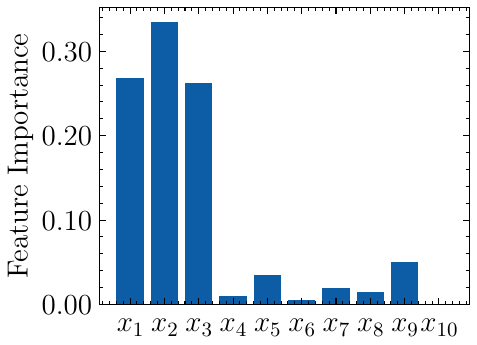} \\ 
P2 ($N=1000$)  & \includegraphics[width=\linewidth]{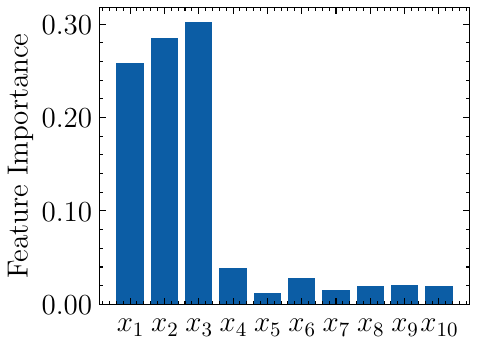} & \includegraphics[width=\linewidth]{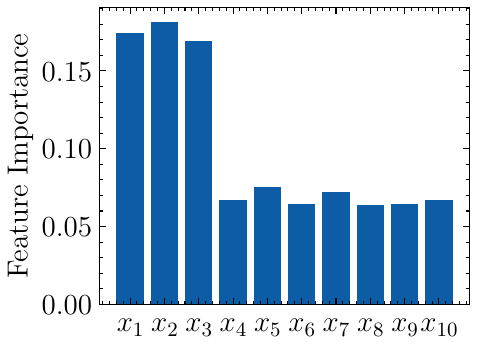} & \includegraphics[width=\linewidth]{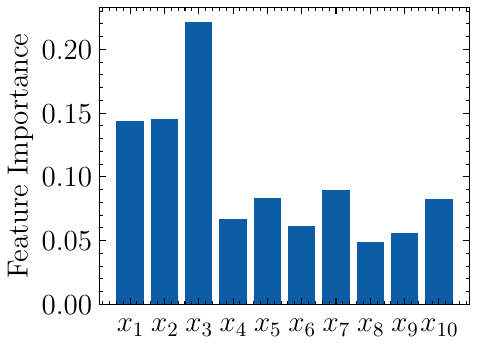} & \includegraphics[width=\linewidth]{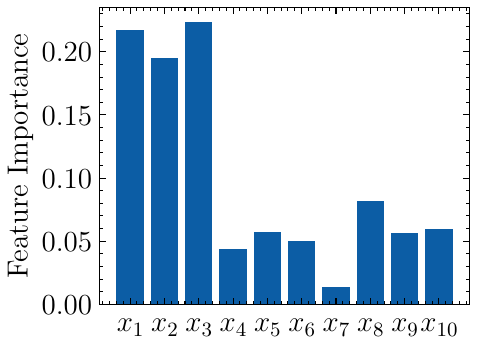} &
\includegraphics[width=\linewidth]{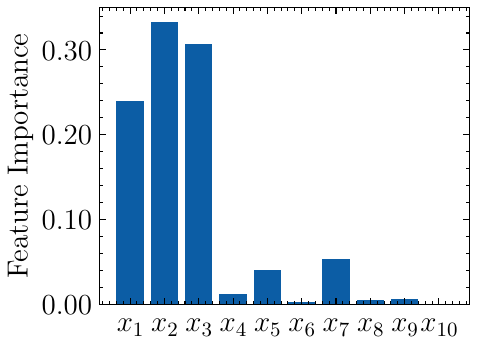} \\ 
\bottomrule 
\end{tabular}
\label{tab:P2_FI}
\end{table}

Problem P3 is a nonlinear regression problem with the DFS showing the best performance for $N=100$ (statistically significant) while for $N=500$ and $N=1000$ the GRBFNN\textsubscript{$c$} (statistically significant) obtains the lowest RMSE.  
In Tab. \ref{tab:P3_FI}, we can analyze the feature importance related to problem P3. 
We discuss GRBFNN\textsubscript{$c$} for $N=100$, $N=500$ and $N=1000$ and referred as GRBFNN.

For $N=100$, seems that all the models recognize that only the first five features are important in problem P3. 
Also for $N=500$ and $N=1000$ the GRBFNN, RF, XGB, and FIDL correctly ignore the contribution of the last five features, differently from DFS. 
Interestingly, for $N=500$ and $N=1000$, the feature importance from GRBFNN differs from all the other models where they recognize the feature $x_4$ as the most important only for $N=100$. 

\begin{figure}[!ht]
\centering
    \begin{subfigure}{0.32\linewidth}
        \includegraphics[width=\linewidth]{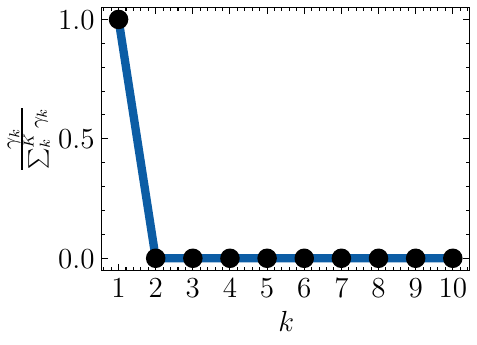}
    \caption{$N = 100$.}
    \label{fig:gammas_friedman_100_10_1_GR-RBF}
    \end{subfigure}
    \begin{subfigure}{0.32\linewidth}
        \includegraphics[width=\linewidth]{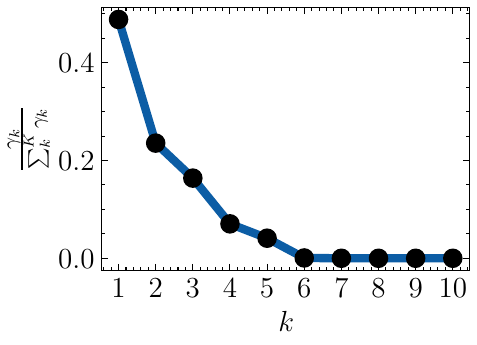}
    \caption{$N = 500$.}
    \label{fig:gammas_friedman_500_10_1_GR-RBF}
    \end{subfigure}
    \begin{subfigure}{0.32\linewidth}
        \includegraphics[width=\linewidth]{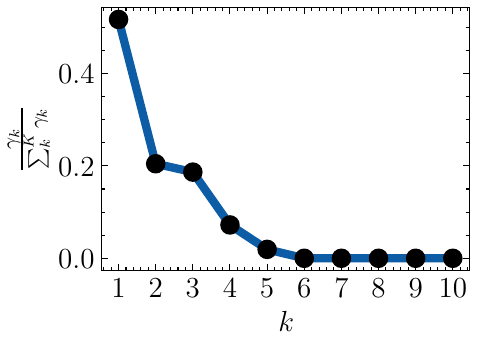}
    \caption{$N = 1000$.}
    \label{fig:gammas_friedman_1000_10_1_GR-RBF}
    \end{subfigure}
\caption{Eigenvalues decay for problem P3.}
\label{fig:gammas_P3}   
\end{figure}

In Fig. \ref{fig:gammas_P3}, we have the eigenvalues decay for problem P3 for the values of $N$ considered. For $N=100$ the model varies only along one latent variable while $N=500$ and $N=1000$, we have approximately the first five eigenvalues that are different from zero.

\begin{table}[!htb]
\centering
\caption{Summary of the models feature importance obtained on problem P3. Note that the first five features are relevant in P3. The GRBFNN\textsubscript{$c$} or GRBFNN\textsubscript{$k$} are renamed as GRBFNN depending on which one achieved superior performance according to Table \ref{tab:num_res2}.}
\begin{tabular}{l*5{C}@{}}
\toprule
Model & GRBFNN & RF & XGB & DFS & FIDL \\ 
\midrule
P3 ($N=100$) & \includegraphics[width=\linewidth]{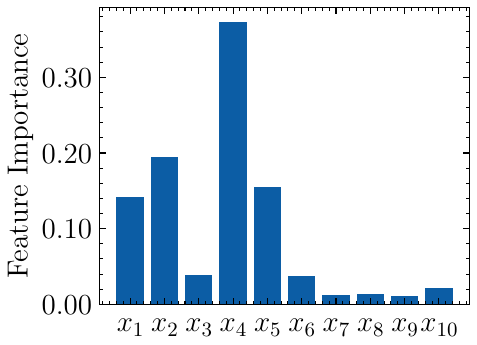} & \includegraphics[width=\linewidth]{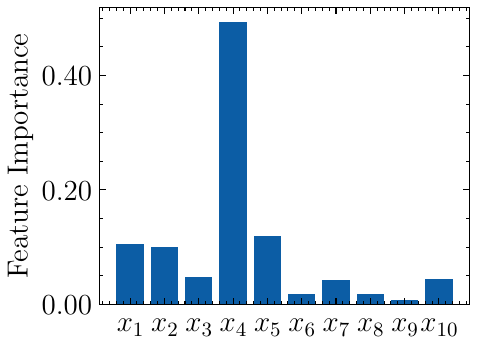} & \includegraphics[width=\linewidth]{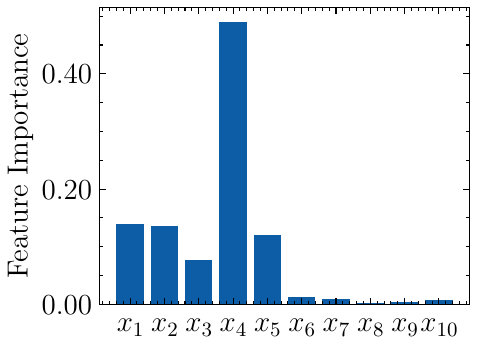} & \includegraphics[width=\linewidth]{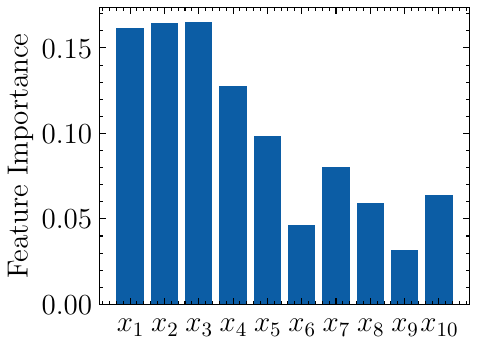} &
\includegraphics[width=\linewidth]{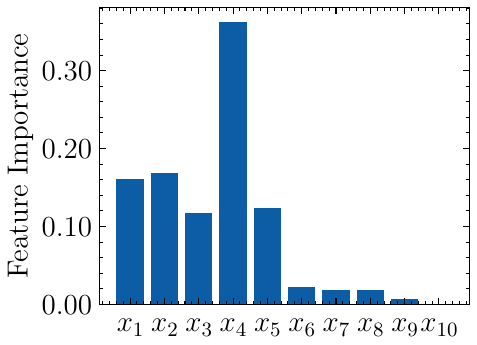} \\ 
P3 ($N=500$)  &  \includegraphics[width=\linewidth]{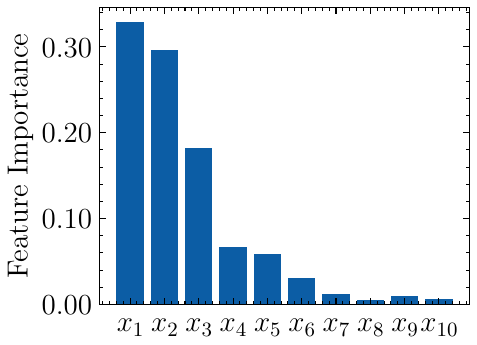} & \includegraphics[width=\linewidth]{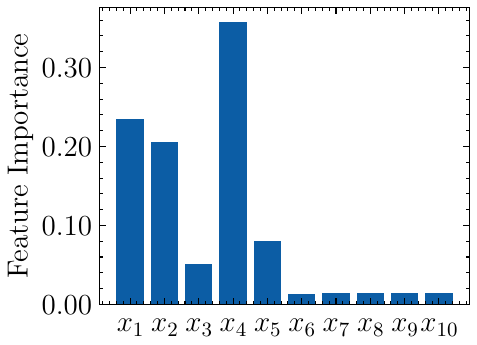} & \includegraphics[width=\linewidth]{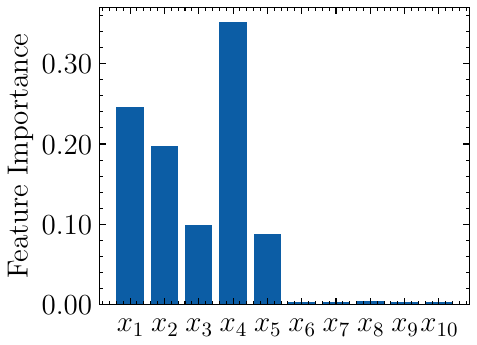} & \includegraphics[width=\linewidth]{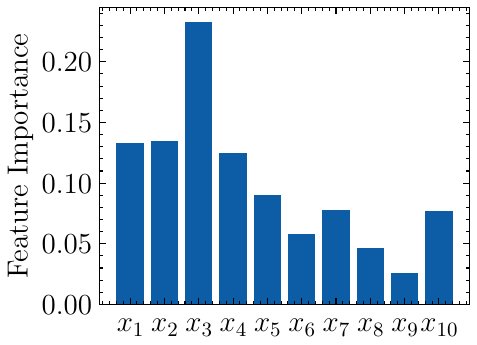} &
\includegraphics[width=\linewidth]{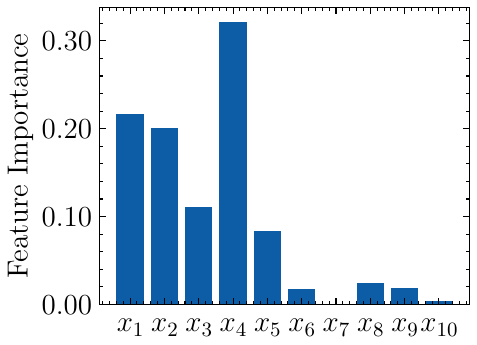} \\ 
P3 ($N=1000$)  & \includegraphics[width=\linewidth]{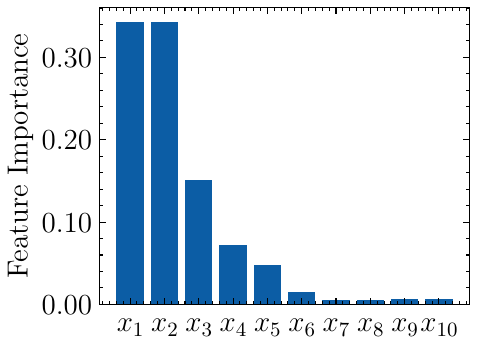} & \includegraphics[width=\linewidth]{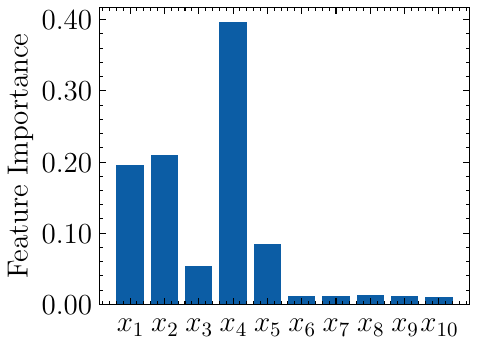} & \includegraphics[width=\linewidth]{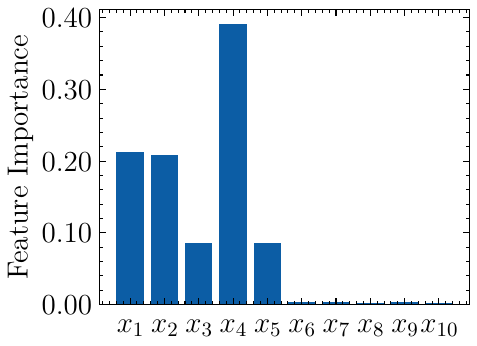} & \includegraphics[width=\linewidth]{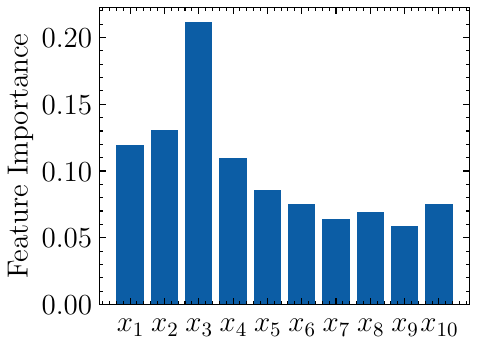} &
\includegraphics[width=\linewidth]{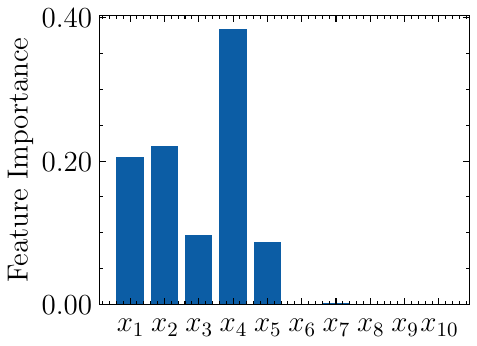} \\  
\bottomrule 
\end{tabular}
\label{tab:P3_FI}
\end{table}
\FloatBarrier
\section{Conclusion and Future Work}
In this paper, we proposed modifications to the classical RBFNN model,  to enhance its interpretability by highlighting the important features and estimating the active subspace of the model. 
This is achieved by incorporating a learnable precision matrix into the Gaussian kernel. The latent information about the learned model can be extracted from the eigenvalues and eigenvectors of the estimated precision matrix.

Our extensive numerical experiments covered regression, classification, and feature selection tasks, where we compared our proposed model with widely used methods such as SVM, MLP, RF, XGB, FT-T (a transformer architecture for tabular data), and state-of-the-art deep learning-based embedding methods such as DFS and FIDL. 

The results demonstrated that our GRBFNN model achieved attractive prediction performance while providing meaningful feature importance rankings. 
One of the key observations from our experiments was the impact of the regularizer $\lambda_{\mathbf{u}}$ on the performance of the GRBFNN, which often prevails the effect of the weight regularizer $\lambda_{\mathbf{w}}$. 
This finding suggests that prioritizing the regularization of the precision matrix yields more significant improvements in the model generalization performance. 

Moreover, the model enables supervised dimensionality reduction in the active subspace, facilitating visualization and comprehension of complex phenomena. In parallel, it offers insights into the impact of individual input features through a feature importance ranking. This capability not only facilitates a clearer understanding of the model but also allows for effective feature selection tasks.

In summary, by combining predictive power with interpretability, the GRBFNN offers a valuable tool for understanding complex nonlinear relationships in the data.  
Overall, our work contributes to bridging the gap between black-box neural network models and interpretable machine learning, enabling users to not only make accurate predictions but also gain meaningful insights from the model behavior and improve decision-making processes in real-world applications.

Looking ahead, we plan to apply our model to tackle the so-called curse of dimensionality in expensive engineering optimization problems. By leveraging the active subspace estimation, we aim to reduce the dimensionality of optimization problems without relying on direct gradient computations as in the classical active subspace method, which is not desirable in noisy simulation scenarios. 
Additionally, we plan to broaden the application of our model in domains where accurate predictions and interpretable insights are crucial such as in healthcare. 
\section{Acknowledgement}
The research is supported by Prof. Shoemaker’s Distinguished Professor Chair fund from National University of Singapore (NUS) and her startup funds from NUS.

\FloatBarrier
\bibliographystyle{plain}
\bibliography{biblio} 
\newpage
\section{Supplementary Material}\label{sec:supplementary}
\subsection{Datasets Description}\label{sec:supplementary_data}
The following is a detailed description of the real-world datasets used in this study:
\begin{itemize}
\item \textbf{Digits} \cite{alimoglu1996methods}: They created a digit database by collecting 250 samples from 44 writers. The samples written by 30 writers are used for training, cross-validation, and writer-dependent testing, and the digits written by the other 14 are used for writer-independent testing. For the current experiment, we use the digits $3$ and $8$ for feature selection purposes so that $N = 357$ and $D=64$.
\item \textbf{Iris} \cite{fisher1936use}: One of the most famous datasets in the pattern recognition literature, contains 3 classes of 50 instances each ($N=150, D=4$), where each class refers to a type of iris plant. One class is linearly separable from the other 2; the latter are not linearly separable from each other.
\item \textbf{Breast Cancer} \cite{street1993nuclear}: Features are computed from a digitized image of a fine needle aspirate (FNA) of a breast mass. They describe the characteristics of the cell nuclei present in the image. In this dataset, $N = 569$, $D=30$, and two classes.
\item \textbf{Wine} \cite{aeberhard1994comparative}: The data is the results of a chemical analysis of wines grown in the same region in Italy by three different cultivators. There are thirteen different measurements taken for different constituents found in the three types of wine. In this dataset, $N = 173$, $D=13$, and three classes.
\item \textbf{Australian} \cite{Dua:2019}: This is the famous Australian Credit Approval dataset, originating from the StatLog project. It concerns credit card applications. All attribute names and values have been changed to meaningless symbols to protect the confidentiality of the data. In this dataset, $N = 600$, $D=15$, and two classes.
\item \textbf{Credit-g} \cite{Dua:2019}: This dataset classifies people described by a set of attributes as good or bad credit risks, there are $D=20$ features, $N=1000$ data points, and two classes in this dataset.
\item \textbf{Glass} \cite{evett:1987:rifs} : The Glass identification database. The study of the classification of types of glass was motivated by criminological investigation. There are $D=9$ features, $N=214$ data points, and two classes in this dataset. 
\item \textbf{Blood} \cite{YEH20095866}: Data taken from the Blood Transfusion Service Center in Hsin-Chu City in Taiwan. The center passes its blood transfusion service bus to one university in Hsin-Chu City to gather blood donated about every three months. The target attribute is a binary variable representing whether he/she donated blood in March 2007. This dataset has $D=4$ features, $N=748$ data points, and two classes. 
\item \textbf{Heart Disease} \cite{Dua:2019}: This database contains 76 attributes, but all published experiments refer to using a subset of $D=14$ of them and $N=270$. The goal is to predict the presence of heart disease in the patient.
\item \textbf{Vowel} \cite{deterding1989speaker}: Speaker-independent recognition of the eleven steady state vowels of British English using a specified training set of lpc derived log area ratios.
This dataset has $D=12$ features, $N=990$ data points, and eleven classes. 
\item \textbf{Dehli Weather} \cite{DWdataset}: The Delhi weather dataset was transformed from a time series problem into a supervised learning problem by using past time steps as input variables and the subsequent time step as the output variable, representing the humidity. In this dataset $N=1461$ and $D=7$.
\item \textbf{Boston} \cite{pace1997sparse}: This dataset contains information collected by the U.S Census Service concerning housing in the area of Boston Mass and has been used extensively throughout the literature to benchmark algorithms for regression. In this dataset $N=506$ and $D=14$.
\item \textbf{Diabetes} \cite{smith1988using}: Ten baseline variables, age, sex, body mass index, average blood pressure, and six blood serum measurements were obtained for each of $N = 442$ diabetes patients, as well as the response of interest, a quantitative measure of disease progression one year after baseline.
\item \textbf{Prostatic Cancer} \cite{stamey1989prostate}: The study examined the correlation between the level of prostate-specific antigen (PSA) and several clinical measures, in $N=97$ men who were about to receive radical prostatectomy. The goal is to predict the log of PSA (lpsa) from given measurements of $D=4$ features.
\item \textbf{Liver} \cite{MCDERMOTT201641}: It is a regression problem where the first 5 variables are all blood tests that are thought to be sensitive to liver disorders that might arise from excessive alcohol consumption. Each line in the dataset constitutes the record of a single male individual. There are $D=5$ features, $N=345$ data points.
\item \textbf{Plasma} \cite{nierenberg1989determinants}: A cross-sectional study has been designed to investigate the relationship between personal characteristics and dietary factors, and plasma concentrations of retinol, beta-carotene, and other carotenoids. Study subjects ($N = 315$) were patients who had an elective surgical procedure during a three-year period to biopsy or remove a lesion of the lung, colon, breast, skin, ovary, or uterus that was found to be non-cancerous.
\item \textbf{Cloud} \cite{Dua:2019}: The data sets we propose to analyze are constituted of $N=1024$ vectors, each vector includes $D=10$ parameters. Each image is divided into super-pixels 16*16 and in each super-pixel, we compute a set of parameters for the visible (mean, max, min, mean distribution, contrast, entropy, second angular momentum) and IR (mean, max, min).
\item \textbf{DTMB-5415} \cite{d2024generative}: The DTMB-5415 datasets come from a real-world naval hydrodynamics problem. The $21$ input variables represent the design variables responsible for the shape modification of the hull while the output variable represents the corresponding total resistance coefficient of the simulated hull through a potential flow simulator. We propose two versions of the same problem: DTMB-$5415^{1}$ where all the $21$ design variables are related to the output variable and DTMB-$5415^{2}$ where $5$ of the $21$ design variables are not related to the output so that in this manner we can evaluate the models also from a feature selection perspective.

\item \textbf{Body Fat} \cite{penrose1985generalized}:  Estimates of the percentage of body fat are determined by underwater weighing and various body circumference measurements for $N=252$ men and $D=14$ different input features. 
\end{itemize}

\subsection{Numerical Set-up}
In this section, we present the numerical details of our experiment. We evaluate the performance of the GRBFNN model for feature selection with unsupervised (GRBFNN$_{k}$) and supervised (GRBFNN$_{c}$) center selection, as defined in Eq.\ref{eq:min_rbf_uns} and Eq. \ref{eq:min_rbf_sup}, respectively. To compute the centers for GRBFNN$_{k}$, we use the popular $k$-means clustering algorithm. Both GRBFNN\textsubscript{$c$} and GRBFNN\textsubscript{$k$} were optimized using Adam \cite{kingma2014adam} for a maximum of 10000 epochs. We implemented the GRBFNN in Pytorch \cite{NEURIPS2019_9015}.

We perform a grid search to approximately find the best set of hyperparameters of all the models considered in these numerical experiments.\\
For the GRBFNN the grid search is composed as follows:
\begin{itemize}
    \item Number of centers: For regression problems, the number of centers $M$ can take the following values $M\in\{8, 32, 128\}$, where for classification $M \in \{2, 4, 8, 16, 32\}$. 
    \item Regularizers: $(\lambda_{\mathbf{w}}$, $\lambda_{\mathbf{u}}) \in\{0, 10^{-3}, 10^{-2}, 10^{-1}, 1, 10^{1}, 10^{2}, 10^{3}\}$. 
    \item Adam learning rate: $\alpha \in \{10^{-3}, 10^{-2}\}$.
\end{itemize}
For the SVM model, we used a Gaussian kernel, and the grid search is composed as follows: 
\begin{itemize}
    \item Gaussian kernel width: $\sigma^2\in\{10^{-3}, 10^{-2}, 10^{-1}, 1, 10^{1}, 10^{2}, 10^{3}\}$. 
    \item Regularizer: $C\in\{0, 10^{-3}, 10^{-2}, 10^{-1}, 1, 10^{1}, 10^{2}, 10^{3}\}$. 
\end{itemize}
For the RF model, the grid search is composed as follows: 
\begin{itemize}
    \item Depth of the tree: $d_t \in\{2, 4, 8, 16, 32, 64, 128\}$.
    \item Minimum number of samples required to be a leaf node: $s_t \in \{1, 5, 10, 20\}$.
    \item Number of decision trees: $n_t \in \{10, 20, 50, 100, 200, 400, 800\}$. 
\end{itemize}
For the XGB model, the grid search is composed as follows: 
\begin{itemize}
    \item Learning rate: $l_b \in\{10^{-3}, 10^{-2}, 10^{-1}, 1\}$.
    \item Number of boosting stages: $n_b \in \{10, 20, 50, 100, 200, 400, 800\}$.
    \item Maximum depth of the individual regression estimators: $d_b \in \{2, 4, 8, 16, 32, 64, 128\}$.
\end{itemize}
For the MLP the grid search is composed as follows:  
\begin{itemize}
    \item Regularizer ($L^2$-norm): $\lambda \in \{0, 10^{-3}, 10^{-2}, 10^{-1}, 1, 10^{1}, 10^{2}, 10^{3}\}$
    \item Network architecture: A two-hidden-layer architecture with the following combinations of the number of neurons in the two layers is considered $\{(D, \lceil D/2 \rceil)), (2D, D), (2D, \lceil D/2 \rceil)\}$ with rectifiers activation functions \cite{fukushima1975cognitron}.
    \item Adam learning rate: $\alpha \in \{10^{-3}, 10^{-2}\}$.
\end{itemize}
For the DFS we use the same grid search hyperparameters as in the MLP case. This is because the DFS is the same as an MLP but with an additional
sparse one-to-one layer added between the input and the first hidden layer, where each input feature is weighted. We use the implementation available on the following link \footnote{\url{https://github.com/cyustcer/Deep-Feature-Selection}}. \\
For the FIDL, we use the author's implementation of the algorithm available at the following link \footnote{\url{https://github.com/maksym33/FeatureImportanceDL}} and the following grid search hyperparameters:
\begin{itemize}
    \item Network architecture: A two-hidden-layer architecture with the following combinations of the number of neurons in the two layers is considered $\{(D, \lceil D/2 \rceil)), (2D, D), (2D, \lceil D/2 \rceil)\}$.
    \item Number of important features: $s = \lceil D/2 \rceil$.
\end{itemize}
Always referring to FIDL, for datasets that are used also in their paper, we use the optimal set of hyperparameters found by them.
In this method, the user has to choose in advance, and before training the model, the number of important features $s$ that the problem might have. 
We fix this parameter to $s = \lceil D/2 \rceil$ as used in their paper for some datasets. We fix all the other hyperparameters to their default values provided by the authors. \\
The FT-T model has several hyperparameters together with the AdamW \cite{loshchilov2019decoupled} used for the optimization:
\begin{itemize}
    \item Number of blocks: $\{2, 3, 4\}$
    \item Embedding size of each feature: $\{64, 128\}$
    \item Number of attention heads: $\{4, 8\}$
    \item Attention drop-out: $\{0.1, 0.3\}$
    \item Hidden representation size: 4/3 * Embedding size
    \item Hidden representation drop-out: $\{0.1, 0.3\}$
    \item AdamW learning rate: $\{10^{-4}, 50^{-5}\}$
    \item AdamW weight decay:$\{10^{-5}, 10^{-6}\}$
\end{itemize}
We evaluated the model picking 20 random combinations of the aforementioned hyperparameters.
We used the author's implementation of the FT-T model freely available at the following link \footnote{\url{https://github.com/yandex-research/rtdl-revisiting-models}}.
For the SVM, MLP, and RF, we used the Python package \cite{scikit-learn} while for XGB \cite{Chen_2016}.
\begin{figure}[!htbp]
   \centering
\setlength{\tabcolsep}{12pt}
\begin{tabular}{l*4{C}@{}}
& Digits & Iris & Breast Cancer & Wine\\
Training& 
\includegraphics[width=1.85cm]{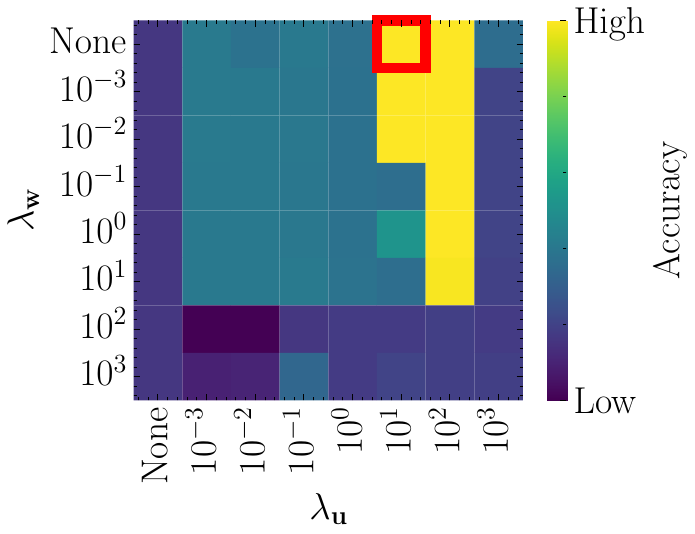}&
\includegraphics[width=1.85cm]{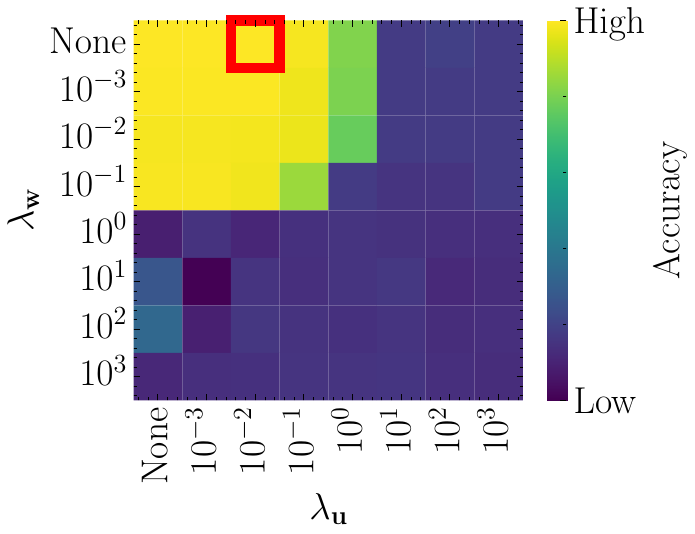}&
\includegraphics[width=1.85cm]{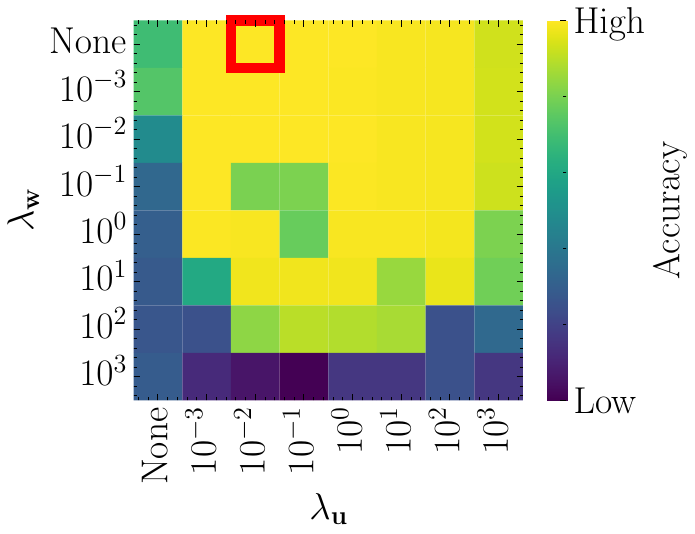}&
\includegraphics[width=1.85cm]{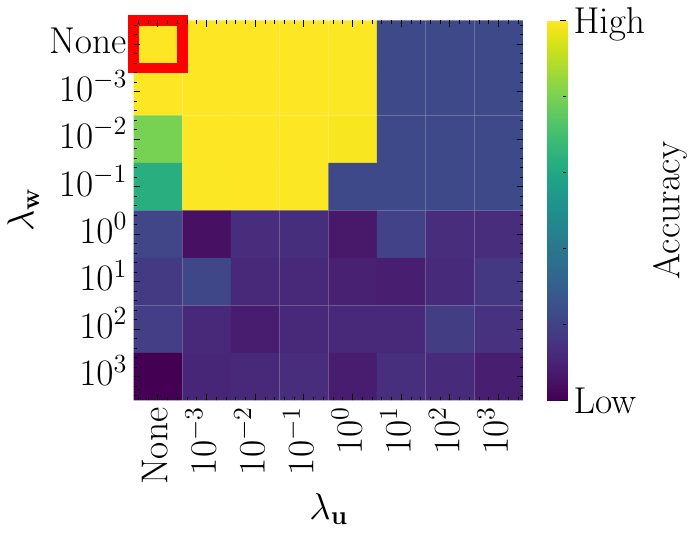}\\
Test&
\includegraphics[width=1.85cm]{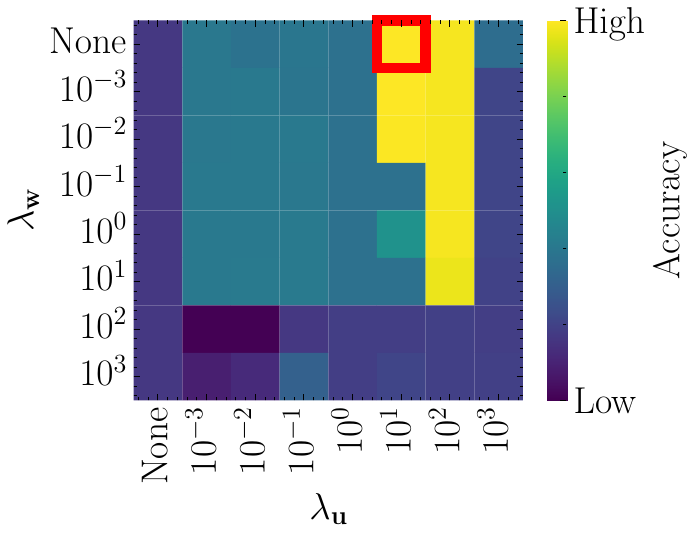}&
\includegraphics[width=1.85cm]{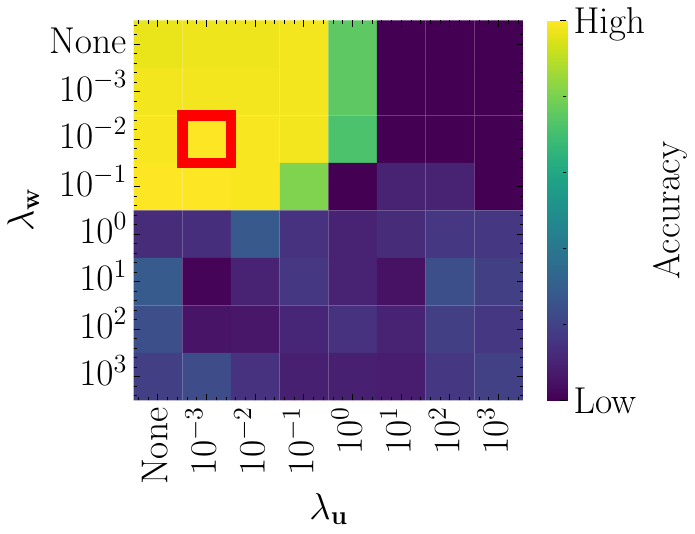}&
\includegraphics[width=1.85cm]{figs/reg_analysis_test_breast_cancer_GR-RBF.pdf}&
\includegraphics[width=1.85cm]{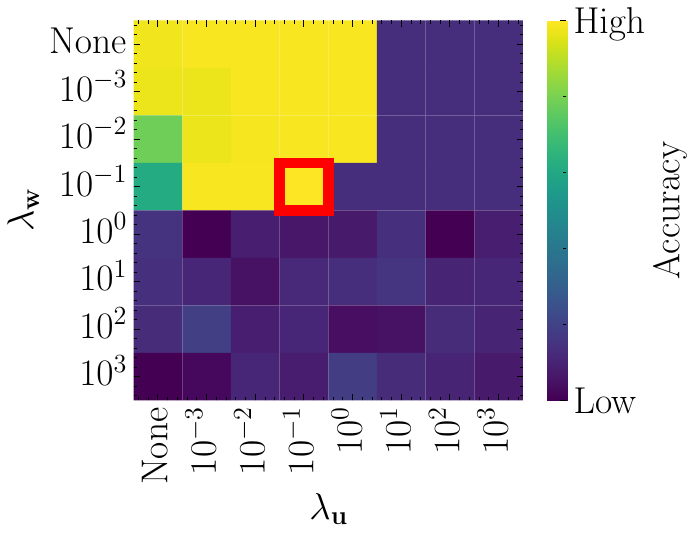}\\
& Australian & Credit-g & Glass & Blood \\
Training& 
\includegraphics[width=1.85cm]{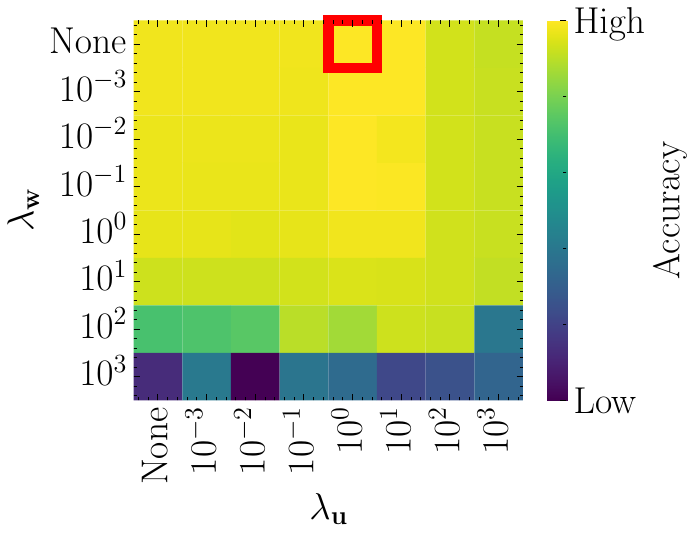}&
\includegraphics[width=1.85cm]{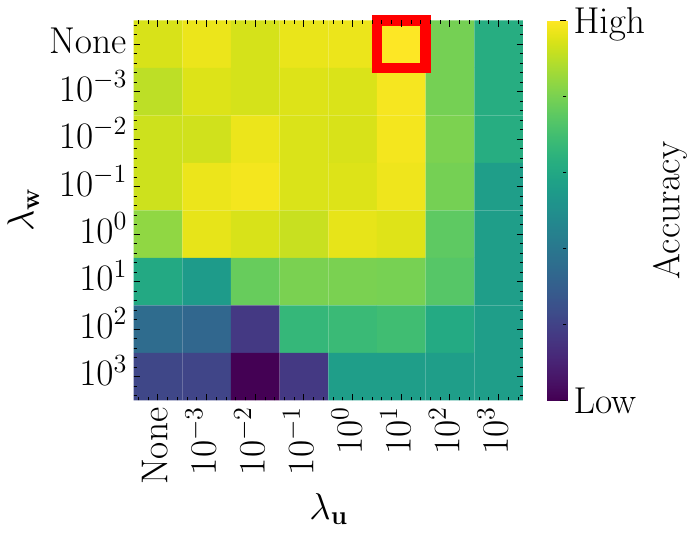}&
\includegraphics[width=1.85cm]{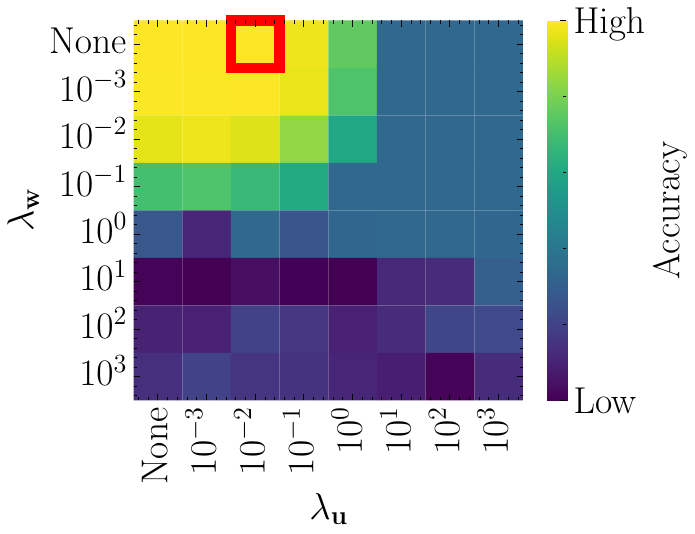}&
\includegraphics[width=1.85cm]{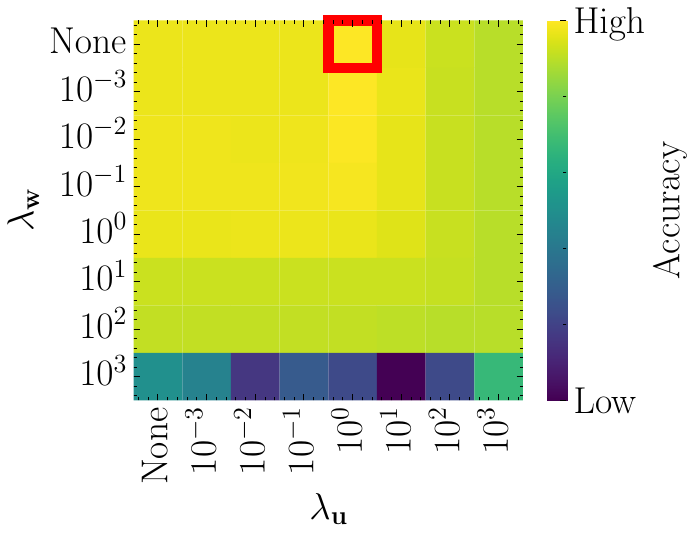}\\
Test&
\includegraphics[width=1.85cm]{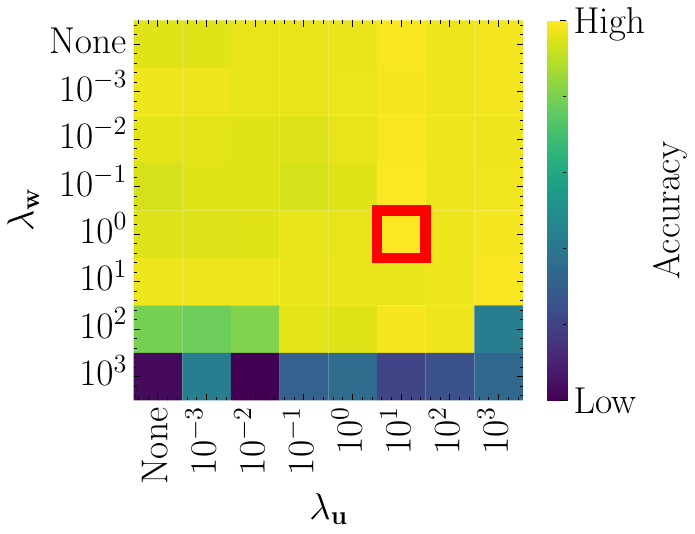}&
\includegraphics[width=1.85cm]{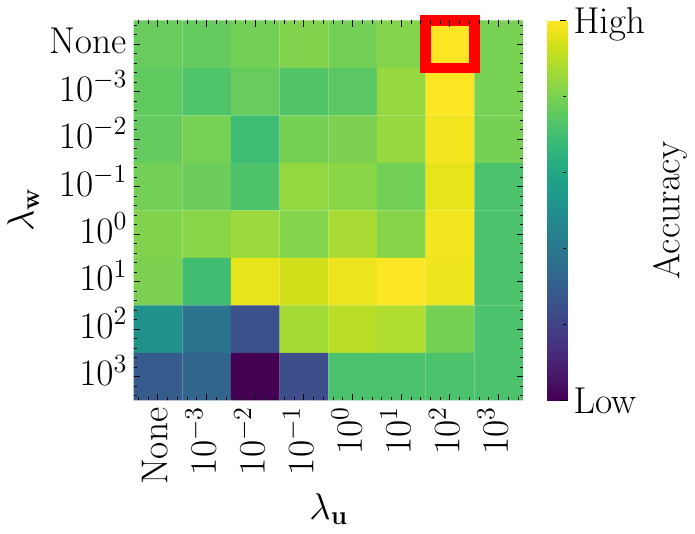}&
\includegraphics[width=1.85cm]{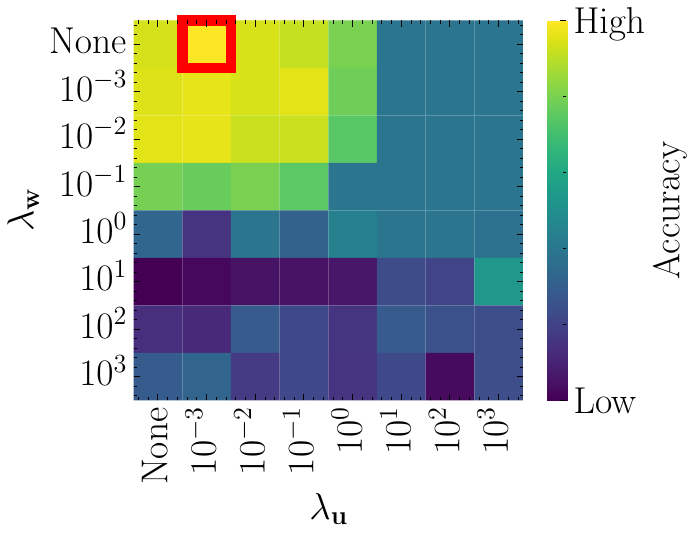}&
\includegraphics[width=1.85cm]{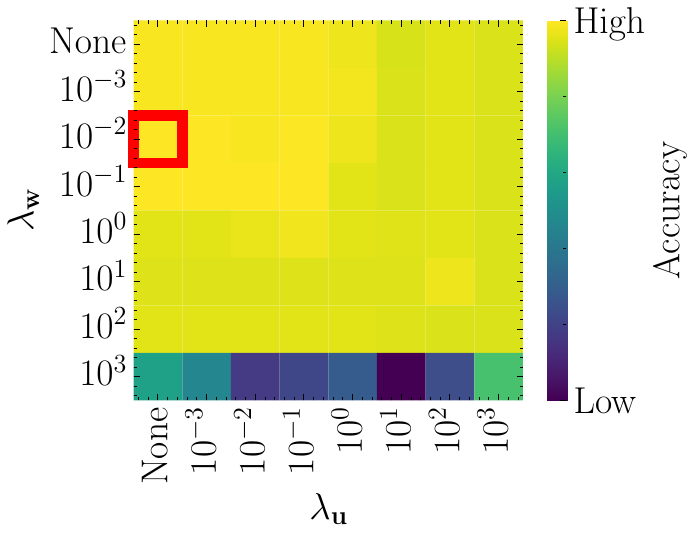}\\
& Heart Disease & Vowel & Delhi Weather & Boston Housing \\
Training&
\includegraphics[width=1.85cm]{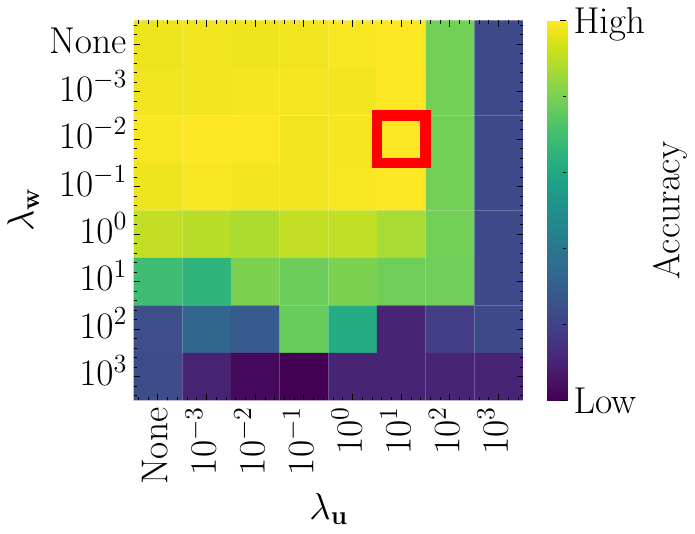}&
\includegraphics[width=1.85cm]{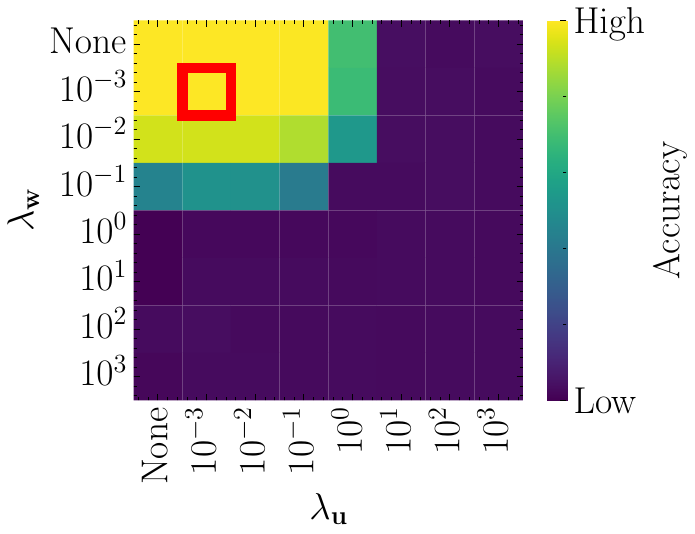}&
\includegraphics[width=1.85cm]{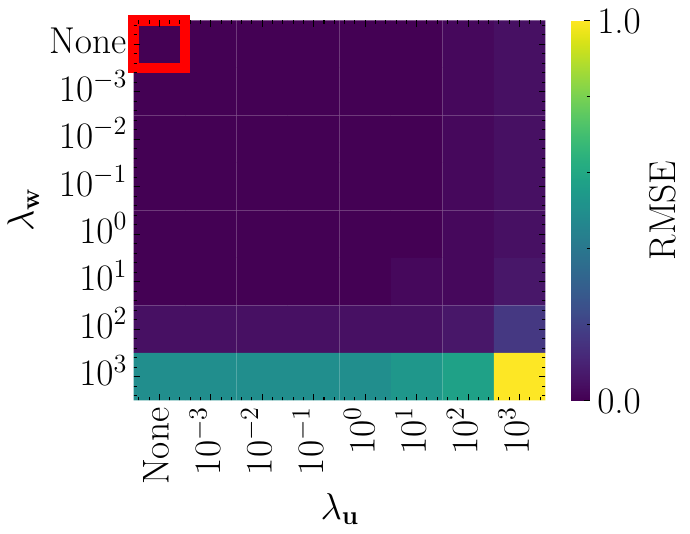}&
\includegraphics[width=1.85cm]{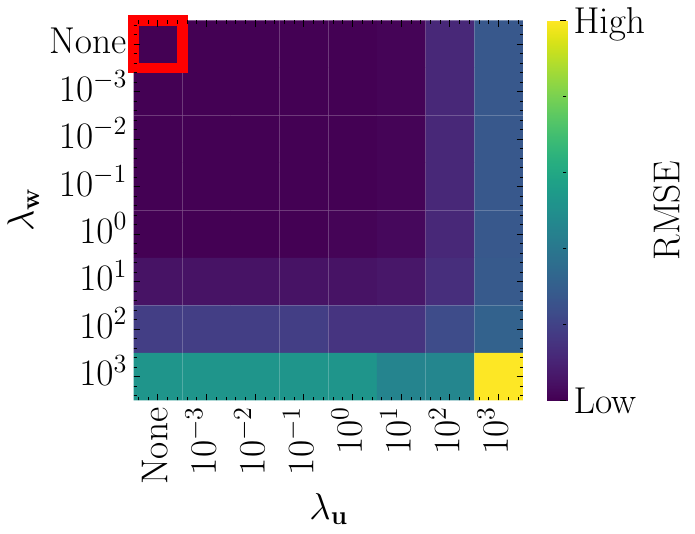}\\
Test&
\includegraphics[width=1.85cm]{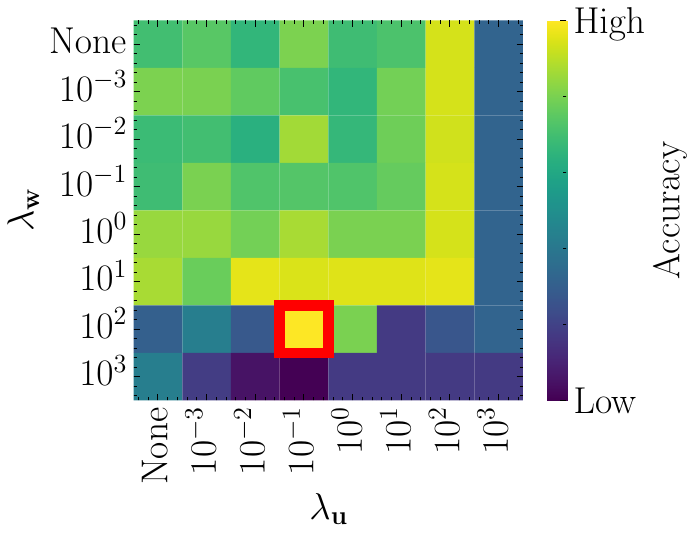}&
\includegraphics[width=1.85cm]{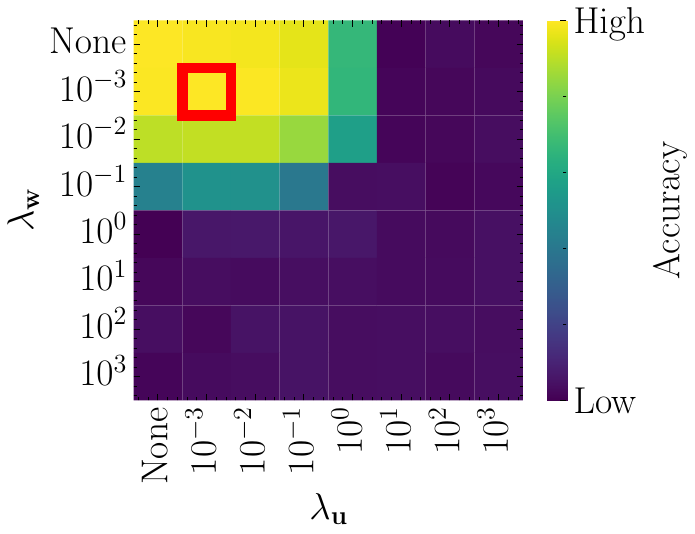}&
\includegraphics[width=1.85cm]{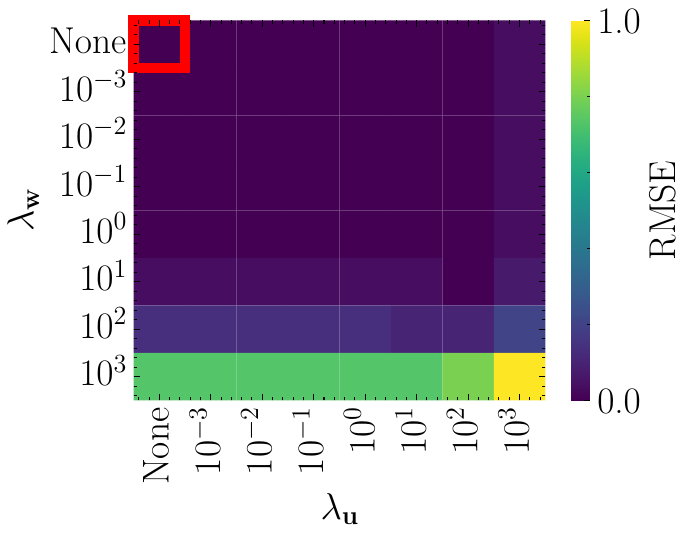}&
\includegraphics[width=1.85cm]{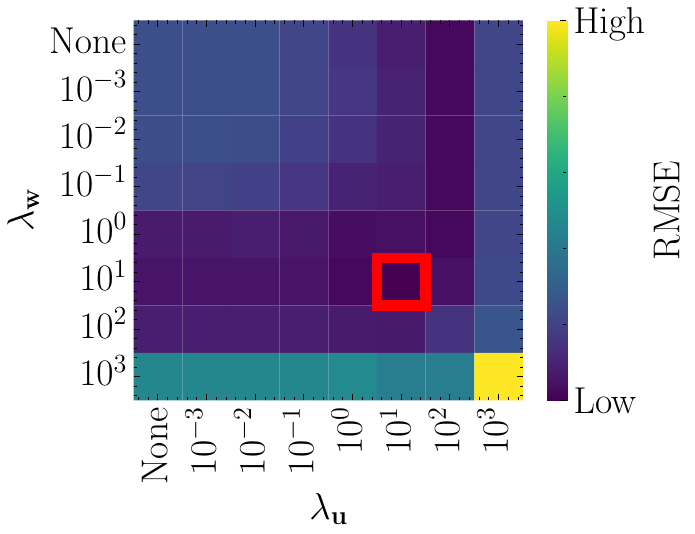}\\
 & Diabetes & Prostatic Cancer & Liver & Plasma\\
 Training&
\includegraphics[width=1.85cm]{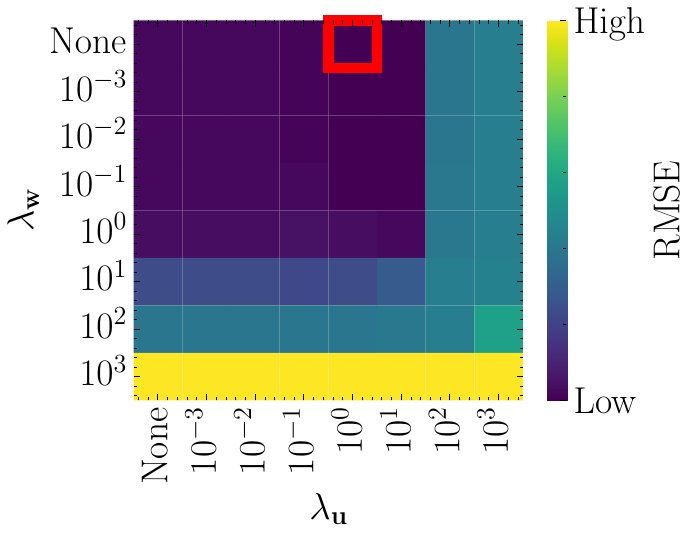}&
\includegraphics[width=1.85cm]{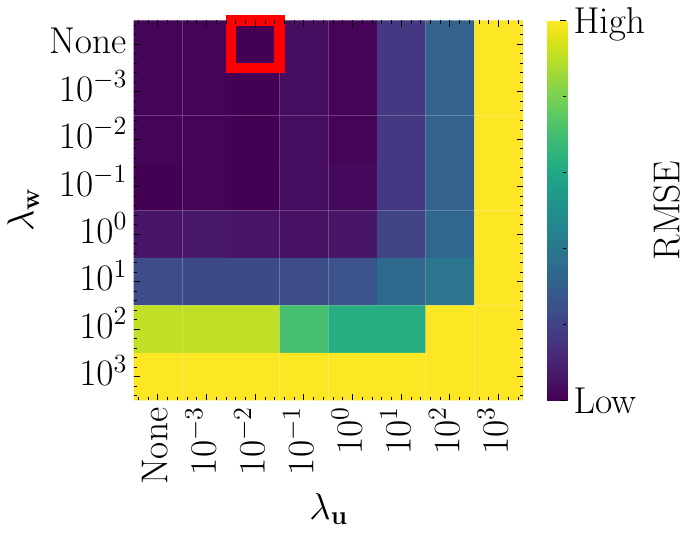}&
\includegraphics[width=1.85cm]{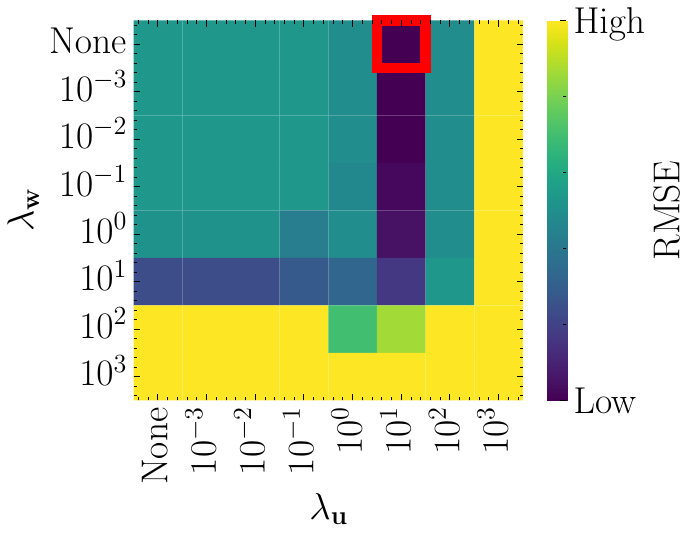}&
\includegraphics[width=1.85cm]{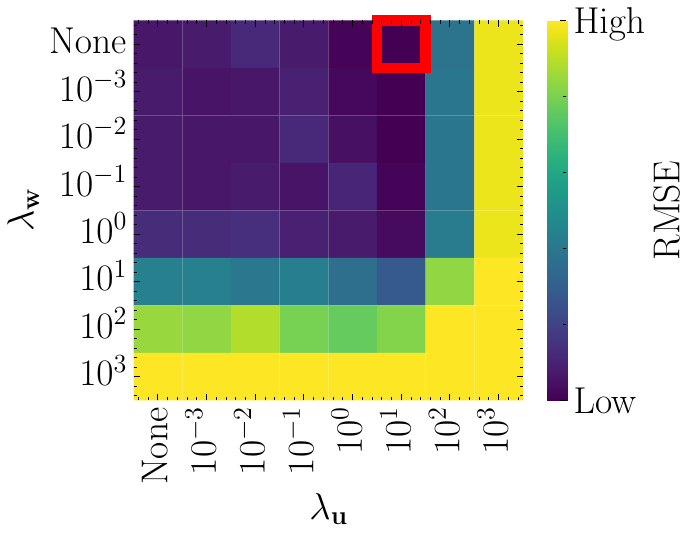}\\
Test&
\includegraphics[width=1.85cm]{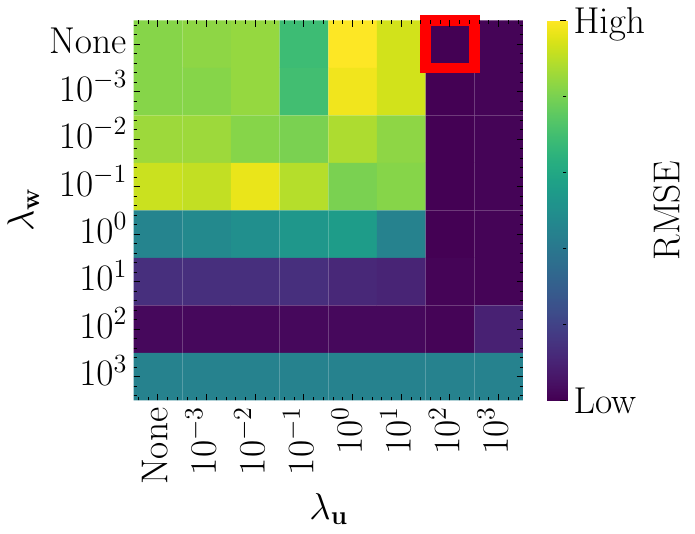}&
\includegraphics[width=1.85cm]{figs/reg_analysis_test_prostate_cancer_GR-RBF.pdf}&
\includegraphics[width=1.85cm]{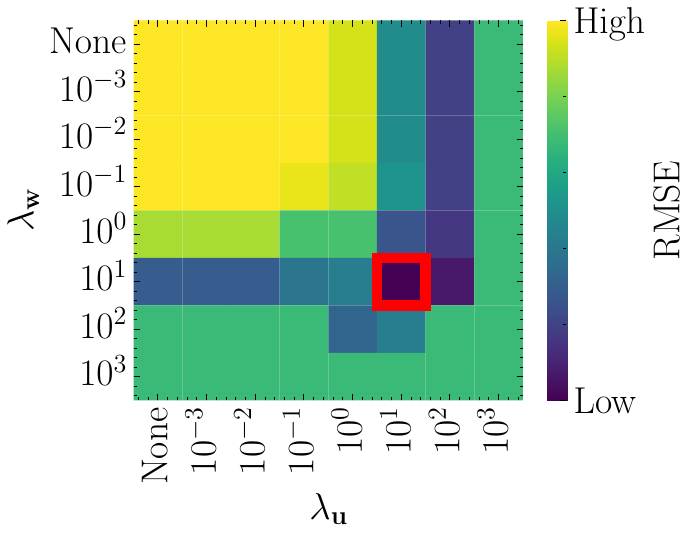}&
\includegraphics[width=1.85cm]{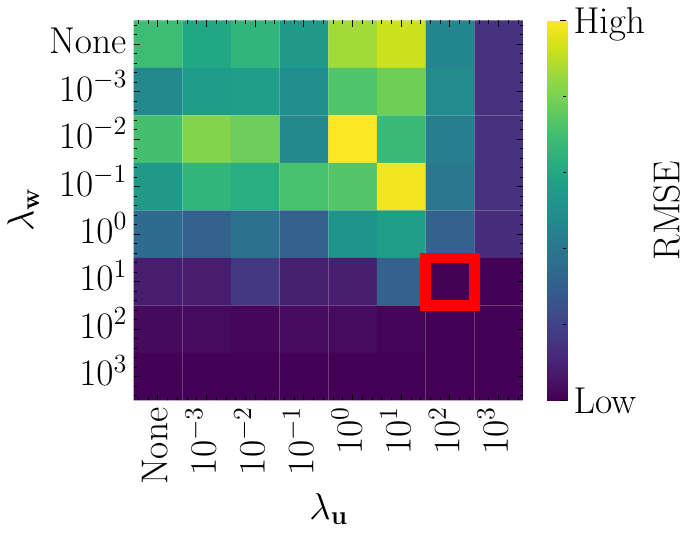}\\
 & Cloud & DTMB-$5415^{(1)}$ &DTMB-$5415^{(2)}$& Body Fat\\
 Training&
\includegraphics[width=1.85cm]{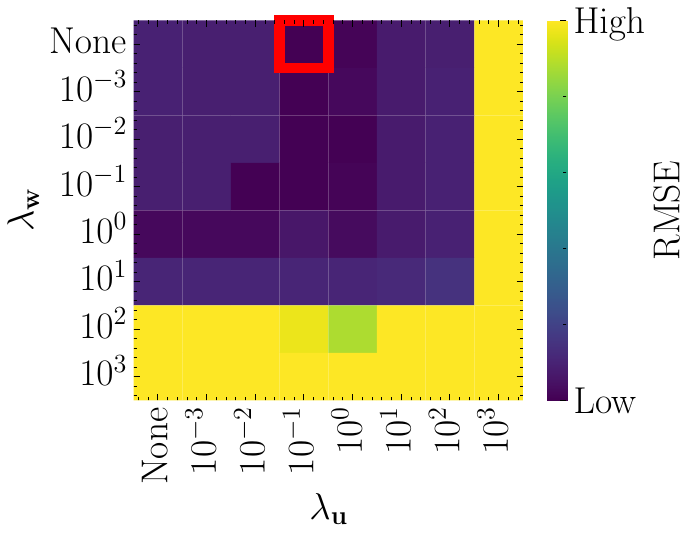}&
\includegraphics[width=1.85cm]{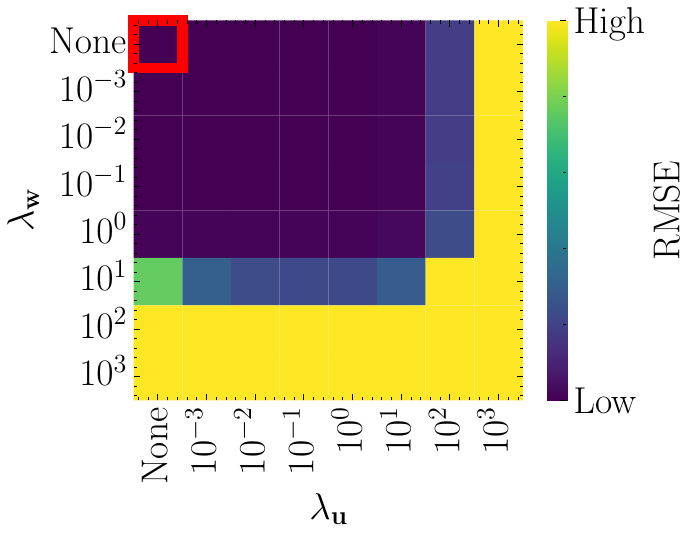}&
\includegraphics[width=1.85cm]{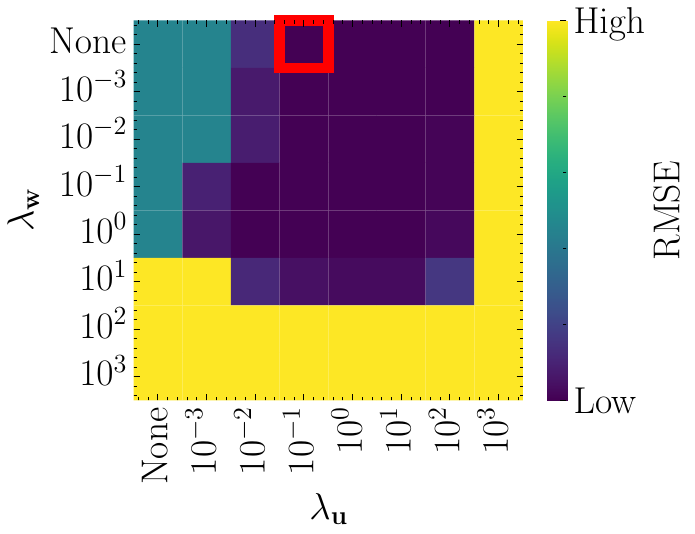}&
\includegraphics[width=1.85cm]{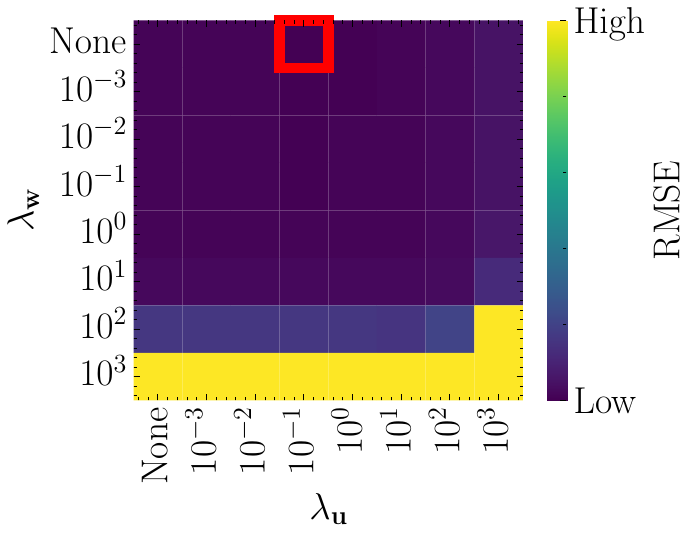}\\
Test&
\includegraphics[width=1.85cm]{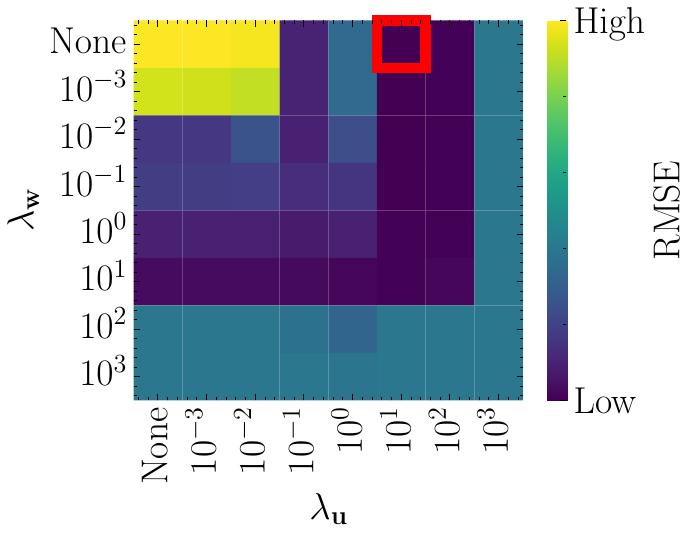}&
\includegraphics[width=1.85cm]{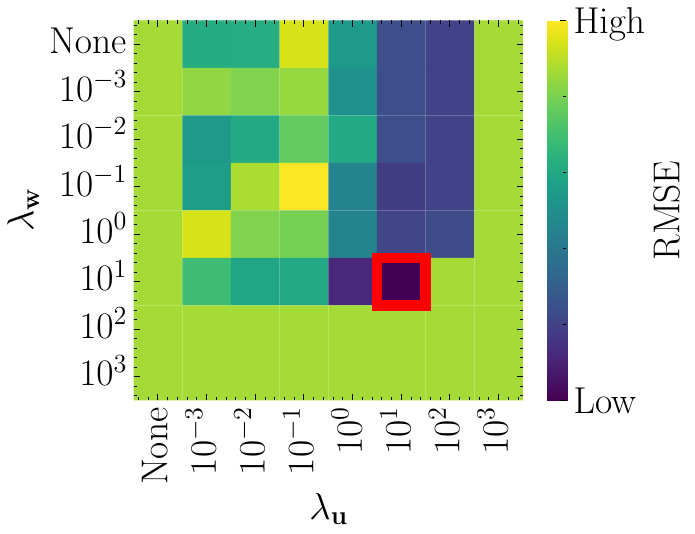}&
\includegraphics[width=1.85cm]{figs/reg_analysis_train_DTMB_5415_GR-RBF.pdf}&
\includegraphics[width=1.85cm]{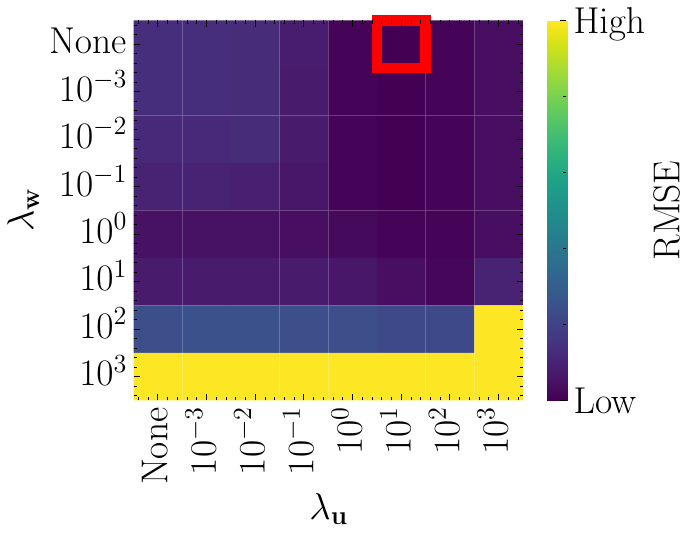}\\
 &
\end{tabular}
\caption{GRBFNN's regularization behavior: dark color for lower RMSE (regression) and lighter for higher accuracy (classification). The red frame indicates the best regularization combination.}
\label{fig:reg_analysis}
\end{figure}
\begin{figure}[!htbp]   
\centering
\setlength{\tabcolsep}{12pt}
\begin{tabular}{l*4{C}@{}}
& Digits & Iris & Breast cancer & Wine \\
Active subspace &
\includegraphics[width=1.96cm]{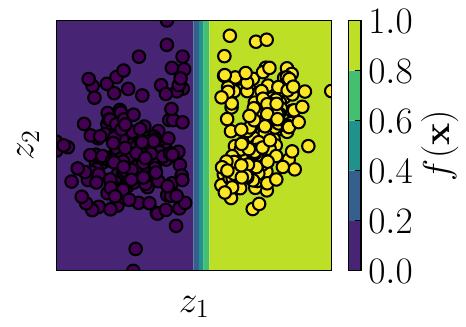}&
\includegraphics[width=1.96cm]{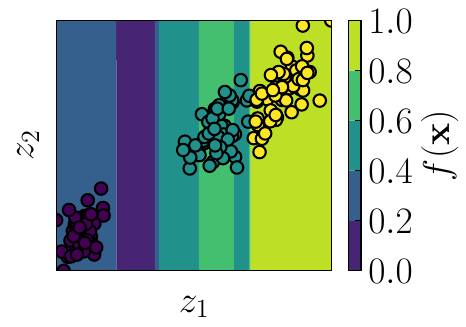}&
\includegraphics[width=1.96cm]{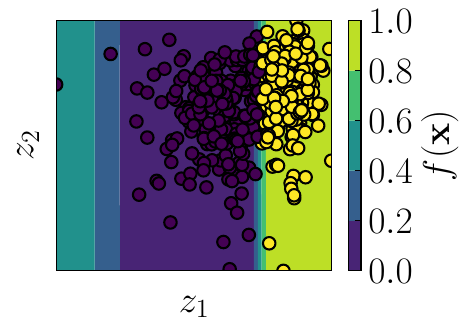}&
\includegraphics[width=1.96cm]{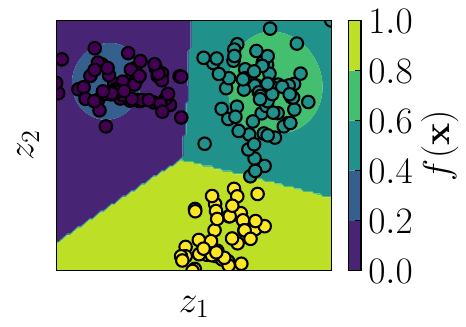}\\
Eigenvalues decay&
\hspace*{-0.55cm}\includegraphics[width=1.96cm]{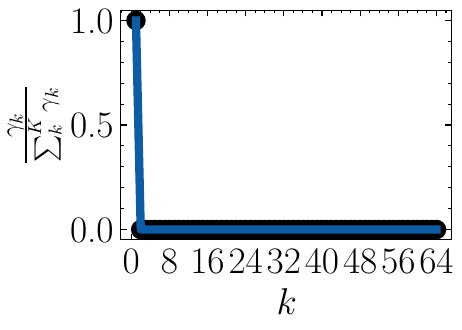}&
\hspace*{-0.55cm}\includegraphics[width=1.96cm]{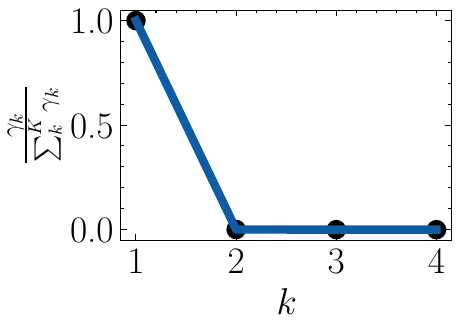}&
\hspace*{-0.55cm}\includegraphics[width=1.96cm]{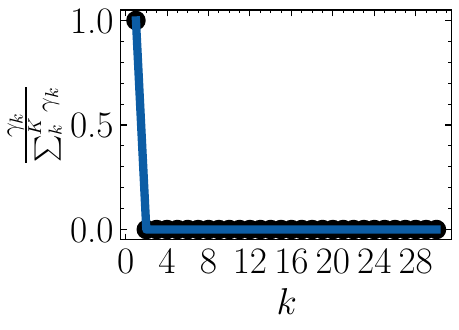}&
\hspace*{-0.55cm}\includegraphics[width=1.96cm]{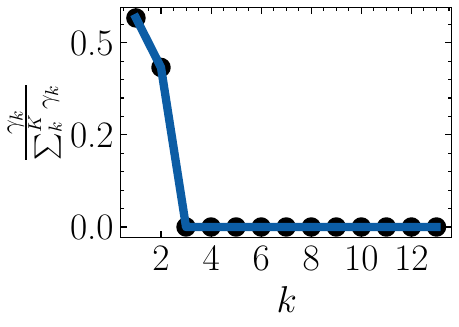}\\
& Australian & Credit-g & Glass & Blood\\
Active subspace&
\includegraphics[width=1.96cm]{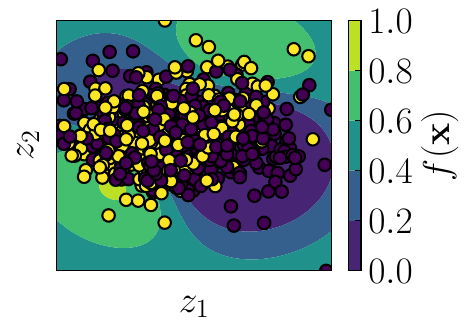}&
\includegraphics[width=1.96cm]{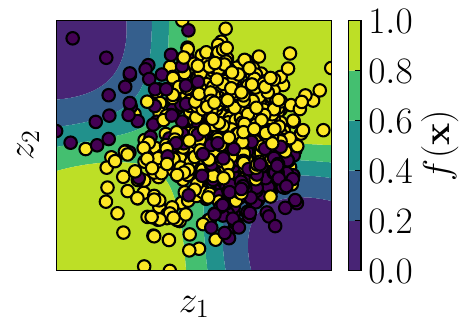}&
\includegraphics[width=1.96cm]{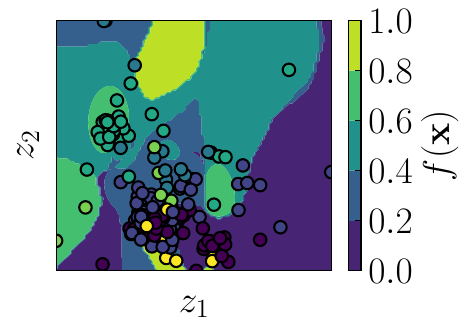}&
\includegraphics[width=1.96cm]{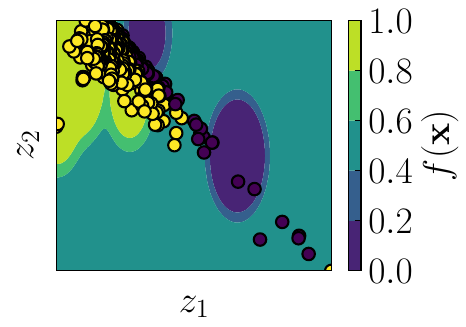}\\
Eigenvalues decay&
\hspace*{-0.55cm}\includegraphics[width=1.96cm]{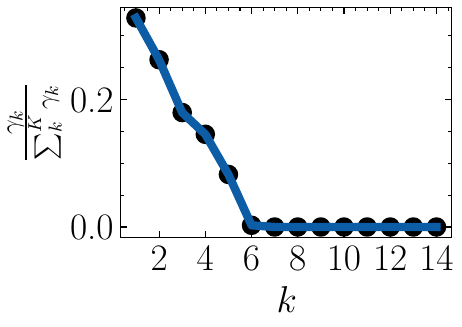}&
\hspace*{-0.55cm}\includegraphics[width=1.96cm]{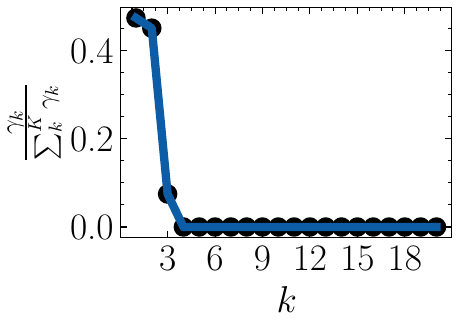}&
\hspace*{-0.55cm}\includegraphics[width=1.96cm]{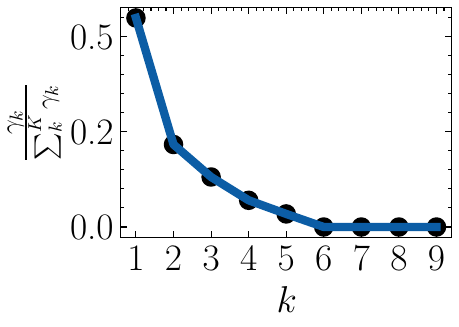}&
\hspace*{-0.55cm}\includegraphics[width=1.96cm]{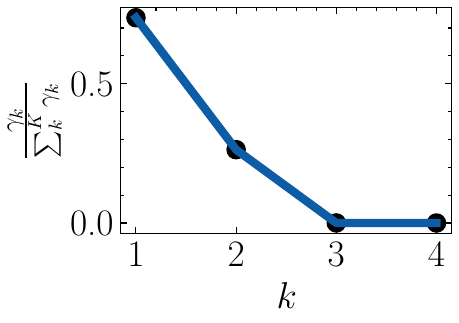}\\
&Heart Disease & Vowel & Delhi Weather & Boston Housing \\
Active subspace&
\includegraphics[width=1.96cm]{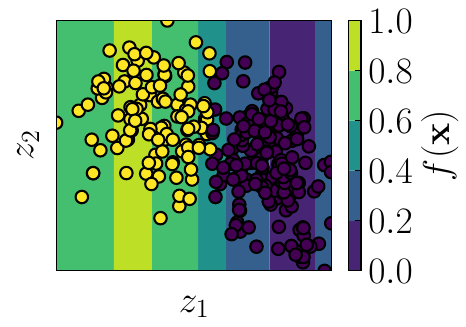}&
\includegraphics[width=1.96cm]{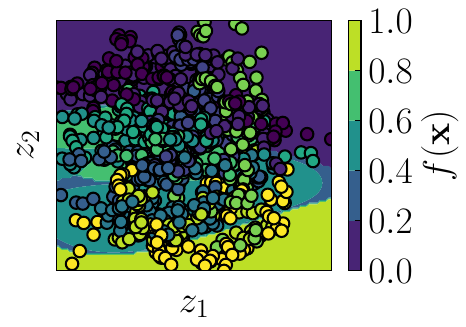}&
\includegraphics[width=1.96cm]{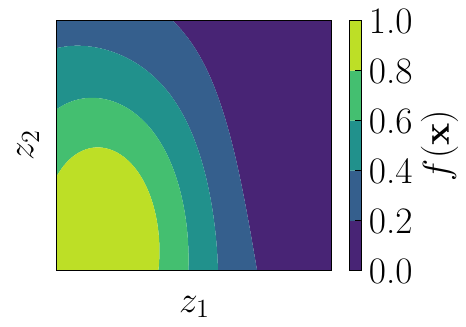}&
\includegraphics[width=1.96cm]{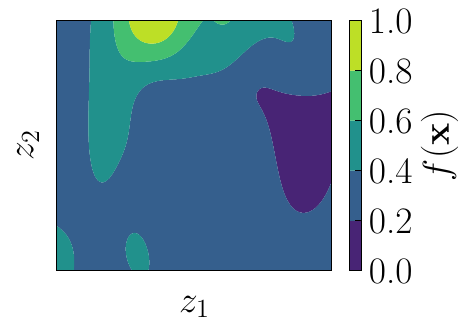}\\
Eigenvalues decay&
\hspace*{-0.55cm}\includegraphics[width=1.96cm]{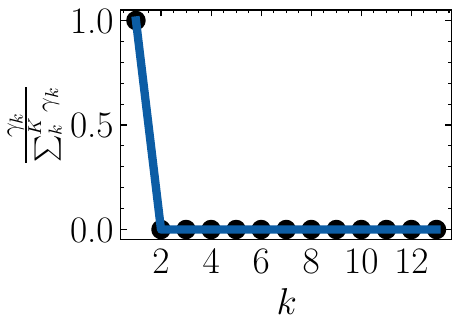}&
\hspace*{-0.55cm}\includegraphics[width=1.96cm]{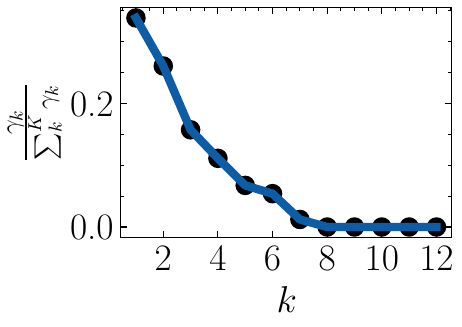}&
\hspace*{-0.55cm}\includegraphics[width=1.96cm]{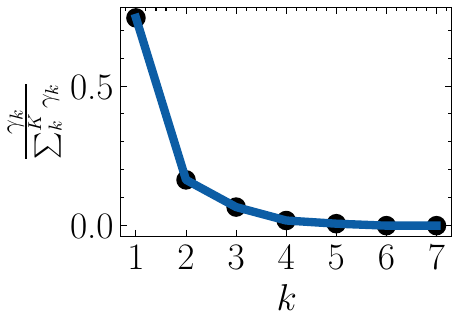}&
\hspace*{-0.55cm}\includegraphics[width=1.96cm]{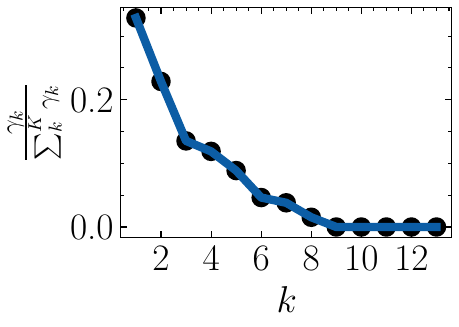}\\
&Diabetes & Prostate cancer & Liver & Plasma\\
Active subspace&
\includegraphics[width=1.96cm]{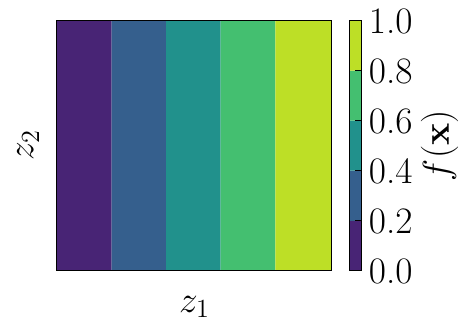}&
\includegraphics[width=1.96cm]{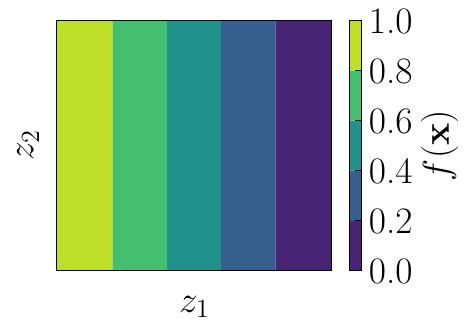}&
\includegraphics[width=1.96cm]{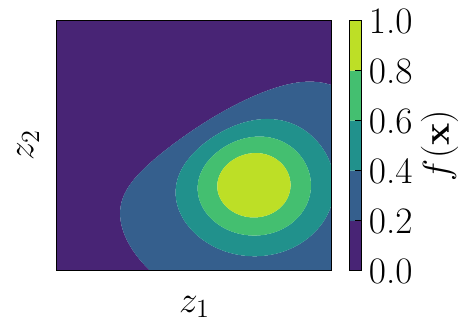} &
\includegraphics[width=1.96cm]{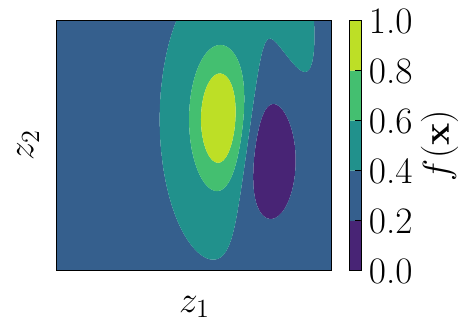}\\
Eigenvalues decay&
\hspace*{-0.55cm}\includegraphics[width=1.96cm]{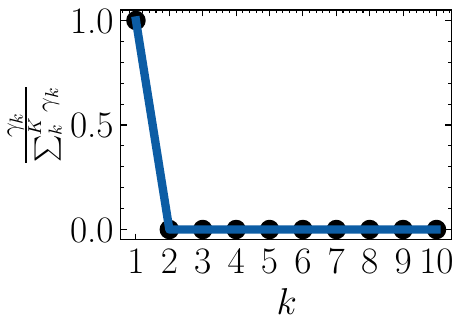}&
\hspace*{-0.55cm}\includegraphics[width=1.96cm]{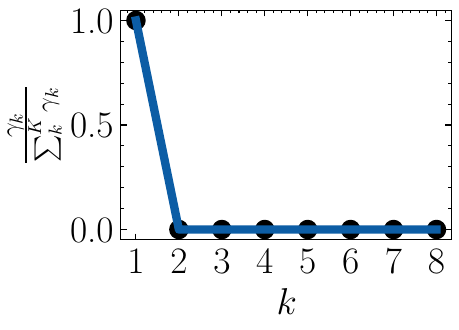}&
\hspace*{-0.55cm}\includegraphics[width=1.96cm]{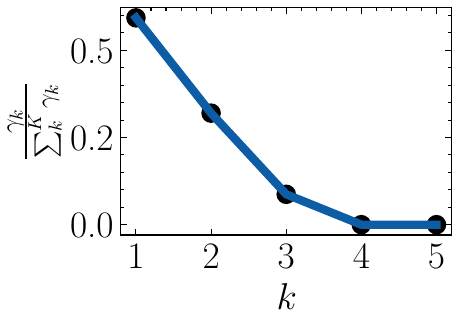}&
\hspace*{-0.55cm}\includegraphics[width=1.96cm]{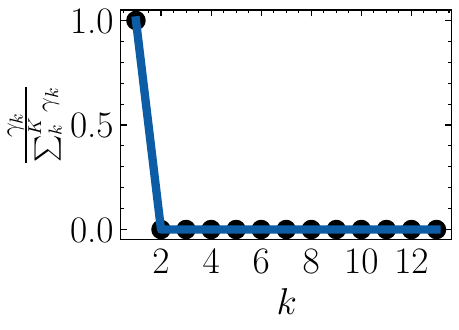}\\
& Cloud & DTMB-$5415^{(1)}$&DTMB-$5415^{(2)}$ & Body Fat\\
Active subspace&
\includegraphics[width=1.96cm]{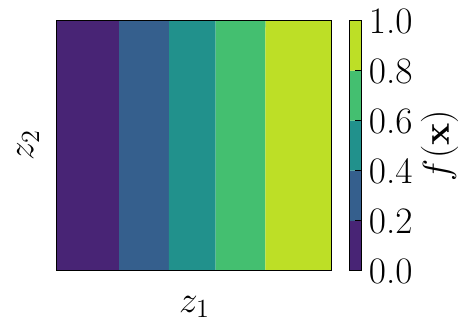}&
\includegraphics[width=1.96cm]{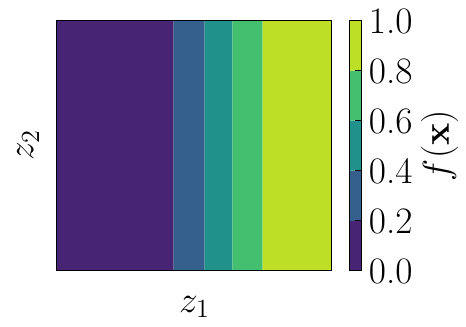}&
\includegraphics[width=1.96cm]{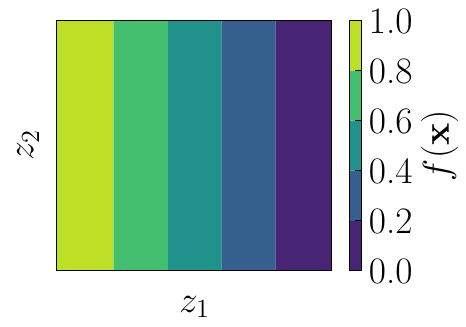}&
\includegraphics[width=1.96cm]{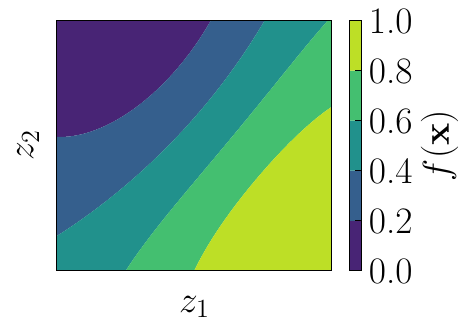}\\
Eigenvalues decay&
\hspace*{-0.55cm}\includegraphics[width=1.96cm]{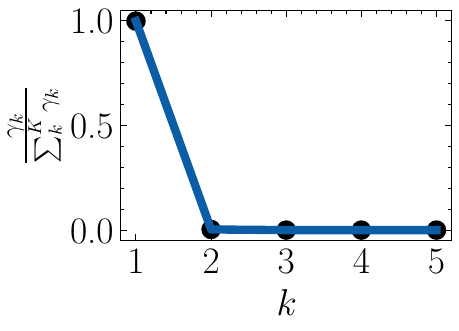}&
\hspace*{-0.55cm}\includegraphics[width=1.96cm]{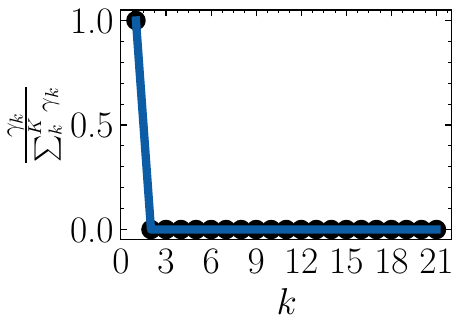}&
\hspace*{-0.55cm}\includegraphics[width=1.96cm]{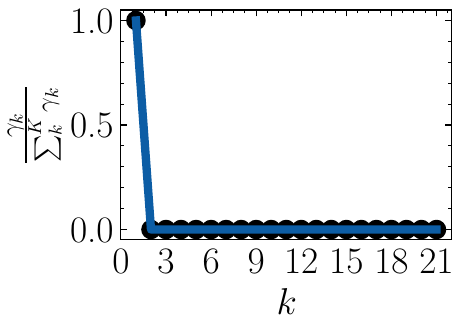}&
\hspace*{-0.55cm}\includegraphics[width=1.96cm]{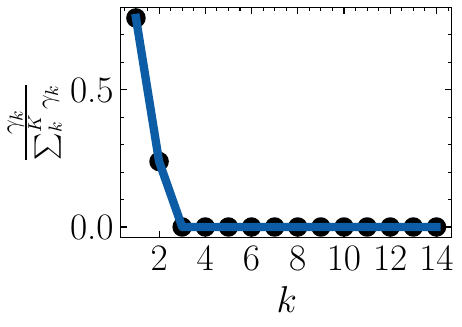}\\
\end{tabular}
\caption{Graphical interpretation of the active subspace in two dimensions in the contour plots and corresponding eigenvalues decay. Function values are normalized.}
\label{fig:eigen_embedding}
\end{figure}
\end{document}